\documentclass[a4paper,12pt,twoside,hidelinks,openright]{report}

\usepackage[english]{babel}
\usepackage[utf8]{inputenc}
\usepackage[usenames, dvipsnames]{color}
\usepackage{graphicx}
\usepackage{enumerate}
\usepackage{multirow}
\usepackage{graphics}
\usepackage{appendix}
\usepackage[nottoc]{tocbibind}
\DeclareMathAlphabet{\pazocal}{OMS}{zplm}{m}{n}
\usepackage{rotating}
\newcommand{\Lb}{\pazocal{L}}
\usepackage{times}
\usepackage{epsfig}
\usepackage{amsmath}
\usepackage{amssymb}
\usepackage[normalem]{ulem}
\useunder{\uline}{\ul}{}
\usepackage{nicefrac}
\usepackage{balance}
\usepackage{wrapfig}
 
\usepackage{tabularx}
\usepackage{booktabs}
\usepackage{tikz}
\usetikzlibrary{shapes,arrows}
\usepackage{subcaption}
\usepackage[final]{pdfpages}
\usepackage[]{placeins,flafter}
\usepackage[none]{hyphenat} 
\usepackage{xcolor}
\usepackage{adjustbox}
\usepackage[authoryear]{natbib}
\usepackage{hyperref}
\usepackage{makeidx}
\makeindex
\index{key}

\newenvironment{dedicatory}
  {\vspace*{6cm}\begin{quotation}\begin{center}\begin{em}}
  {\par\end{em}\end{center}\end{quotation}}

\begin{document}


\begin{titlepage}

\vspace*{-4mm}
\begin{figure}[!h]
  \centering
	\includegraphics[width=69.62mm]{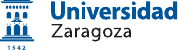}
\end{figure}

\vspace*{17mm}

\fontsize{28pt}{28pt}\selectfont
\begin{center}
\setlength{\fboxsep}{3.4mm}
\adjustbox{minipage=14.4cm,cfbox=blue,center}{\begin{center}Tesis Doctoral\end{center}}
\end{center}

\vspace*{18.7mm}

\fontsize{20pt}{20pt}\selectfont
\begin{center}
Uncertainty and Self-Supervision in Single-View Depth
\end{center}

\vspace*{0cm} 
\baselineskip 36pt
\begin{center}
\fontsize{12pt}{12pt}\selectfont
\center{\rm  Autor}

\vspace*{3.65mm} 
\fontsize{18pt}{18pt}\selectfont
\center{Javier Rodríguez Puigvert}
\vspace*{1cm}
\baselineskip 36pt
\fontsize{12pt}{12pt}\selectfont
\center{\rm  Directores}
\vspace*{3.56mm}
\fontsize{14pt}{14pt}\selectfont
\center{Javier Civera Sancho}
\center{Rubén Martínez Cantín}
\end{center}

\setcounter{footnote}{1}

\vspace*{16.45mm}
\fontsize{12pt}{12pt}\selectfont
\begin{center}
ESCUELA DE DOCTORADO\\
2023\\
\end{center}

\renewcommand{\thefootnote}{\arabic{footnote}}
\end{titlepage}
\newpage

\title{Título del trabajo final de máster}
\author{Autor Apellido Apellido}

\pagebreak
\cleardoublepage
\baselineskip 19pt

\renewcommand{\labelitemi}{$-$}
\renewcommand{\tablename}{Table}


\begin{dedicatory}
Für Nina, meine Frau.\\A Carolina y Nuria, mis hermanas.\\ A Nuria y Alfonso, mis padres.
\end{dedicatory}
\cleardoublepage

{\Large \bfseries Agradecimientos}
\vspace{1.5cm}

Hace cuatro años, decidí embarcarme en un camino apasionante, lleno de desafíos, dejando atrás una vida acomodada en Alemania, pero llevándome lo mejor de ella. Mi mujer, Nina, muchas gracias por estar a mi lado y apoyarme incondicionalmente. Gracias por atreverte a tomar la valiente decisión de trasladarnos a Zaragoza y emprender esta aventura, \textit{fortis fortuna iuvat}.

Me gustaría expresar mi gratitud más sincera a mis directores Javier y Rubén, por motivarme y guiarme durante estos años de mi tesis. Muchas gracias por vuestro tiempo y dedicación. Bajo vuestra supervisión, he aprendido las habilidades analíticas  para desafiar y abordar problemas complejos.  A ti Javier, muchas gracias por darme la oportunidad de hacer el doctorado bajo tu supervisión, sin conocerme demasiado. Todavía me acuerdo de la primera conversación que tuvimos en el ISMAR de Munich, hace ya 5 años y la trascendencia que tuvo en mi vida.

Quisiera agradecer a Pascal, Mingo y Monti, por darme siempre un feedback sincero y constructivo. Trabajando con vosotros, me habéis trasladado la importancia de tener un pensamiento crítico y positivo sobre la investigación. Quisiera agradecer también a César que me diera la oportunidad de trabajar junto a él durante mi estancia en la ETH Zurich.
Muchas gracias a todos los compañeros de unizar y  del L.108, en especial a, Juanjo, Sergio, Julio, Richard, David, León, Edu, Lorenzo, Tomás, Morlana, Victor y Bruno. Nuestras conversaciones siempre han sido super enriquecedoras, siempre he aprendido algo nuevo con vosotros.

Gracias mis amigos Tono, por ayudarme con cosas de Unity, Alberto, por ayudarme distintas discusiones y Arnau por ayudarme con su conocimiento de programación web. A David y Luis Miguel, por estar ahí desde 2008 cuando empezábamos nuestro camino para ser ingenieros. 

Muchas gracias a mi Familia, Rodríguez, Puigvert y Hoppmann, por su amor y apoyo. Especialmente a mis hermanas Nuria y Carolina que siempre han sido un ejemplo para mí. 

Finalmente, este doctorado está dedicado a mis padres Nuria y Alfonso.
Mi padre, Alfonso Rodríguez Cid, la persona más perseverante que he conocido, inteligente y perspicaz donde los haya. Siento tu apoyo diario, aunque no estés con nosotros.

\newpage
{\Large \bfseries Resumen}

\vspace{1cm}
La estimación de profundidad a partir de una sola vista se refiere a la capacidad de obtener información tridimensional por píxel a partir de una imagen RGB bidimensional. La profundidad a partir de una sola vista tiene gran relevancia en una amplia gama de aplicaciones en campos como la realidad aumentada, la realidad virtual, la robótica, la medicina o la conducción autónoma.
Desde un punto de vista geométrico, la reconstrucción 3D a partir de una imagen es un problema mal condicionado, porque existen múltiples soluciones de profundidad que explican una imagen 2D.
Recientemente, las redes neuronales profundas han demostrado ser muy eficaces para aprender los patrones de apariencia visual que corresponden con ciertas estructuras 3D. Sin embargo, la mayoría de los métodos son deterministas y no estiman la incertidumbre asociada a sus predicciones. Esto puede tener consecuencias desastrosas cuando se aplica a campos como la conducción autónoma o la robótica médica. 
En esta tesis hemos abordado este problema cuantificando la incertidumbre de la profundidad de una sola vista para redes neuronales profundas bayesianas.

La estimación de la profundidad en una sola vista puede ser muy eficaz cuando se dispone de suficientes datos de profundidad anotados para el entrenamiento supervisado. Sin embargo, existen escenarios, por ejemplo nuestra aplicación con imágenes endoscópicas, en los que no es posible obtener dichos datos. La estimación de profundidad a partir de imágenes endoscópicas es un requisito para una amplia gama de tecnologías de asistencia al personal médico basadas en IA, como la localización y medición precisas de tumores o la identificación de zonas no inspeccionadas.
En esta tesis, presentamos un método que facilita la transición de los métodos entrenados en datos sintéticos al dominio real teniendo en cuenta las incertidumbres asociadas. En concreto, introducimos una arquitectura maestro-estudiante (teacher-student en inglés) consciente de la incertidumbre que se entrena de forma autosupervisada, teniendo en cuenta la incertidumbre del maestro entrenado en un conjunto de datos sintéticos.

Finalmente, en algunos dispositivos médicos como los endoscopios, la cámara y las fuentes de luz están rígidamente unidas y situadas a una pequeña distancia de las superficies a explorar. Como última contribución de esta tesis doctoral, modelamos el hecho de que para cualquier albedo y superficie dados, el brillo del píxel es inversamente proporcional al cuadrado de la distancia a la superficie y proponemos el uso de la iluminación como una potente señal de autosupervisión de una sola vista para redes neuronales profundas.

\vspace{2.5cm}
\newpage
\cleardoublepage
{\Large \bfseries Acknowledgements}
\vspace{1.5cm}

Four years ago, I decided to embark on an exciting and challenging journey, leaving behind a comfortable life in Germany, but taking the best of it with me. My wife, Nina, thank you so much for being by my side and supporting me unconditionally. Thank you for bravely taking the courageous decision to move to Zaragoza and start this adventure (fortis fortuna iuvat). I would like to express my most sincere gratitude to my directors Javier and Rubén, for motivating and guiding me during these years of my thesis. Thank you very much for your time and dedication. Under your supervision, I have learned the analytical skills to challenge and tackle complex problems. I still remember the first conversation we had at ISMAR in Munich, 5 years ago now, and how significant it was in my life.
I would like to thank Pascal, Mingo and Monti for always giving me honest and constructive feed-back. By working with you, you have shown me how important it is to think critically and positively about research.
I would also like to thank César for giving me the opportunity to work with him during my stay at ETH Zurich. Many thanks to all my colleagues at unizar and L.108, especially Juanjo, Sergio, Julio, Richard, David, León, Edu, Lorenzo, Tomás, Morlana, Victor and Bruno. Our conversations have always been very enriching, I have always learned something new from you.

Thanks to my friends Tono, for helping me with Unity stuff, Alberto, for helping me with different discussions and Arnau for helping me with his programming knowledge. To David and Luis Miguel, for being there since 2008 when we were starting our way to become engineers. 

Many thanks to my family, Rodríguez, Puigvert and Hoppmann, for their love and support. Especially to my sisters Nuria and Carolina who have always been an inspiration to me. 
Finally, this PhD is dedicated to my parents Nuria and Alfonso.
My father, Alfonso Rodríguez Cid, is the most persevering person I have ever met. I feel your daily support, even though you are not with us.
\newpage
{\Large \bfseries Abstract}

\vspace{1cm}
Single-view depth estimation refers to the ability to derive three-dimensional information per pixel from a single two-dimensional RGB image. Estimating depth from a single image is required in a wide range of applications in fields such as augmented reality, virtual reality, robotics, medicine or autonomous driving. 

Single-view depth estimation is an ill-posed problem because there are multiple depth solutions that explain 3D geometry from a single view. While deep neural networks have been shown to be effective at capturing depth from a single view, the majority of current methodologies are deterministic in nature. Accounting for uncertainty in the predictions can avoid disastrous consequences when applied to fields such as autonomous driving or medical robotics. We have addressed this problem by quantifying the uncertainty of supervised single-view depth for Bayesian deep neural networks.
When there is enough ground truth depth data for supervised training, single-view depth estimation can be remarkably effective. There are scenarios, especially in medicine in the case of endoscopic images, where such annotated data is not available. Nevertheless, the estimation of depth information from endoscopic images is a prerequisite for a wide range of AI-based technologies, such as the accurate localisation and measurement of tumours or the identification of non-inspected areas. 

To alleviate the lack of data,  we present a method that improves the transition from synthetic to real domain methods. We introduce an uncertainty-aware teacher-student architecture that is trained in a self-supervised manner, taking into account the teacher uncertainty trained on a synthetic dataset. 

Given the vast amount of unannotated data and the challenges associated with capturing annotated depth in medical minimally invasive procedures, we advocate a fully self-supervised approach that only requires RGB images and the geometric and photometric calibration of the endoscope.
In endoscopic imaging, the camera and light sources are co-located at a small distance from the target surfaces. This setup indicates that brighter areas of the image are nearer to the camera, while darker areas are further away. Building on this observation, we exploit the fact that for any given albedo and surface orientation, pixel brightness is inversely proportional to the square of the distance. We propose the use of illumination as a strong single-view self-supervisory signal for deep neural networks.

\vspace{2.5cm}
\newpage
\newpage

\cleardoublepage
\renewcommand{\contentsname}{Index}
\tableofcontents


\chapter{Introduction}
\begin{figure}
    \centering
    \includegraphics[width=0.9\textwidth]{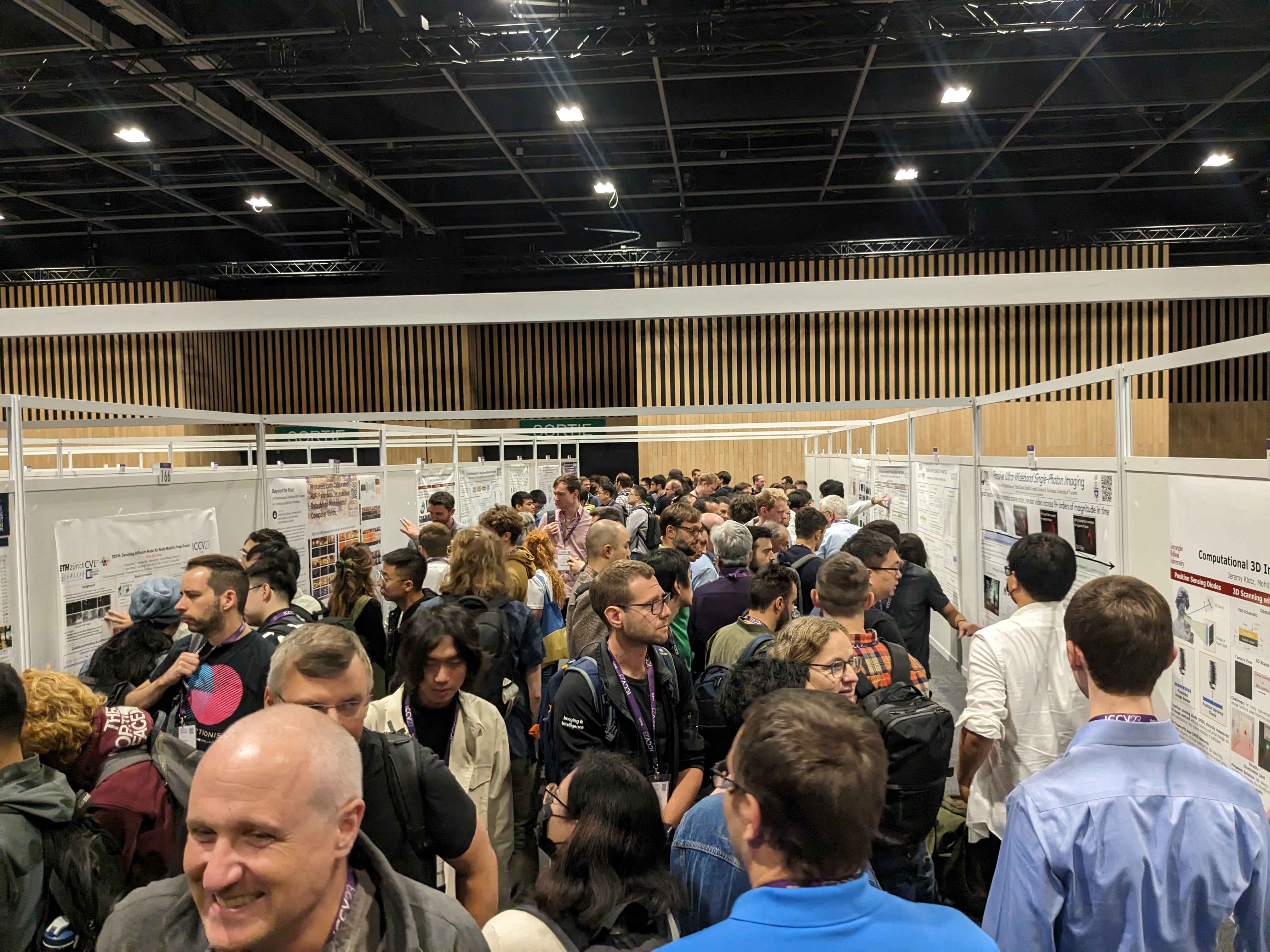}
    \caption{ICCV 2023 Poster Session }
    \label{fig:iccv}
\end{figure}

Vision is tremendously powerful; we as humans are able to navigate through the world and most often it feels effortless, as our visual system has evolved through millennia to process information seamlessly. However, even if humans generally excel at making qualitative judgments about visual scenes, we are not so good on quantifying our perceptions. For example, we can easily qualify an object as near or far, but when it comes to an accurate quantitative measurement, such as estimating an object's distance in centimetres, we might be off by a gross margin. Looking at the more especific example in Figure \ref{fig:iccv}, it is easy for us to understand the rough geometry and the semantic structure of the scene. We can make spatial sense of the space, inferring a back wall at the end of the two rows of posters, plan a path to get there and navigate through the scene, even if it is crowded. Such understanding is an extremely complex task, only made possible by our prior knowledge and our extraordinary pattern recognition abilities.

Computer vision aims to emulate, and hopefully surpass, human visual perception. In fact, computer vision's capabilities have already begun to outpace human abilities in specific areas \citep{kaufmann2023champion}.
The field of computer vision is currently on a growth curve never seen before. 
From the early days of deep learning, when convolutional neural networks were used for image classification of some simple entities \citep{lecun2015deep}, to recent advances in visual transformers \citep{dosovitskiy2020image} that masterfully create features with a consistent global perspective across images or Neural Radiance Fields \citep{mildenhall2020nerf} (NERFs), the field of computer vision is experiencing an extraordinary evolution. Even more recently, the introduction of diffusion models \citep{ho2020denoising} has started yet another revolution in generative modelling and opened up new avenues in image synthesis and rendering.

As the capabilities of computer vision continue to expand and evolve, we expect it to have a disruptive effect on technologies such as vision-assisted robotics or medical image analysis.
However, and despite the impressive recent advances, several computer vision tasks remain unsolved. Specifically and among others, single-view depth estimation still presents significant challenges. Although depth can also be estimated from multiple views, single-view depth may be relevant when only one view of the scene is available. In addition, for deforming scenes, small camera motion or large illumination changes, multi-view reconstruction may present significant challenges and single-view depth be a reasonable alternative.  

Single-view depth may be applied in a wide range use cases such as augmented reality \citep{luo2020consistent}, computational photography \citep{Barron2015A} or medical robotics \citep{14500}. This thesis addresses this specific topic, with particular focus on endoscopic imaging, and our work contains several contributions and experimental evaluations in this field with focus on self-supervision and uncertainty quantification. 

\section{Depth perception}
Depth perception is a fundamental aspect of visual cognition. It refers to the ability to perceive the 3D spatial relationships of the objects within a scene, from the sole input of one or many 2D images. This cognitive capability is the basis for 3D reconstruction from a single or multiple views.
The importance of depth perception covers a wide range of industries and applications.

In immersive technologies, depth perception is a must. For example, in augmented reality, it enables virtual objects to be placed in the real world with appropriate depth and hence coherence with the rest of the viewed scene. In virtual reality, users can experience a virtual environment while the device perceives the real environment and prevents collisions. 
In robotics and self-driving cars, depth perception is a cornerstone to enable full autonomy. 3D reconstruction plays a crucial role in the understanding of a robot's surroundings, as autonomous robots should ground their behaviours on a comprehensive model of their environment, whether navigating indoors or exploring outdoor terrain. 3D models estimated from onboard sensor data provide robots with detailed spatial information, enabling among others collision-free navigation, object recognition and safe interaction with the environment.

When focusing on 3D reconstruction, the task of single-view depth estimation has its own unique challenges and importance. It is critical in scenarios where there is frequent deformation, small baseline between frames, or the camera is frequently occluded.
All these aspects are frequent in the medical arena, that has adopted vision technologies in many tasks. Nowadays, surgeons have the ability to visualise internal body structures in three dimensions, which allows for more precise planning and execution of surgical procedures. As another example, dentists also use 3D scanning to create accurate models for the design and manufacture of dental appliances.

Recovering depth from images has been studied extensively in the computer vision community. Multi-view, shape-from-X, and structure-from-motion are effective in recovering the three-dimensional structure of a scene using multiple images taken from different viewpoints. The estimation of depth from a single view is a potential complement to the many challenging conditions that may occur in multi-view setups, like low textured scenes, insufficient motion between views, deformations and illumination changes. While 3D reconstruction often leverages multiple views to recreate a scene or object in three dimensions, single-view depth estimation seeks to achieve a similar goal from only one perspective. 
However, it is an inherently complex problem due to its ambiguous nature. This ambiguity arises because multiple 3D configurations can result in the same 2D image, making it impossible from a general geometric perspective to identify the one that produce the image. This challenge has motivated considerable research into the use of additional cues or sophisticated algorithms to approximate depth more accurately.
Despite its complexity, mastering single-view depth estimation is critical for scenarios where multi-view or motion-based methods are impractical or impossible, e.g., in endoscopy. As depth estimation algorithms from single views mature in precision, they also offer potential for integration into multi-view 3D reconstruction pipelines \citep{Facil_2017}.

In this thesis, we tackle the problem of single-view depth from different perspectives, but always in a data-driven manner and deep neural networks. Firstly, we use data-driven methods and extract the uncertainty associated with depth maps. Secondly, we reduce the complexity of the problem by using illumination decline in the scenarios where the camera and illumination are co-located.

Deep neural network developments have swiftly progressed, becoming the predominant approach for various tasks within computer vision like semantic segmentation, classification object detection, or depth estimation. Specifically, CNNs \citep{lecun2015deep} can identify basic features in the early layers and more complex spatial associations in the deeper layers. CNNs apply a composition of convolutions to the input image, extracting hierarchically organized feature maps. The feature maps correspond to patterns in the input data such as corners, edges, textures, and in general local or mid-level patterns in the image.

A pioneering work in the recovery of 3D from 2D images was done by \citet{saxena2005}, who propose multi-spatial monocular clues that models the relative depths. \citet{eigen2014} were the first ones to demonstrate an effective use of deep learning in this problem. They propose a supervised learning pipeline using a convolutional neural networks with a coarse-to-fine architecture, being the input a RGB image and the target per-pixel depth estimations. 
By improving the network architecture, \citet{Laina2016} proposed deeper fully convolutional models and later \citet{fu2018deep} formulated the problem from an ordinal regression perspective in a discrete depth space.

Visual Transformers are the latest significant innovation in the field of deep learning. Based on self-attention mechanisms, visual transformers weigh the relevance of different image patches relative to each other. This mechanism helps to focus the attention on the most important areas, having a global context of the image. In contrary to the fundamentals of convolutional neural networks, visual patches are of fixed size and non-overlapping, that help to learn recognize patterns across different patches of the image. The architecture of transformers has been adapted to work with image patches visual transformers \citep{dosovitskiy2020image} and later applied to a single-view depth estimation \citep{Ranftl_2021_ICCV}.

Annotated depth data is very diverse depending of the captured scene (indoors or outdoors), the nature of annotation (be it relative or absolute), and the accuracy (from lasers, synthetic data, or Structure from Motion). \citet{ranftl2020towards} proposed the use of different source of depth using a scale and shifting invariant losses.
In the medical domain, single-view depth estimation has been extensively studied for endoscopic purposes. \citet{Visentini-Scarzanella2017} used computed tomography renderings for depth supervision in bronchoscopies. However, computed tomography scans in particular and ground-truth depth annotations in general are very rare in endoscopy, which makes self-supervision methods essential for a practical application of this technology.

\section{Self-supervised single-view depth}
Capturing large volumes of annotated data is resource-intensive, often requires domain expertise and specific hardware. In areas such as medical imaging, where expert annotation is both essential and limited, the challenge is particularly pronounced. In endoscopy, for example, the size of the endoscopes can make it challenging to obtain such annotations. Self-supervised methods are emerging as an alternative to traditional ground-truth supervision, especially given the vast difference between the amount of data with and without annotations, the last one being several orders of magnitude larger.

In the field of single-view depth, seminal contributions have been made by \citet{monodepth17}. They proposed a self-supervised learning method that enforces multi-view photometric consistency using a left-right stereo camera. Similarly, \citet{Zhou2017} explored the use of monocular ego-motion to enforce photometric consistency with a monocular image, where the motion is predicted by a network. \citet{Godard2018} introduced a pipeline for multi-scale self-supervision using monocular, stereo and combined cameras. However, this type of supervision can introduce noise due to various factors such as inaccuracies in camera motion estimation, perspective distortions, occlusions, and non-Lambertian effects.
In scenarios where obtaining real labeled data (or ground truth data) is challenging or it is not possible, there has been a tendency to generate synthetic data from simulators and apply learning techniques using synthetic data as reference. Even if the progress of realistic simulations is notable, the issue of domain change (from simulation to real) remains a persistent challenge. This domain change refers to the discrepancies between synthetic data generated by simulators and real-world data, which can hinder the performance of models trained only on simulated environments. 
The singularity of the medical domain has resulted in considerable research in the transition from synthetic to real-world scenarios. For example, the research by \citet{shen2019context} uses a conditional GAN for depth recovery, incorporating SLAM and multiview data.  \citet{karaoglu2021adversarial} focus on depth estimation from monocular images for broncoscopy navigation. The authors propose a domain adaptive pipeline in two steps: training a network in synthetic labels and use advesarial domain feature adaptation to enhance the performance on real images. In the work of \citet{chen2019slam}, a depth network is trained with synthetic images of a simple colon model and fine-tuned with domain-randomized photo-realistic images rendered from computed tomography scans.
In this thesis, we have proposed several contributions in the area of self-supervised learning for single-view depth. Firstly, in Chapter \ref{chap:2}, we evaluate how uncertainties affect the domain change scenario and propose an uncertainty aware teacher-student architecture that mitigates the effect of domain change. Secondly, in Chapter \ref{chap:3}, we introduce a new self-supervised learning technique that employs the information of the illumination decline as supervisory signal. 

\section{Single-view depth uncertainty}

The aforementioned research on image-based depth detection uses deterministic deep learning models. This means they always produce the same output for a same input without a measure of how certain they are about the prediction. During training, a certain loss function is minimized in order to find the set of parameters that will give the best performance in new data. Such models do not take into account the inherent uncertainties in their predictions, and this is relevant for their use in real-world scenarios. 
These uncertainties can occur due to the distribution of the data, the lack of data or the inherent limitations of the models. Uncertainty allows us to assess the reliability of predictions, which can be used to guide subsequent actions, particularly in fields such as medicine and autonomous driving, where decisions can have fatal consequences.
There are two types of uncertainty that can be considered in deep learning: aleatoric uncertainty and epistemic uncertainty \citep{KIUREGHIAN2009105}. Aleatoric uncertainty is the uncertainty of the observations, and it is inherent to the data, i.e. it would still exist even if more data is available. This uncertainty can be thought of as noise in the sensor . For example, in the case of single-view depth, when RGB-D cameras are used, the camera depth sensor noise is modelled by the aleatoric uncertainty. 
Aleatoric uncertainty is further subdivided into homoscedastic uncertainty, that is independent of the inputs (constant), and heteroscedastic uncertainty that is dependent on the inputs to the model. 
Epistemic uncertainty refers to the uncertainty of the model, and consequently, Bayesian deep learning is a natural way to capture epistemic uncertainty in neural networks. 
In Bayesian deep learning, the epistemic uncertainty can be obtained using a prior distribution $p(\mathbf{\theta})$ over the neural network parameters $\mathbf{\theta}$ and computing its posterior distribution $p(\theta|\mathcal{D})$ given a dataset $\mathcal{D}$ using Bayes rule: $p(\theta|\mathcal{D}) =\frac{p(\mathcal{D}|\theta)p(\theta)}{p(\mathcal{D})}$. This equation is intractable for deep learning architectures. However, there are scalable approaches to deep learning such as MC dropout and deep ensembles. 
In \citep{Kendall2017}, a seminal effort is made to highlight the significant types of uncertainty in deep learning, namely aleatoric and epistemic uncertainties.  \citet{Kendall2017} combine epistemic and aleatoric uncertainty by using a MC dropout approximation of the posterior distribution. Modelling aleatoric uncertainty as observational noise through a network output and epistemic uncertainty as the variations in the prediction by several forward passes.
Deep ensembles \citep{Lakshminarayanan2017} consist of training the same architecture multiple times, assuming different random initialisation of its parameters. Each ensemble is considered to be trained to a local minimum. The set of ensembles are an approximation of the model distribution.

For single-view self-supervised learning, \citet{poggi2020uncertainty} propose to extract depth uncertainty of depth from labels of a pretrained self-supervised network by introducing a teacher-student architecture.
It is important to note that uncertainty measures are of limited value unless they are calibrated with the corresponding errors.
In this thesis, we explore deeply MC dropout and deep ensembles in a supervised single-view depth setting, evaluating both their relative and absolute calibration of uncertainty.

\section{Endoscopic single-view depth estimation}
Colonoscopy and gastroscopy are among the most common procedures performed in hospitals. During an endoscopy, a physician uses an endoscope to examine a patient. This procedure involves navigating through the patient's body without damaging tissues and identifying and treating areas that represent a health risk.

Navigating environments such as the colon is a complex task that largely depends on the expertise of the doctor and the preparation of the patient. The primary tool aiding this procedure is the endoscope, with the camera at its tip serving as the unique sensor. This navigation relies heavily on a doctor prior anatomical knowledge and the real-time visual and mechanical feedback from the endoscope. Despite these aids, there remains a risk of perforation, emphasizing the need for enhanced depth perception to assist clinicians during navigation.

Concretely, single-view depth estimation may potentially revolutionize endoscopic mapping and navigation. This method emphasizes creating 3D maps of the inspected areas, allowing for a more detailed understanding of the internal environment \citep{Amit49144}. Given the predominance of monocular cameras in endoscopic procedures, deep learning algorithms present significant promise for advancing this technology. However, images from endoscopic procedures often contain challenges such as liquids, specular lighting, tissue deformations, drastic illumination changes, camera movements, and minimal parallax. Therefore, we argue in favor of systems based on single-view depth, which do not necessitate calculations of the camera relative position.

During screening procedures, physicians aim to identify polyps and tissue anomalies. Assessing these findings is crucial for timely and accurate intervention \citep{14500}. In this context, single-view depth can be instrumental in measuring polyps and in assessing areas affected by conditions like Crohn's disease.

Furthermore, the integration of single-view depth technology can significantly enhance robot-assisted, computer-aided interventions. By offering enhanced perception during navigation, single-view depth will benefit both robotic and manual procedures, potentially reducing associated risks to patients.

\section{Our contribution to single-view depth:}

The work described in this thesis presents significant improvements in the field of single-view depth estimation by using deep neural networks. The improvements encompass uncertainty quantification for single-view depth, a self-supervised approach based on a teacher-student architecture that models the teacher uncertainty, and a new single-view self-supervision method based on illumination decline.  

Previous research have shown that deep neural networks have significant potential to make depth estimations, which gives a promising basis to further develop their usage. They offer an appearance-based approach to the ill-posed problem of depth perception from 2D images. In addition, deep ensembles and MC dropout have shown to be a scalable alternative to Bayesian deep learning methods. 

The first of our contributions is the identification of deep ensembles as the best calibrated method for scalable Bayesian depth deep learning methods. We demonstrate empirically that adding dropout in all layers of the encoder delivers better results than other variations of MC dropout. This result is of practical relevance in the estimation of depth and uncertainty, because MC dropout requires much less memory than deep ensembles. It is particularly attractive for systems with limited resources due to its reduced memory requirements. As a result, MC dropout emerges as a potentially more adaptable approach when considering a Bayesian approach for real-world applications.
We also show the application of Bayesian depth networks in the context of 3D reconstruction by applying pseudo-RGB Depth Iterative Closest Point (ICP), showing that relative transformation can be improved by excluding the points with higher uncertainties.

Supervised methods outperform self-supervised methods in terms of accuracy, primarily because the benefit from training with accurate and annotated data. There are situations where acquiring labeled data is difficult, being an inclination to use synthetic data from simulators as a reference for learning. However, the domain shift problem is still a challenge because of the difference between synthetic and real images. 
In medical settings, especially in procedures like colonoscopies, the challenges presented by domain change become especially pronounced.

We pioneer the exploration of uncertainty quantification for single-view depth specifically focusing on medical colonoscopy images, marking a novel contribution to the existing literature.
Colonoscopy images are particularly difficult as they present many challenges. They have numerous discontinuities mainly due to partial occlusions in folds or haustra. The lighting conditions within the colon can vary significantly depending on the endoscope electronics, leading to frequent illumination changes. On top of that, the presence of specular light, which results from the reflection of light off the wet surfaces inside the colon, adds another layer of complexity to the image interpretation.
Due to this conditions, colonoscopy images are an ideal benchmark for evaluating the effectiveness and reliability of single-view depth estimation methods. 
The importance of understanding uncertainty in this domain is crucial, as inaccurate or misleading predictions could lead to misdiagnoses or oversight of critical medical conditions. Therefore, it is critical to have a robust system that not only estimates depth, but also provides a measure of the uncertainty associated with that estimation. Our main research advance is the introduction of a self-supervised method anchored in an uncertainty-aware teacher-student architecture. Our results demonstrate that our approach excels other types of learning in terms of depth and uncertainty quantification. Moreover, the results show the advantages of weakly supervised learning with respect to self-supervised learning.

Until now, single-view depth has traditionally been classified in two categories: supervised learning, which relies on depth annotations or annotations derived from a multi-view system; and self-supervised learning, which leverages geometric and/or photometric consistency across multiple views. We introduce a new form of depth supervision derived from the illumination decline principle. The main result of our work is a unique single-view self-supervised learning method for depth estimation. This method is applicable in dark environments, where the light source and camera are co-located. Such conditions are prevalent in numerous real-world scenarios, including within the human body as seen in clinical procedures which involve examining the interior of hollow organs or cavities, underwater domains like in submarines or specific seabed regions, enclosed natural formations such as caves, rigid enclosures often encountered during engine inspections or examinations of other mechanical parts, and confined cylindrical spaces like pipes or other similar structures that are typically hard to access.

For the development of this method, we grounded our research within a medical context, specifically focusing on endoscopes including procedures like colonoscopies and gastroscopies.

The medical field, especially when it comes to gastroscopy, is full of challenges reminiscent of those described for colonoscopy. Gastroscopy images often show foam, varying textures, numerous specular reflections and a variety of shapes that differ from what we observe in other endoscopic procedures, particularly when examining regions such as the stomach or other related areas.
Despite these complexities, this environment is a natural platform for our research.
The particularity lies in endoscopic imaging, where the imaging device is the only light source illuminating the scene.
The foundation of our approach is based on a simple yet profound observation: In endoscopic images, brighter areas tend to be closer to the camera, while darker areas tend to be further away. This insight differs significantly from conventional self-supervised learning paradigms. 
Our method is unique in that it requires only a single reference image during the learning process. Our technique not only outperforms other self-supervised methods, but also rivals depth-supervised strategies. In the scientific literature, self-supervised techniques have always lagged far behind supervised ones, being ours the first method that is competitive with ground truth supervision. 
An additional point in favour of our single-view self-supervision is its resilience to domain changes. Specifically, a distinct advantage of our method is the capability to refine the depth estimations on-the-fly at test time, which opens the way for improved predictions.

In addition to predicting depth, our technique is adept at assessing the albedo of the tissue. This advance is also significant, given the inherent value of this prediction mode. By predicting the albedo, our method can become instrumental in detecting flat lesions and texture variations, which are critical in identifying conditions such as Barrett's oesophagus in gastroscopy or Crohn's disease indications in colonoscopy. In particular, albedo predictions can improve the effectiveness of other machine learning systems. These diverse applications highlight the transformative potential of our method in medical diagnostics.

\section{List of publications}
Throughout this thesis, the following publications have been published (although the thesis document focuses in the first three): 

\begin{itemize}
    \item Rodríguez-Puigvert, J., Martínez-Cantín, R.,  Civera, J. (2022). \textbf{Bayesian deep neural networks for supervised learning of single-view depth.} IEEE Robotics and Automation Letters (R-AL), 7(2), 2565-2572. presented at  IEEE International Conference on Robotics and Automation (ICRA) 2022 
    \item Rodriguez-Puigvert, J., Recasens, D., Civera, J.,  Martinez-Cantin, R. (2022, September). \textbf{On the uncertain single-view depths in colonoscopies.} In International Conference on Medical Image Computing and Computer-Assisted Intervention (MICCAI) (pp. 130-140). Cham: Springer Nature Switzerland.
    \item Rodriguez-Puigvert, J., M. Battle, V., Montiel, J., Cantin, R., Fua, P., Tardos, J.,  Civera, J. (2023).\textbf{ LightDepth: Single-View Depth Self-Supervision from Illumination Decline.} In International Conference on Computer Vision (ICCV) 2023.
    \item Patent Pending: EP23382614.8. \textbf{SELF-SUPERVISED METHOD FOR OBTAINING DEPTH, ALBEDO AND SURFACE ORIENTATION ESTIMATES OF A SPACE ILLUMINATED BY A LIGHT SOURCE.}    
    \item Looper, S., Rodriguez-Puigvert, J., Siegwart, R., Cadena, C.,  Schmid, L. (2023, May). \textbf{3D vsg: Long-term semantic scene change prediction through 3D variable scene graphs.} In 2023 IEEE International Conference on Robotics and Automation (ICRA) (pp. 8179-8186). IEEE.
\end{itemize}
\section{Manuscript organization}

This thesis is organized as follows. In Chapter 2, we discuss about scalable Bayesian deep neural networks for supervised learning of single-view depth and we explore the importance of uncertainty quantification in robotics perception.
In Chapter 3, we discuss about uncertainty of deep ensembles applied to different types of learning for single-view depth from colonoscopy images. 
In Chapter 4, we discuss the advantages of using a physical-model light-camera based to introduce the first single-view self-supervised method for depth learning. Finally, in Chapter 5, we summarize the conclusions of the thesis work.
\chapter{Bayesian deep neural networks for supervised learning of single-View depth}
\label{chap:1}
Understanding and quantifying uncertainty is crucial when using deep neural networks in real-world scenarios. The aim of this chapter is to explain how uncertainty can be quantified for deep neural networks and to evaluate the calibration of the captured uncertainty.
We distinguish between two forms of uncertainty: aleatoric and epistemic. Aleatoric uncertainty arises from the inherent randomness or variability in the data, and represents the intrinsic noise of the system. Epistemic uncertainty, on the other hand, arises from either insufficient knowledge or a lack of data. However, as more data are collected and models refined, epistemic uncertainty can be reduced or even eliminated. 
To investigate these uncertainties, our approach focuses on modelling the aleatoric directly in the network and capturing the epistemic using two scalable techniques: MC dropout and deep ensembles.
With respect to MC dropout, we explore the importance of dropout layers within the network architecture. Deep ensembles are a form of Bayesian model averaging, and in practice they can be used as an approximation of the full posterior. 
\section{Introduction}
\begin{figure}
    \centering
    \includegraphics[width=0.5\textwidth]{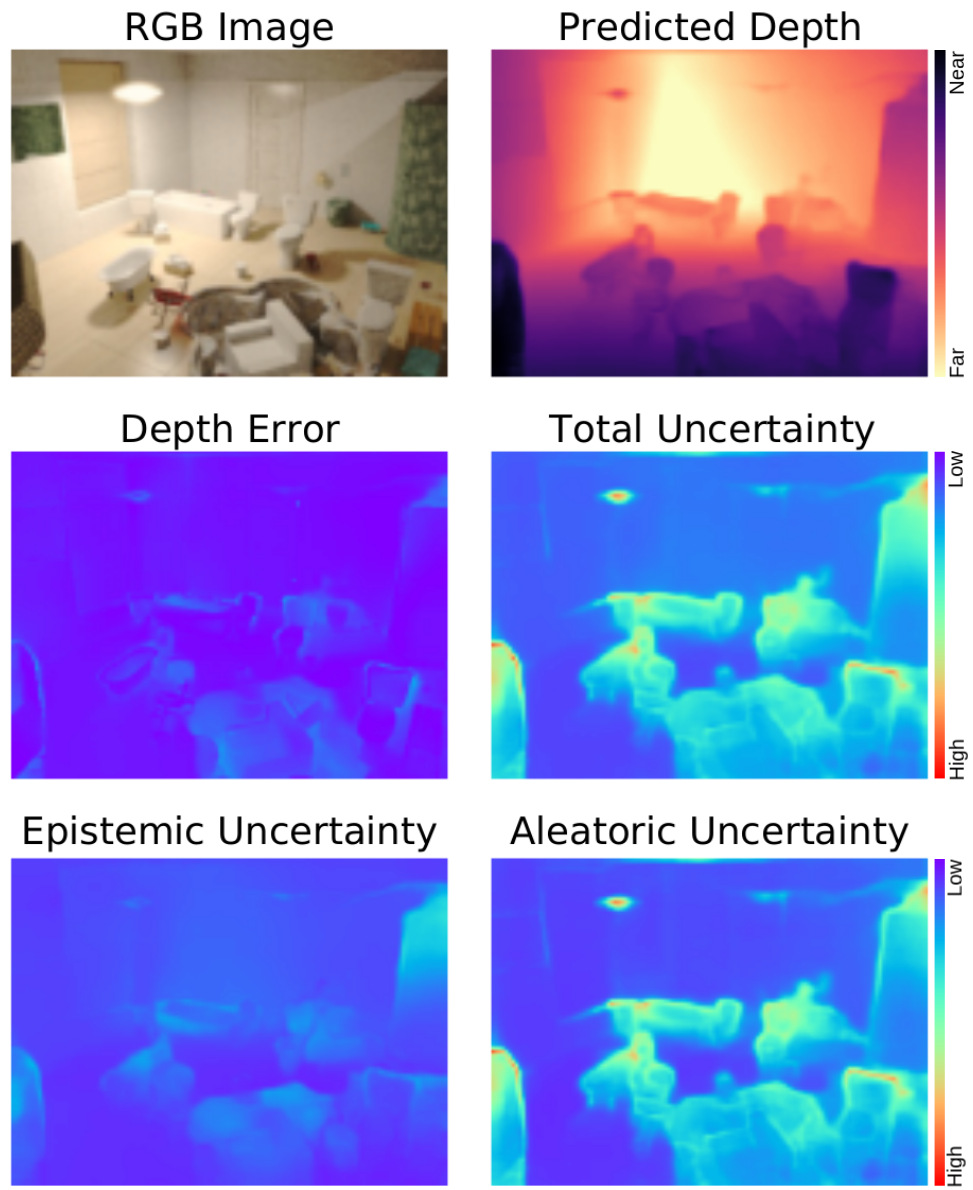}
    \caption[Bayesian single-view depth prediction for a SceneNet image]{Bayesian single-view depth predicton for a SceneNet image.
    In the middle row the small depth error, and how the total uncertainty models it accurately. In the bottom row how epistemic and aleatoric sources are both significant and relevant for uncertainty quantification.}
    \label{fig:teaserRAL}
\end{figure}
The quantification of the uncertainty is critical in robotics, in order to implement systems that are robust and reliable in real-world applications. Point estimators, which dominate the landscape of multi-view~\citep{engel2017direct} and single-view~\citep{fu2018deep,Godard2018} scene reconstruction, do not typically compute the full distribution but just the maximum-likelihood state. Higher-level decision blocks have no means to judge how accurate these estimates are, and hence its use for safe planning might be questionable. Uncertainty is often present in the formulation of model-based estimators (e.g., \citep{barfoot2017state}), but much less in the training of deep learning models. Furthermore, learning-based approaches tend to overfit on standard datasets, which might lead us to assume a reasonable general performance while they are strongly biased. In such cases, their outputs should not be trusted in real-world applications, for which we would need generalizable and self-aware models. Bayesian learning is one of the approaches that can address such application challenges. In a supervised learning approach using depth annotations, we investigate epistemic and aleatoric uncertainties to determine the accuracy of the model's predictions and its suitability for use in robotic environments. In neural networks, uncertainty can stem from the input data or the network weights. For the latter, scalable approaches to Bayesian deep learning have shown to be effective to model uncertainty \citep{Gustafsson2019}.
Figure~\ref{fig:teaserRAL} shows the output of our Bayesian framework for supervised single-view depth learning. The small depth error is noticeable, but more importantly, the fact that this error is mostly coherent with the predicted uncertainty. Note also how the aleatoric uncertainty is greater in this case, but the epistemic uncertainty is still relevant for an accurate quantification. 
In this chapter, we provide a unified framework and a thorough evaluation of scalable uncertainty estimation methods, namely Monte Carlo (MC) dropout and deep ensembles, for supervised single-view depth learning with deep convolutional networks. We propose to apply MC dropout in the encoder, contrary to recent works~\citep{Gustafsson2019,poggi2020uncertainty} that apply it in the decoder. We demonstrate in our evaluations that, in the particular task of single-view depth supervised learning, the dropout in the encoder achieves a better performance than the dropout in the decoder.
Such a result has relevant practical implications, as MC dropout has a lower memory footprint than deep ensembles. Finally, we also provide results in pseudo-RGBD ICP, a potential application for our single-view depth uncertainty models. In our experiments, we demonstrate that our uncertainty estimates are reasonably well calibrated and has significant potential to provide accurate and scaled motion estimates from monocular views.

\section{Background and related work}
\subsection{Structure estimation and learning from images}

Reconstructing a 3D scene from visual data has been addressed from a wide variety of perspectives, the more typical being based on multiple images either alone~\citep{mur2017orb,engel2017direct} or fused with other sensors (e.g., visual-inertial setups~\citep{qin2018vins}). However, multi-view and visual-inertial pipelines present two limitations: they require sufficiently textured scenes to find correspondences, and also sufficient motion for observability. Single-view depth estimation can help with these two issues, although it is significantly more challenging due to its ill-posed nature.

Following the seminal work of \citet{saxena2005} on single-view depth, \citet{eigen2014} were the first ones that used depth supervision to train a deep network for such task. 
Many works followed with different contributions: \citet{Laina2016} proposed deeper fully convolutional models. \citet{fu2018deep} proposed a spacing-increasing depth discretization that learns depth from an ordinal regression perspective. \citet{Dijk_2019_ICCV} evaluated a self-trained networks to investigate which visual cues they use. From their conclusions, depth networks favor vertical positions and disregard obstacles in their apparent size.
Similarly, we contribute to understanding the behavior of depth networks from a Bayesian perspective.  

Several works have proposed self-supervised approaches, using photometric reprojection losses between stereo or multiple views~\citep{monodepth17,Zhou2017}. Self-supervised approaches still underperform compared to supervised ones. For this reason, we focus on supervised methods.

\subsection{Bayesian deep learning}
\label{sec:bayesdl}
Bayesian deep learning combines the strengths of deep neural architectures with the uncertainty quantification of probabilistic (Bayesian) learning and inference methods. Regarding uncertainty, we must differentiate between what the model does not know and what is missing from the input data. Accordingly, the uncertainty sources can be classified into two: aleatoric and epistemic. 
Aleatoric (also referred to as statistical) uncertainty, refers to the variations caused by the realization of different experiments with stochastic components. In our models, it encodes the variability in the different inputs from the test data and hence cannot be reduced by increasing the amount of training data. Some models assume that the aleatoric uncertainty is homoscedastic; that is, it is independent of the input data. In this chapter, we train the network to predict the uncertainty for each input datum resulting in a heteroscedastic uncertainty model~\citep{Kendall2017, Kiureghia}.
Epistemic (also known as systematic) uncertainty, represents the lack of knowledge of a trained model. This type of uncertainty is deeply related to the training data and the model ability to generalize. For example, epistemic uncertainty is high for out-of-distribution data or extrapolation in regions where training data was scarce. In Bayesian deep learning, the epistemic uncertainty can be estimated from the uncertainty in the model parameters, assuming that the model architecture is correct. In this case, epistemic uncertainty can be obtained using a prior distribution $p(\mathbf{\theta})$ over the neural network parameters $\mathbf{\theta}$ and computing its posterior distribution $p(\theta|\mathcal{D})$ given a dataset $\mathcal{D}$ using Bayes rule: $p(\theta|\mathcal{D}) =\frac{p(\mathcal{D}|\theta)p(\theta)}{p(\mathcal{D})}$. In general, this equation is intractable for state-of-the-art deep architectures, but there are several approaches to tackle this problem that we describe below. 

\subsubsection{Variational inference (VI).} VI proposes the use of a tractable approximation $q(\theta)$ to the posterior distribution $p(\theta|\mathcal{D})$. Mean-field variational inference assumes an isotropic Gaussian distribution for $q(\theta)\sim \mathcal{N}(\theta|\mu,\mathbf{I}\sigma)$. The parameters of the approximate distribution $q(\theta)$ are optimized by minimizing the KL-divergence between the approximate distribution and the true posterior $D(q||p)$. Mean-field variational inference with Gaussian approximation suffers from the soap-bubble effect, reducing the predictive performance as most samples fall in a ring. The Radial Bayesian Neural Networks \citep{farquhar_radial_2020} avoid that effect, but the distribution is biased towards the center resulting in uncertainty underestimation. Furthermore, VI methods are very sensitive to calibration and configuration. A natural-gradient VI method \citep{osawa2019practical} was introduced to improve the robustness of the optimization. However, it requires strong approximations of the Hessian, resulting in lower performance.

\subsubsection{Monte Carlo (MC) dropout.} MC dropout can be used to approximate the posterior distribution, as proposed by \citet{Gal2016}. It can be considered as a specific case of VI, where the variational distribution includes a set of binary random variables that represent the corresponding unit to be turned off or dropped. The approximation makes the computation tractable and robust. MC dropout is able to approximate multimodal distributions. However, the epistemic distribution on the weight-space only has discrete support. \citet{Kendall2017} presented a framework to combine both aleatoric and epistemic uncertainty, where MC dropout is used to obtain epistemic uncertainty, while the function mapping the aleatoric uncertainty is learned from the input data. 


\subsubsection{Deep ensembles.} Deep ensembles \citep{Lakshminarayanan2017} involves training the same architecture many times optimizing some MAP loss, but starting from different random initialization of its parameters. Therefore, deep ensembles are not truly a Bayesian approach as the samples are distributed according to the different local optima. Conversely, these models in an ensemble perform reasonably well, even considering the small number of random samples considered in practice, as all of the models are optimized and have a high likelihood. Therefore, deep ensembles can be considered an approximate Bayesian model average, although, in practice, they can also be used as a rough posterior approximation. Contrary to MC dropout, where the model weights are shared between samples, in the case of deep ensembles, each \emph{sample} is trained independently. Therefore the number of model parameters required grows linearly with the number of samples. Furthermore, deep ensembles also result in a distribution with discrete support on the weight-space.  

\subsection{Bayesian deep learning in computer vision}

Evaluating uncertainty correctly is still in open discussion, as it is task-related. \citet{Mukhoti2018} evaluated MC dropout for semantic segmentation and designed the metrics for such case. Similarly, \citet{Gustafsson2019} designed a framework to explore uncertainty metrics for semantic segmentation and depth completion, using MC dropout and deep ensembles. 
\citet{ilg2018uncertainty} compare different strategies and techniques for quantifying the uncertainty of optical flow. They also introduce a multi-hypothesis network based on a winner-takes-all loss function that penalizes the best hypothesis result. However, disentangling aleatoric and epistemic uncertainty could be an arduous task in multi-hypothesis approaches. The network that merge all hypotheses contains its own epistemic uncertainty that is not taking into account. Nevertheless, they show to be competitive in comparison to deep ensemble and MC dropout. For depth estimation,  \citet{yang2019inferring} proposes a multinomial distribution to learn uncertainty based on discretizing the depth space.

For our case of single-view depth regression, \citet{poggi2020uncertainty} evaluated the uncertainty in self-supervised learning, which leverage photometric consistency between views. They observed that depth accuracy is improved by uncertainty estimation along the training paradigms. This thesis complements the ones mentioned in this section by evaluating uncertainty quantification in a supervised regression setting. 

\section{Bayesian single-view depth learning from supervised data}
MC dropout and deep ensembles provide a sample representation of the posterior distribution over the network parameters. Here, we introduce a unified formulation to analyze the posterior and predictive distribution for these sample representations. We have particularized our framework for depth perception applications, although it could be extended to other tasks.

\subsection{Architecture and loss}

We adapt a U-Net \citep{Ronneberger2015} encoder-decoder architecture as in  \citep{Godard2018,poggi2020uncertainty}. Our encoder is a Resnet18 \citep{He_2016_CVPR} pre-trained in ImageNet~\citep{Russakovsky2014}. Table \ref{tab:decoder} summarizes our decoder architecture. 

\begin{table}[h!]
\centering
\scriptsize
\begin{tabular}{ |l|l|l|l|l| }
\hline
\multicolumn{4}{ |c| }{Depth Decoder} \\
\hline
\textbf{layer} & \textbf{\# filters} & \textbf{inputs} & \textbf{activation} \\ \hline
upconv5 & 256 & econv5 & ELU\\
iconv5 & 256 & ↑upconv5, econv4 & ELU \\\hline
upconv4 & 128 & iconv5 & ELU \\ 
 iconv4 & 128 & ↑upconv4, econv3 & ELU \\
depth\_unc4 & 2 &  iconv4 & - \\ \hline
upconv3 & 64 & iconv4 & ELU\\
iconv3 & 64 & ↑upconv4, econv3 & ELU \\ 
depth\_unc3 & 2  &  iconv3 & - \\\hline
upconv2 & 32 & iconv3 &ELU\\
iconv2 & 32& ↑upconv2, econv1  &  ELU\\
depth\_unc2 & 2 & iconv2 & - \\\hline
upconv1 & 16 & iconv2 & ELU\\
iconv1 & 16 & ↑upconv1 & ELU\\
depth\_unc1 & 2 & iconv1& -\\
\hline
\end{tabular}
    \caption[Details of the decoder architecture.]{Decoder architecture. Kernels are always $3\times 3$ with stride $1$. ↑ stands for $2 \times 2$ nearest-neighbor upsampling.}
    \label{tab:decoder}
\end{table}

Our training data $\mathcal{D}=\{ \{I_1,d_1 \},\hdots,\{I_N,d_N \}\}$ is composed by $N$ supervised pairs, each pair $i \in \{1,\hdots,N\}$ containing a RGB image $I_i \in \{0,\hdots,255\}^{w \times h \times 3}$ and its ground truth depth $d_i \in \mathbb{R}^{w \times h}_{>0}$. For a single input image, the network $f_\theta(I)$ outputs two channels: per-pixel depth $\widehat{d}(I)$ and uncertainty $\sigma_d(I)$. The later corresponds to aleatoric uncertainty, which can also be interpreted as heteroscedastic observation noise. 
We incorporate both output channels in a single loss per image by using a standard Laplace log-likelihood \citep{Kendall2017}:

\begin{equation}
    \Lb(\theta) = \frac{1}{w \cdot h} \sum_{j \in \Omega} \frac{||d[j] - \widehat{d}[j]||}{ \sigma_{d}[j]} + \log \sigma_{d}[j]
\end{equation}
where $j \in \Omega$ is the pixel index in the image domain $\Omega$.

For deep ensembles, the loss function is evaluated independently for each sample model as they are trained separately, resulting in $M$ sets of parameters $\{\theta_m\}_{m=1}^M$. Although the sample models are not drawn from the posterior distribution, it still can be considered an approximation in practice. Deep ensembles are especially suitable for our problem as we need to maintain the number of samples small to keep it tractable. Therefore, it is important that we are not wasting valuable resources in low probability models that might reduce the overall performance.
In the case of MC dropout, the loss function can be used for approximate variational inference on the posterior distribution of the weights by training with dropout {after} every layer. The actual Monte Carlo phase is done by also performing random dropout at test time to sample from the variational distribution computed during training \citep{Kendall2017}. This sampling at test time results again in a set of $\{\theta_m\}_{m=1}^M$ different parameters. This time they are all generated from the same trained model, resulting in a much lower memory and computational footprint compared to deep ensembles or other variational methods. 

In practice, we found that adding dropout at every layer reduced the predictive performance considerably for our application, which is consistent with previous results \citep{Mukhoti2018}. Therefore, in section \ref{sec:results-bayesdepth}, we study different configurations of dropout and compare their quality both in terms of depth error and uncertainty quantification.
\subsection{Bayesian prediction of sample-based deep networks}
The predictive distribution for a pixel depth can be computed by integrating over the model parameters. We use the same strategy for MC dropout and deep ensembles, as they both use sample representations of the model parameters: 
\begin{equation}
    \begin{split}
        p(d | I, \mathcal{D} ) &= \int p(d | I, \theta) p(\theta|\mathcal{D}) d\theta\\
        & \approx \sum_{m=0}^M p(d | I, \theta_m)
    \end{split}
\end{equation}

\begin{figure} 
    \centering
    \centering
    \includegraphics[width=\textwidth]{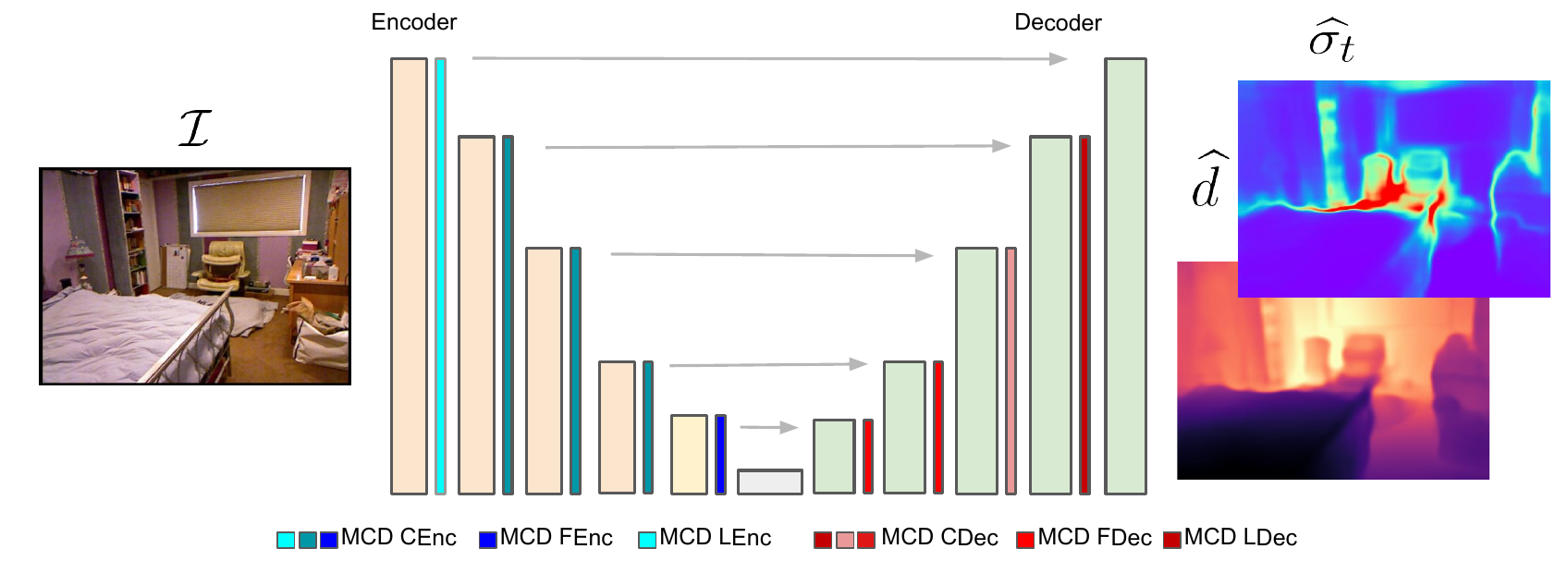}
    \caption{Variations of MC dropout in our experiments. }
    \label{fig:arch}
\end{figure}
As our architecture generates a Gaussian prediction for the pixel depth $\mathcal{N}(\widehat{d},\sigma^2_d)$, the sample-based output is a mixture of Gaussians that can be approximated by a single Gaussian. In particular, for the $\{\theta_m\}_{m=1}^M$ model samples (MC dropout) or models (deep ensembles) with respective outputs $\widehat{d}(m)$ and $\sigma_d(m)$, we approximate the total predictive distribution per pixel as a Gaussian $p(d | I, \mathcal{D}) \approx \mathcal{N}(\widehat{d}_t,\sigma^2_t)$ with:
\begin{equation}
    \begin{split}
        \widehat{d}_t &= \frac{1}{M} \sum_{m=1}^{M} \widehat{d}_m\\
        \sigma^2_t &= \underbrace{\frac{1}{M} \sum_{m=1}^{M} \left(\widehat{d}_t - \widehat{d}(m)\right)^2}_{\text{epistemic}} + \underbrace{\frac{1}{M} \sum_{m=1}^{M} \sigma^2_d(m)}_{\text{aleatoric}}
    \end{split}
\end{equation}

In the experiments section, we will show that identifying and quantifying the epistemic from the aleatoric uncertainty will be fundamental to finding the uncertainty source and improving the quality of the model and the predictions.

\section{Experiments}
\subsection{Datasets} 
\textbf{SceneNet RGB-D dataset~\citep{McCormac2016}} contains photorealistic sequences of synthetic indoor scenes from general camera trajectories, along with their ground truth. Our models are trained over $210,000$ synthetic images of $700$ scenes and tested on $90,000$ images of $300$ different scenes. We chose this dataset as it provides a wide variety of viewpoints and scenes, challenging occlusions and different lighting conditions, which are relevant for the network generalization.

\textbf{NYU Depth V2 \citep{Silberman:ECCV12}} consists of $120,000$ RGB-D images in $464$ indoor scenes. For training, we use $36,253$ images of $249$ scenes as proposed by  Lapdepth \citep{Lapdepth}. We test our models in the official split of $654$ images \citep{eigen2014}. 

\subsection{Metrics}

We use the depth error metrics that are standard in literature: Absolute Relative difference: Normalize per pixel error according to the real depth, reducing the effect of large error with the distance Eq. \ref{absrel}, Square Relative difference: penalize larger square errors, Root mean square error (RMSE), Root mean square error Log ($RMSE_{Log}$) and $\delta < 1.25^i$ with $i \in \{1,2,3\}$ \citep{eigen2014}: 
\begin{equation}
\label{absrel}
   Abs Rel = \frac{1}{w \cdot h} \sum_{\boldsymbol{j} \in \Omega_{i}}\frac{|d[j]-\widehat{d}[j]|}{\widehat{d}[j]}
\end{equation}
\begin{equation}
Sq Rel = \frac{1}{w \cdot h} \sum_{\boldsymbol{j} \in \Omega_{i}}\frac{(d[j]-\widehat{d}[j])^2}{\widehat{d}[j]}
\end{equation}
\begin{equation}
    RMSE = \frac{1}{w \cdot h} \sum_{\boldsymbol{j} \in \Omega_{i}}  (d[j]-\widehat{d}[j])^2 )^{1/2}
\end{equation}
\begin{equation}
    RMSE_{Log} =(\frac{1}{w \cdot h} \sum_{\boldsymbol{j} \in \Omega_{i}}  (\log d[j]- \log \widehat{d}[j])^2 )^{1/2}
\end{equation} 
\begin{equation}
     \delta < 1.25^i =  \frac{1}{w \cdot h} \sum_{\boldsymbol{j} \in \Omega_{i}} \max(\frac{d[j]}{\widehat{d}[j]},\frac{\widehat{d}[j]}{d[j]}) < 1.25^i
\end{equation}

For uncertainty, we use the Area Under the Calibration Error curve (AUCE) and Area Under the Sparsification Error curve (AUSE)  \citep{Gustafsson2019}. Since our methods output a Gaussian distribution $ \mathcal{N}(\widehat{d}, \sigma^2)$ per pixel, we generate prediction intervals ~$\widehat{d} \pm \phi^{-1} (\frac{p+1}{2})\sigma$ of confidence level $p \in [0, 1]$ being $\phi$ the CDF of the standard normal distribution. In a perfectly calibrated model, the proportion of pixels for which the prediction intervals covers the ground truth coincides the confidence level. AUCE is an absolute uncertainty metric, we use AUSE, in terms of RMSE, as relative measure of uncertainty. This metric compares the ordering of the per-pixel uncertainties against the order of the per-pixel depth errors. The ordering should be similar for a well-calibrated uncertainty, as uncertain predictions will tend to have larger errors.

For the pseudo-RGBD Bayesian ICP in Section \ref{sec:pseudorgbdicp}, we report the translational and rotational RMSE. 

\subsection{Bayesian single-view depth}
\label{sec:results-bayesdepth}

We evaluate several variations of MC dropout and deep ensembles. Specifically, for MC dropout, we report depth and uncertainty metrics for dropouts at different layers and with $p = 0.3$ and $p = 0.5$, $p$ being the probability of an element to be zeroed.

\subsubsection{MC dropout.} We consider seven variations (see Figure \ref{fig:arch} for a summary plot):
In the decoder, dropout after every convolutional layer, (\textbf{MC D}ropout \textbf{C}omplete \textbf{Dec}oder), the first two convolutional layers (\textbf{MC D}ropout \textbf{F}irst \textbf{Dec}oder), and the last convolution layer of the decoder (\textbf{MC D}ropout \textbf{L}ast \textbf{Dec}oder). And, in the encoder, dropout after every convolutional layer (\textbf{MC D}ropout \textbf{C}omplete \textbf{Enc}oder), the first convolutional layer (\textbf{MC D}ropout \textbf{F}irst \textbf{Enc}oder), and after the last convolution layer (\textbf{MC D}ropout \textbf{L}ast \textbf{Enc}oder). Finally, in NYU RGB-D v2 we also evaluate the effect of the dropout in all layers of the architecture (\textbf{MC D}ropout \textbf{C}omplete \textbf{Enc}oder-\textbf{C}omplete \textbf{Dec}oder). 

\subsubsection{Deep ensembles.} We examine uncertainty and depth by averaging an ensemble composed of a variable number of networks. We initialise the network weights with different seeds from a normal distribution $\mathcal{N}(0,10^{-2})$.

\subsubsection{Results.}
\begin{figure*}
    \centering
    \includegraphics[width=\textwidth]{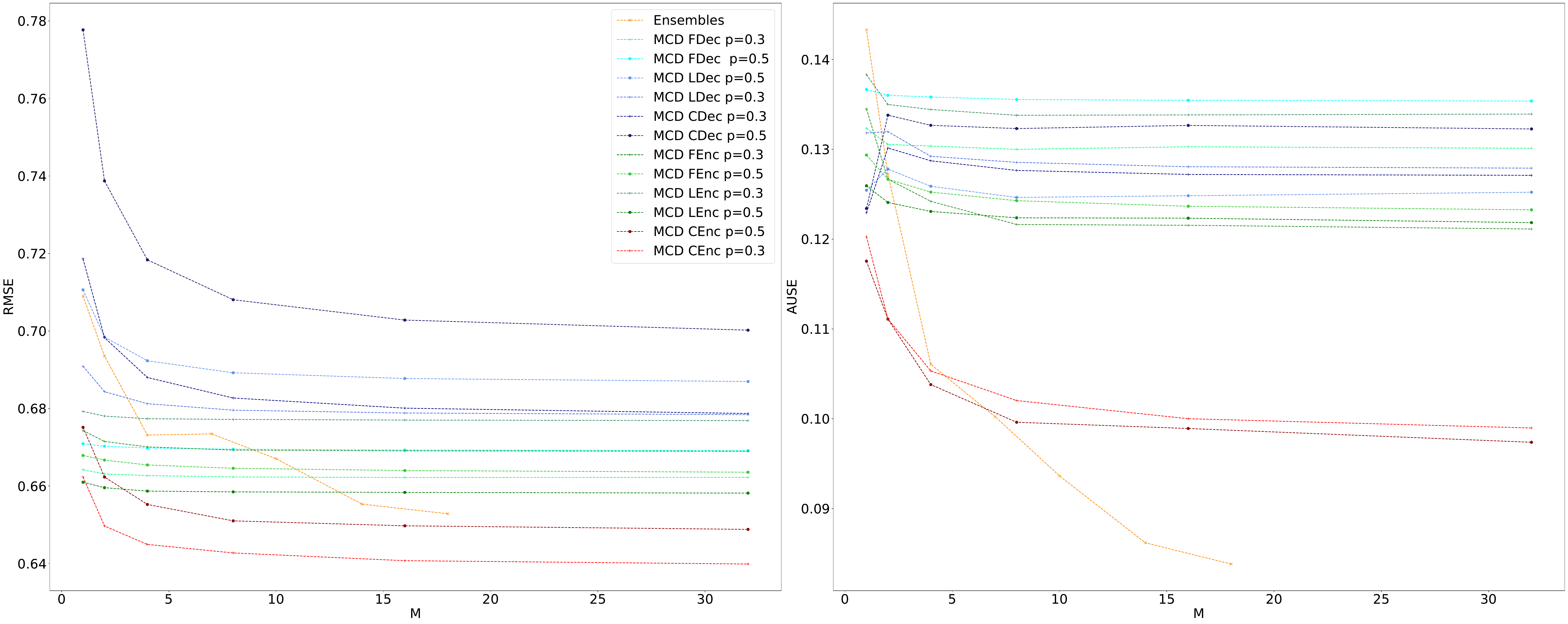}
    \caption[Comparison of MC dropout variations and deep ensembles for different numbers of forward passes.]{Comparison of MC dropout variations and deep ensembles for different numbers of forward passes $M$. Left: RMSE. Right: AUSE.  The higher $M$ is, the better the performance, but with slight improvements for $M>18$.} 
    \label{fig:results}
\end{figure*}
\begin{table*}
\scriptsize
    \centering
    \begin{tabular}{|c|c|c|c|c|c|c|c|c|c|}
    \hline
        Model & M & Abs Rel  & Sq Rel & RMSE & RMSE Log & $\delta< 1.25$ & $\delta < 1.25^2$ & $\delta < 1.25^3$ &  AUSE\\\hline
MCD CEnc p=0.3 & 64 & \textbf{0.1231} & \textbf{0.1222} & \textbf{0.6396} & \textbf{0.1982} & \textbf{0.8719} & \textbf{0.9635} & \textbf{0.9850} & 0.0985\\\hline
MCD CEnc p=0.5 & 64 & \underline{0.1243} & \underline{0.1228} & \underline{0.6484} & 0.2023 & \underline{0.8675} & \underline{0.9615} & 0.9842 &  \underline{0.0968}\\\hline
        MCD FEnc p=0.3  & 64 & 0.1291 & 0.1326 & 0.6688 & 0.2047 & 0.8591 & 0.9591 & 0.9834  & 0.1210\\\hline
MCD FEnc p=0.5 & 64 & 0.1320 & 0.1363 & 0.6633 & 0.2062 & 0.8583 & 0.9586 & 0.9833 & 0.1229\\\hline
MCD LEnc p=0.5 & 64 & 0.1244 & 0.1245 & 0.6582 & 0.2010 & 0.8658 & 0.9607 & 0.9846  & 0.1219\\\hline
MCD LEnc p=0.3 & 64 & 0.1244 & 0.1248 & 0.6769 & 0.2000 & 0.8629 & 0.9595 & 0.9848  & 0.1327\\\hline
MCD CDec p=0.3 & 64 & 0.1316 & 0.1322 & 0.6781 & 0.2044 & 0.8567 & 0.9597 & 0.9841 & 0.1268\\\hline
MCD CDec p=0.5 & 64 & 0.1369 & 0.1378 & 0.6988 & 0.2080 & 0.8494 & 0.9579 & 0.9835 & 0.1323\\\hline
MCD FDec p=0.5 & 64 & 0.1263 & 0.1252 & 0.6690 & 0.2035 & 0.8604 & 0.9593 & 0.9841  & 0.1353\\\hline
MCD FDec p=0.3 & 64 & 0.1264 & 0.1263 & 0.6622 & 0.2016 & 0.8641 & 0.9604 & 0.9842  & 0.1304\\\hline
MCD LDec p=0.5 & 64 & 0.1336 & 0.1371 & 0.6866 & 0.2065 & 0.8568 & 0.9588 & 0.9833  & 0.1241\\\hline
MCD LDec p=0.3 & 64 & 0.1293 & 0.1303 & 0.6782 & 0.2042 & 0.8589 & 0.9593 & 0.9837  & 0.1285\\\hline
Deep ensembles & 18 & 0.1283 & 0.1244 & 0.6529 & \underline{0.1993} & 0.8617 & 0.9613 & \textbf{0.9850} & \textbf{0.0838}\\\hline

    \end{tabular}
    \caption[Depth and uncertainty metrics for MC Dropout and Deep ensembles in SceneNet dataset.]{Depth and uncertainty metrics for several variations of MC Dropout and Deep ensembles in SceneNet RGB-Depth. Best results are boldfaced, second best ones are underlined.}
    \label{tab:metrics}
\end{table*}
Table \ref{tab:metrics} shows the depth and uncertainty metrics for the MC dropout variations and deep ensembles  on SceneNet RGB-D.
We observe that the uncertainty metric for deep ensembles outperforms all variants of MC dropout. 
This is due to the fact that deep ensembles are optimized to be close to a minimum. 
However, the depth error metrics are consistently better for MCD CEnc, which also show the second best uncertainty metrics.
    
The performance of the different MC dropout models varies significantly, which is a novel result of our analysis. We found that introducing dropout in all layers of the encoder (MCD CEnc) improves the results with respect to applying it to a few layers of the encoder (MCD FEnc, LEnc) or to the decoder (MCD FDec, LDec and CDec). Again, this result is of relevance as MC dropout is commonly applied in the decoder for depth estimation \citep{Gustafsson2019,poggi2020uncertainty}. Our rationale for this result is as follows: we believe that applying MC dropout only in the decoder makes the network learn deterministic representations (the encoder is deterministic). In contrast, applying MC dropout in the image encoder allows us to learn probabilistic representations modeling uncertainty in the feature space. This seems to be a more appropriate choice for Bayesian image processing.

Our experiments show that there is small variations in the MC samples when we solely apply dropout close to the code, as done in MCD FDec or MCD Lenc. This results in worse calibrated models and also less satisfactory depth estimations. We also observed that applying MC dropout after the first and before the last layers of the network leads to poor performance (see MCD LDec, MCD FEnc).

Comparing the two dropout probabilities, $p = 0.3$ and $p = 0.5$, we observe that the depth prediction is usually better for $p = 0.3$ but the uncertainty is better calibrated for $p = 0.5$, as it introduces greater variability. One or the other should be preferred depending on the application. 

MC dropout has a much smaller memory footprint than deep ensembles and our results indicate that the MCD CEnc performs similarly to ensembles, so it could be relevant option in certain applications. 
Specifically, only $56$ Mb are required for our MC dropout models, while the memory for deep ensembles grows linearly with the number of samples (for $M=18$, around $1$ Gb). The run time grows linearly with the number of samples in both cases (around 3--4 ms per forward pass). All data stems from a NVIDIA GeForce RTX 3090 GPU.

\begin{figure}
\centering
\begin{subfigure}{.49\columnwidth}
  \includegraphics[width=\linewidth]{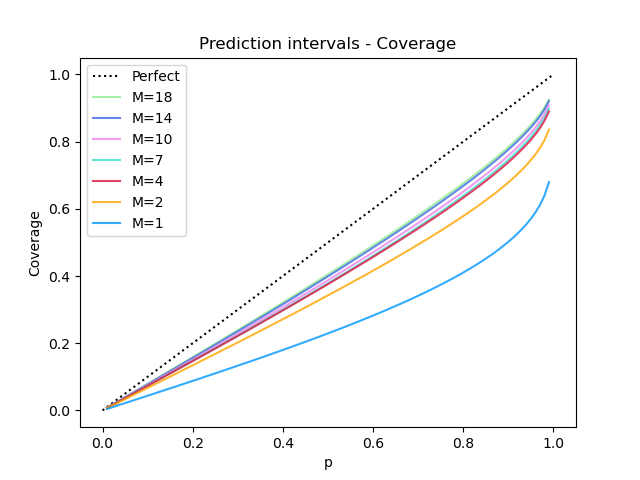}
  \caption{\scriptsize AUCE Deep ensembles}
\end{subfigure}
\begin{subfigure}{.49\columnwidth}
  \includegraphics[width=\linewidth]{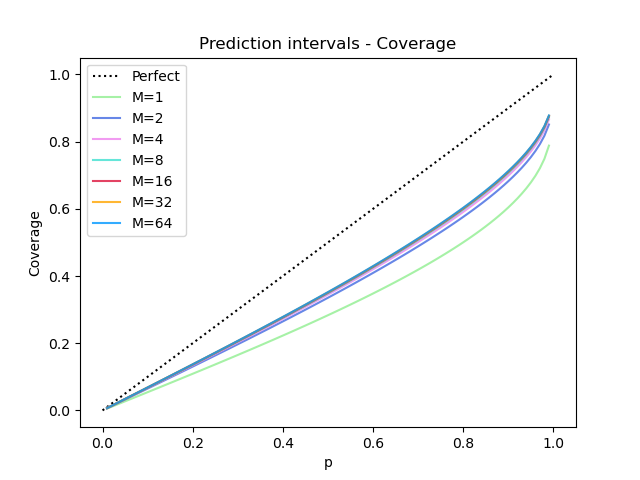}
  \caption{\scriptsize AUCE MCD CEnc p=0.5}
\end{subfigure}
\begin{subfigure}{.49\columnwidth}
  \includegraphics[width=\linewidth]{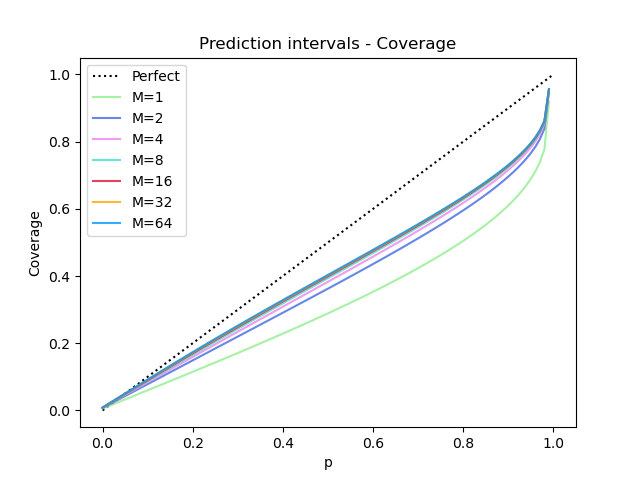}
  \caption{\scriptsize AUCE MCD CDec p = 0.3}
\end{subfigure}
\begin{subfigure}{.49\columnwidth}
  \includegraphics[width=\linewidth]{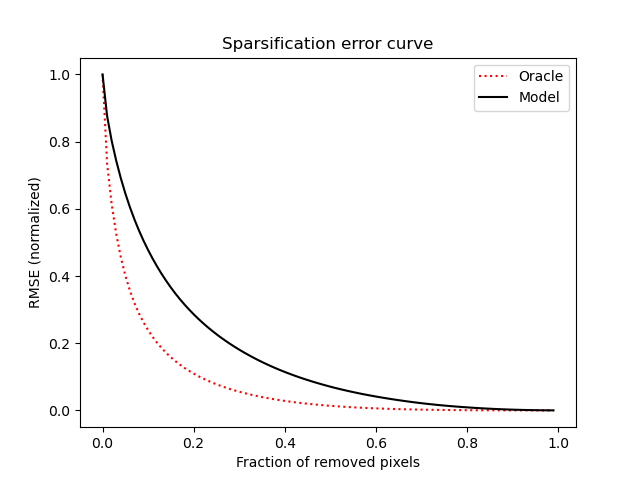}
  \caption{\scriptsize AUSE Deep ensembles M=18}
\end{subfigure}
\begin{subfigure}{.49\columnwidth}
  \includegraphics[width=\linewidth]{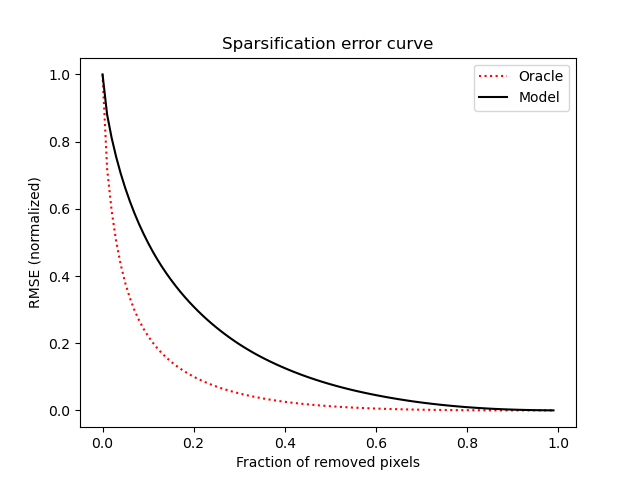}
  \caption{\scriptsize AUSE MCD CEnc p=0.5 M=64}
\end{subfigure}
\begin{subfigure}{.49\columnwidth}
  \includegraphics[width=\linewidth]{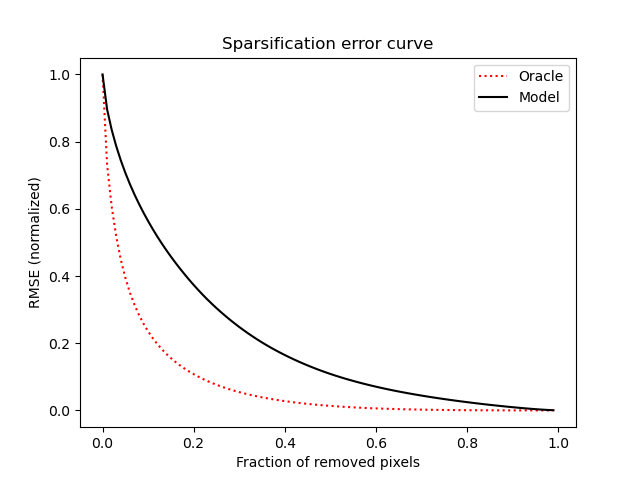}
  \caption{\scriptsize AUSE MCD CDec p=0.3 M=64}
\end{subfigure}
\caption{Calibration curves (AUCE and AUSE) for MC dropout and deep ensembles.}
\label{fig:u_metrics}
\end{figure}

Figure \ref{fig:results} shows the evolution of the metrics with the number of forward passes $M$. 
We can take $M=1$ as the baseline, since there is no epistemic contribution. In general, predictions improve as $M$ increases. However, such improvements are hardly noticeable for $M>18$.
Interestingly, for MCD CDec, the epistemic uncertainty results in a worse uncertainty calibration in comparison to a single forward pass. 

Figure~\ref{fig:u_metrics} shows the calibration error curves and sparsification error curves from where AUSE and AUCE were extracted. In these figures it can be seen that all models are overconfident, and the similarity between the sparsification error curves that leads to similar AUSE values.

\begin{figure}
    \centering
    \includegraphics[width=\columnwidth]{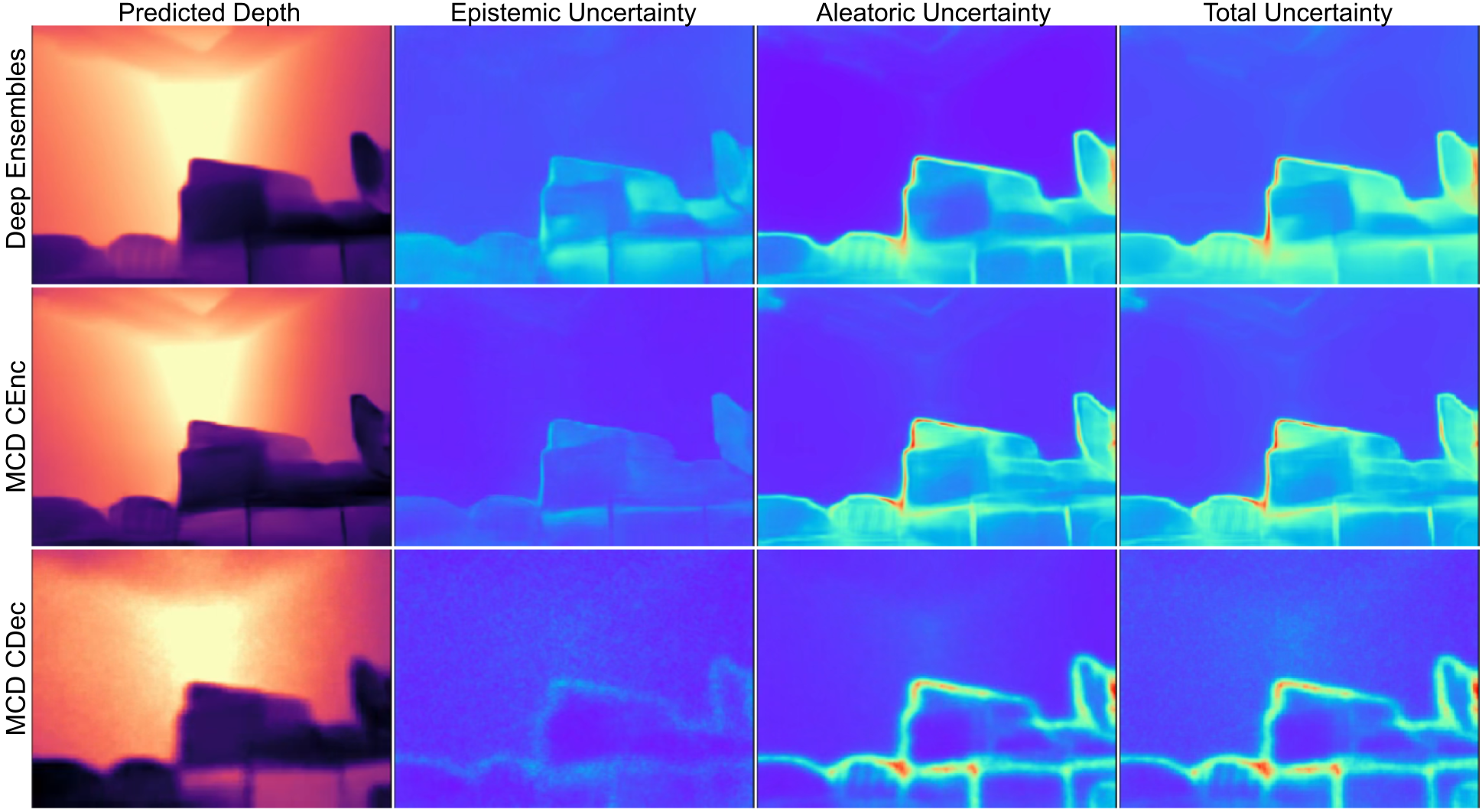}
    \caption{Depth and uncertainty predictions in a SceneNet RBG-D image.}
    \label{fig:qualitative_res}
\end{figure}

\begin{figure}
    \centering
    \includegraphics[width=\columnwidth]{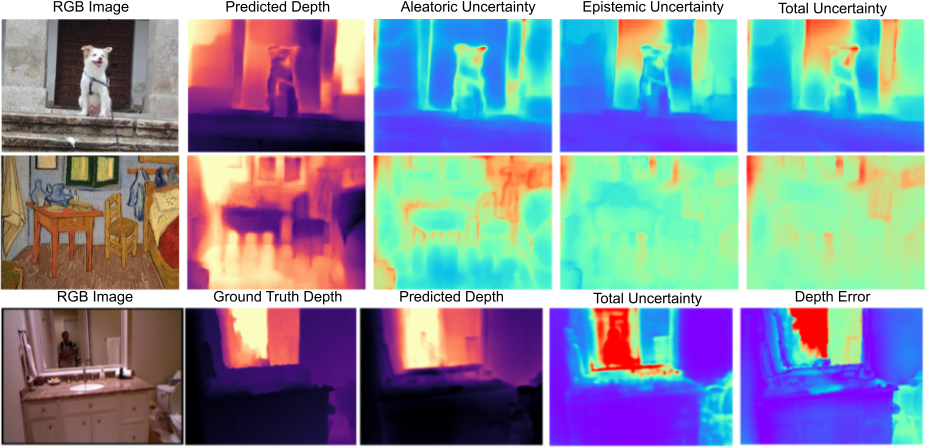}
    \caption[Predictions for three out-of-distribution images, showing high uncertainty for unfamiliar objects and textures.]{Predictions for three out-of-distribution images, showing high uncertainty for unfamiliar objects and textures. Top row: the aleatoric uncertainty is large in depth discontinuities and object boundaries, and the epistemic one concentrates in unknown patterns (the dog and the door). Middle row: aleatoric and epistemic uncertainties are both large due to large differences in appearance with respect to the training data. Bottom row: prediction in NYU Depth V2 image by an ensemble trained in SceneNet. The uncertainty is only very high for the human, not present in the training data.}
    \label{fig:outofdistr}
\end{figure}

\begin{table*}
\scriptsize
    \centering
    \begin{tabular}{|c|c|c|c|c|c|c|c|c|c|c|}
    \hline
        Model & Abs Rel  & Sq Rel & RMSE & RMSE Log & $\delta_1$ & $\delta_2$ & $\delta_3$ & AUCE  & AUSE\\\hline
Deep Ensembles&\textbf{0.1431}&\textbf{0.1052}&\textbf{0.5842}&\textbf{0.1973}&\textbf{0.8157}&\textbf{0.9596}&\textbf{0.9894}&\underline{0.1302}&\textbf{ 0.1588}\\\hline

MCD CENC-CDEC p=0.3&0.1560&0.1250&0.6332&0.2155&0.7867&0.9454&0.9849&0.1453&0.1703\\\hline
MCD CDec p=0.3 &0.1574&0.1286&0.6307&0.2160&0.7850&0.9460&0.9851&0.1382&0.1770\\\hline
MCD CEnc p=0.3 &\underline{0.1495}&\underline{0.1173}&\underline{0.6180}&\underline{0.2090}&\underline{0.8006}&\underline{0.9491}&\underline{0.9858}&0.1436&\underline{0.1653}\\\hline
MCD CENC-CDEC p=0.5 &0.1642&0.1385&0.6592&0.2264&0.7671&0.9385&0.9824&0.1356&0.1668\\\hline
MCD CEnc p=0.5 &0.1525&0.1220&0.6239&0.2127&0.7942&0.9467&0.9848&0.1377&0.1720\\\hline
MCD CDec p=0.5 &0.1578&0.1291&0.6326&0.2164&0.7861&0.9457&0.9843&\textbf{0.1286}&0.1716\\\hline
    \end{tabular}
    \caption[Depth and uncertainty metrics for MC Dropout and Deep ensembles in NYU Depth V2.]{Depth and uncertainty metrics for several variations of MC Dropout and Deep ensembles in NYU Depth V2. Best results are boldfaced, second best ones are underlined.}
    \label{tab:nyumetrics}
\end{table*}

\begin{figure*}[ht!]
    \centering
    \includegraphics[width=0.5\textwidth]{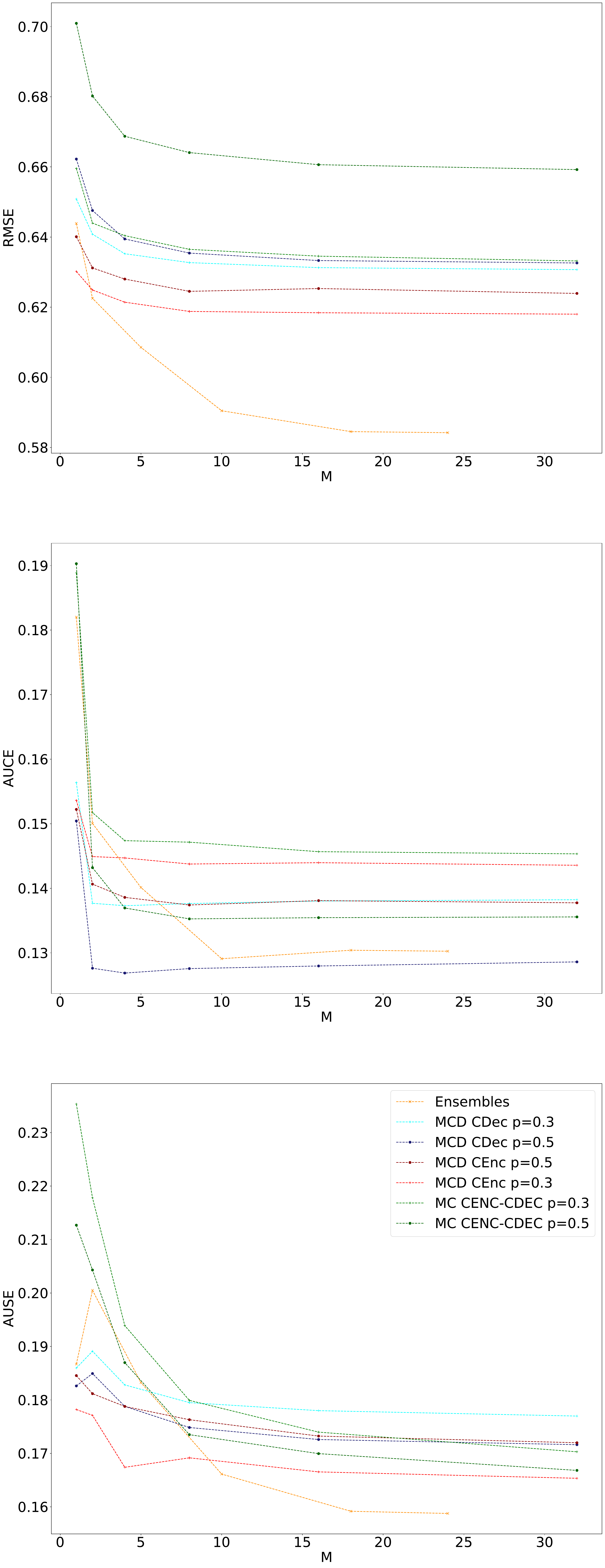}
    \caption[Comparison of MCD CDec, CEnc and Deep ensembles: RMSE, AUSE and AUCE]{
Comparison of MCD CDec, CEnc and Deep ensembles for forward passes $M$. Up: RMSE. Center: AUSE. Down: AUCE. The higher $M$ is, the better the performance, but with slight improvements for $M>18$.} 
    \label{fig:resultsnyu}
\end{figure*}

\begin{figure}
\scriptsize
    \centering
    \includegraphics[width=0.8\textwidth]{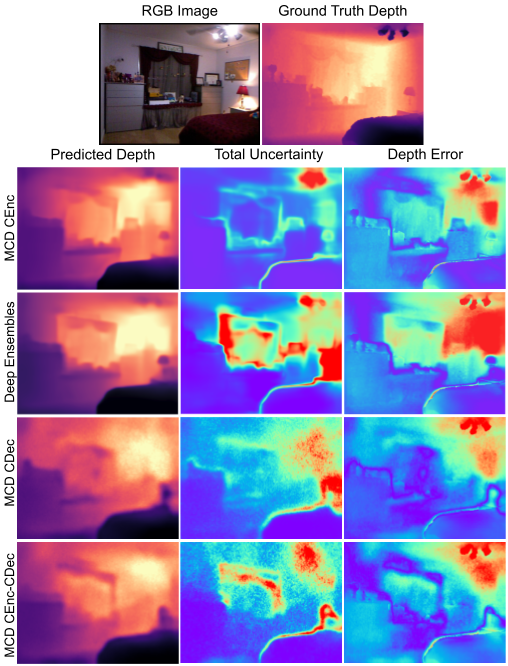}
    \caption[Depth and uncertainty results in NYU Depth v2 for MC dropout variations and deep ensembles.]{
 Depth and uncertainty results in NYU Depth v2 for MC dropout variations and deep ensembles. Top Row: input image and ground truth depth. First column: predicted depth. Second column: predicted uncertainty. Third column: depth error. Colors are equalized per image for better visualization. }
    \label{fig:qualitativeNYU}
\end{figure}

Table \ref{tab:nyumetrics} shows the metrics for depth and uncertainty on NYU Depth v2. We evaluate three MC dropout variations: MCD CEnc, MCD CDec, and MCD CEnc-CDec. In this dataset, deep ensembles show the best results in terms of uncertainty and depth metrics. 
As in SceneNet, applying dropout in the complete encoder (MCD CEnc) shows better results than applying it in the decoder (MCD CDec) or in the whole network (MCD CEnc-CDec). In this last case, the worse performance is caused by a too strong regularization effect. 
It is worth mentioning that the network is unable to recover the object boundaries and that the aleatoric uncertainty becomes diffuse for the high dropout rate case $p=0.5$. This effect becomes more noticeable when the decoder contains dropout layers. 
 Figure \ref{fig:qualitativeNYU} shows results for one sample image. Observe that the areas with higher error correspond with the areas with higher uncertainty.

For all methods, increasing the number $M$ of forward passes improves the performance in uncertainty and depth (see Figure \ref{fig:resultsnyu}). We do not see a noticeable improvement, though, for $M>18$ for deep ensembles and $M>32$ for MC dropout.

Aleatoric uncertainty appears mainly in depth discontinuities, at object edges, and regions with sharp contrast in lighting (see Figure~\ref{fig:qualitative_res}). As additional examples, Figure~\ref{fig:outofdistr} shows highly uncertain depth predictions for three out-of-distribution images. It indicates high uncertainty values for unfamiliar objects not seen during training, like the dog and the door in the background of the first picture, the unrealistic patterns of the painting ``Bedroom in Arles'' by Van Gogh in the second one, and the person in the third one.

\subsection{Bayesian pseudo-RGBD ICP}
\label{sec:pseudorgbdicp}

In this section, we evaluate the application of Bayesian depth neural networks for two-view relative motion. Relative motion can be directly computed from deep neural networks where the inductive bias can be used to estimate absolute values even for uncalibrated cameras and blurry or poorly illuminated images. But geometric methods are more precise in a multi-view setting with a known baseline for scale disambiguation and sensible features can be tracked between multiple views ~\citep{zhou2020learn}. In this work, we present a mixed approach where we use a two-view geometric method (ICP) to accurately compute the relative motion based on the depth prediction from neural networks. Thus, our approach is able to work with poor quality images and provide an unambiguous 3D transformation.


Our proposal leverages the depth predicted by a network to augment monocular images into what we call pseudo-RGBD views, and then aligns them using Iterative Closest Point (ICP). Similar ideas were proposed recently by \citet{tiwari2020pseudo} and \citet{luo2020consistent}. Differently from us, they rely on Structure from Motion~\citep{schonberger2016structure} or visual SLAM~\citep{mur2017orb} to estimate the motion from the pseudo-RGBD views. Also, we use depth uncertainty for a more informed point cloud alignment, specifically excluding highly uncertain points from ICP.

\begin{figure}
    \centering
    \includegraphics[width=\linewidth]{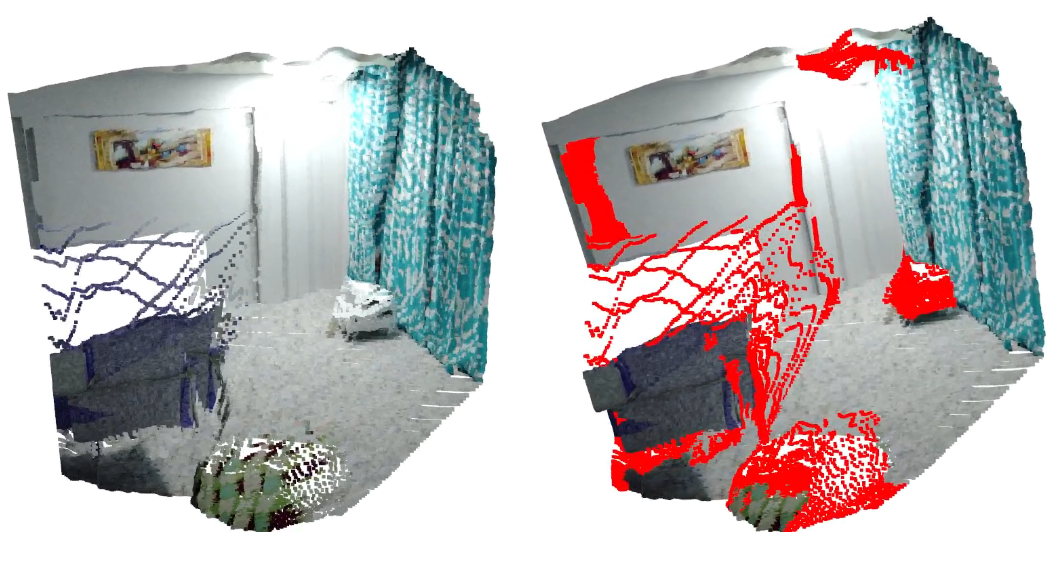}
    \caption[Single-view reconstruction with the $10\%$ most uncertain points plotted in red.]{Left: Single-view reconstruction. Right: Same reconstruction, with the $10\%$ most uncertain points plotted in red. These points corresponds to spurious or high error points that will degrade the performance of ICP.}
    \label{fig:application}
\end{figure}

\begin{table}[h!]
\centering
\scriptsize
\begin{tabular}{ |c|l|l|l|l|l|l|l| }
\hline
\textbf{Percentile} & .30 & .50 & .75 & .90 & .95 & .99 & 1.00\\\hline
\textbf{RMSE t} [m] & 0.238 & 0.216 & 0.188 & \underline{0.182} & \textbf{0.179} &0.190 &0.190\\\hline
\textbf{RMSE r} [$^\circ$] & 1.992 & 1.911 & 1.936 & \textbf{1.847} & \underline{1.888} &1.912 & 1.891\\

\hline
\end{tabular}
    \caption[ICP errors for percentiles .30, .50, .75, .90, .95, .99, 1.00.]{ICP errors for percentiles .30, .50, .75, .90, .95, .99, 1.00. Best are boldfaced, second best underlined.}
    \label{tab:application}
\end{table}

Our experimental setup is as follows. We selected $1408$ random image pairs, separated by at least $4$ frames, from SceneNet RGB-D. 
We excluded pairs with large areas without ground-truth depth (e.g., windows) and small overlap (rotations larger than $60^\circ$). We kept the image pairs for which there is sufficient evidence that ICP converged.

For each pair, we back-projected the estimated depth distributions into a point clouds and applied ICP to the following percentiles of the most certain points according to our estimation: $.30$, $.50$, $.75$, $.90$, $.95$, $.99$, and $1.00$ (the percentile $1.00$ corresponds to the full point clouds). We used our deep ensemble model, as it showed the best uncertainty calibration in Table \ref{tab:metrics}.

Figure \ref{fig:application} illustrates our hypothesis with an example. The left point cloud is the original one, and the right one corresponds to the percentile $.90$. The points highlighted in red represent the $10\%$ most uncertain points according to our uncertainty estimation, and clearly correspond to highly erroneous ones, as they lie on depth discontinuities. Removing such points from ICP will improve its accuracy.

Table \ref{tab:application} shows the translational and rotational errors for all evaluated pairs. As motivated before, removing the most uncertain points (corresponding in well calibrated models to the most errouneous ones) reduces the estimation errors. The best results are for percentiles $.90$ and $.95$ ($1.00$ corresponds to the original point cloud).
When the percentage of points removed is higher the error grows. This effect becomes obvious for percentiles $.30$ and $.50$, for which $70\%$ and $50\%$ of the points with the highest uncertainty were removed respectively. In these cases, the number of points removed is too large and the estimation becomes less accurate.

\section{Conclusions}
In this chapter, we evaluated MC dropout and deep ensembles as scalable Bayesian approaches to uncertainty quantification for single-view supervised depth learning.

We demonstrate empirically that using MC dropout in the encoder outperforms other variations used in the literature, which is a result of practical relevance. The placement of dropout in the architecture indeed has a significant effect in the estimation of depth and uncertainty. As a second conclusion of our analysis, deep ensembles have the best calibrated uncertainty estimations. However, applying dropout in the encoder performs only slightly worse than deep ensembles. 
As MC dropout needs much less memory than deep ensembles, it may be considered the Bayesian approach with more potential for applications.
In our experimental results, we also show the application of Bayesian depth networks to pseudo-RGBD ICP, with the result that relative transformation can be improved by excluding the points with highest uncertainties. 

\chapter{On the uncertain single-view depths in colonoscopies}
\label{chap:2}
Colonoscopy images, due to their complex nature and the environment they capture, are notoriously difficult for computer vision to interpret and analyze. The presence of folds or haustra results in discontinuities. Furthermore, reflections from the wet surfaces inside the colon create specular light, adding another dimension of complexity to understanding the images.
This chapter explores scalable Bayesian deep networks to predict depth and uncertainty in these challenging images under different learning paradigms. Specifically, we first benchmark thoroughly, in synthetic data, supervised and self-supervised learning approaches in the colonoscopic domain. We address the problem of domain change and demonstrate that deep ensembles models trained on synthetic data can be transferred adequately to similar real domains. 
We propose a novel uncertainty aware teacher-student method that models the uncertainty of the teacher in the learning pipeline. Furthermore, we quantify the generalization of depth and uncertainty to real scenarios.

\begin{figure}[t!]
\centerline{\includegraphics[width=0.8\textwidth,keepaspectratio]{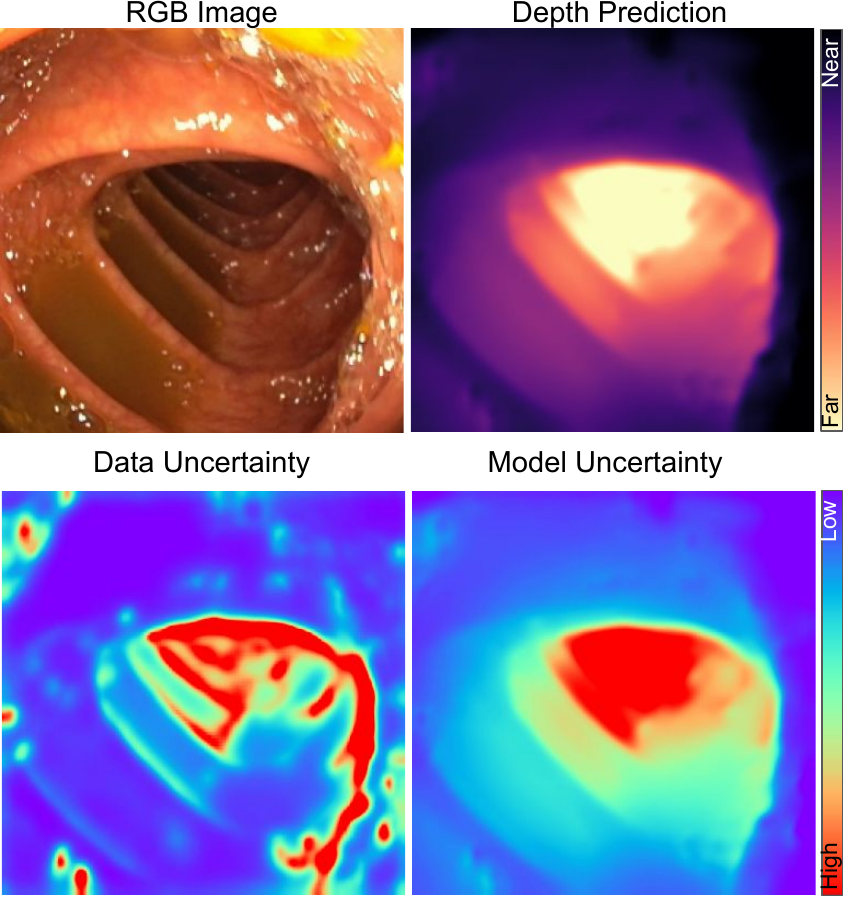}}
\caption[Depth and uncertainty predictions for a colonoscopy image.]{Depth and uncertainty predictions for a colonoscopy image. Dark/bright colors stands for near/far depths, blue/red stands for low/high uncertainties. Note the higher uncertainties in darker and farther areas and in reflections.}
\label{fig:teasermiccai}
\end{figure}
\section{Introduction}
Depth perception inside the human body is one of the cornerstones to enable automated assistance tools in medical procedures (e.g. virtual augmentations and annotations, accurate measurements or 3D registration of tools and interest regions) and, in the long run, the full automation of certain procedures and medical robotics. Monocular cameras stand out as very convenient sensors, as they are minimally invasive for in-vivo patients, but estimating depth from colonoscopy images is a challenge.
Multi-view approaches are accurate and robust in many applications outside the body, e.g.~\citet{schonberger2016structure}, but assume certain rigidity, texture and illumination conditions that are not fulfilled in in-body images. As mentioned in Chapter 2, single-view 3D geometry is ill-posed, since infinite 3D scenes can explain a single 2D view~\citep{Hartley:2003:MVG:861369}.
As discussed in such chapter, deep neural networks have shown impressive results in last years~\citep{eigen2014,fu2018deep,Godard2018}. However, the vast majority of deep learning models lack any metric or intuition about their predictive accuracy.
In a critical environment such as the inside of the human body, uncertainty quantification is essential. Specifically, in medical robotics it allows us to properly account for uncertainty in control and decision making, and in SLAM~\citep{1638022} to safely navigate inside the body. It also provides confidence intervals in in-body measurements (e.g., polyps), which is valuable for doctors to decide how to act. Uncertainty quantification is, a must-have for robust, interpretable and safe AI systems. Bayesian deep learning perfectly combines the fields of deep learning and uncertainty quantification in a sound and grounded approach. However, for high-dimensional deep networks, accurate Bayesian inference is intractable. Only bootstrapping methods such as deep ensembles~\citep{Lakshminarayanan2017} have shown to produce well-calibrated uncertainties in many computer vision tasks at reasonable cost~\citep{Gustafsson2019}. 

\section{Preliminaries and related work}
\label{sec:Related_work}
\textbf{Bayesian deep learning}
is a form of deep learning that performs probabilistic inference on deep network models. This enables uncertainty quantification for the model and the predictions. For high-dimensional deep networks, Bayesian inference is intractable and some approximate inference methods such relying on variational inference or Laplace approximations might perform poorly. In practice, sampling methods based on bootstrapping, such as deep ensembles \citep{Lakshminarayanan2017}, or Monte Carlo, like MC dropout \citep{Gal2016}, have shown to be the most scalable, reliable and efficient approaches for depth estimation and other computer vision tasks \citep{Gustafsson2019}. 
In particular, deep ensembles have shown to perform extremely well even with a reduced number of samples, because each random sample of the network weight is optimized using a maximum a posteriori (MAP) loss $\mathcal{L}_{MAP}=\mathcal{L}_{LL} + \mathcal{L}_{prior}$, resulting in a high probability sample. The MAP loss requires a prior distribution, which unless otherwise stated, we assume to be a Gaussian distribution over the weights $\mathcal{L}_{prior} = ||\theta||^2$. Similarly as chapter 2, for the data likelihood $\mathcal{L}_{LL}$, we use a loss function based on the Laplace distribution for which the predicted mean $\mu(x)$ and the predicted scale $\sigma(x)$ come from the network described in Section \ref{sec:supervisedlearning}, with two output channels \citep{Kendall2017}. The variance associated with the scale term represents the uncertainty associated with the data, also called \emph{aleatoric uncertainty} or $\sigma_a(x)$. Furthermore, in deep ensembles, the variance in the prediction from the multiple models of the ensemble is the uncertainty that is due to the lack of knowledge in the model, which is also called \emph{epistemic uncertainty} or $\sigma_e(x)$. For example, data uncertainty might appear in poorly illuminated areas or with lack of texture, while model uncertainty arises from data that is different from the training dataset. Note that while the model uncertainty can be reduced with larger training datasets, data uncertainty is irreducible. 
Model uncertainty is particulary relevant to address domain changes. To illustrate that, in Section \ref{sec:Experimental_results} we present results of models trained on synthetic data and tested on real data. \\ \\ \noindent \textbf{Single-View Depth Learning} has demonstrated a remarkable performance recently. Some methods rely on accurate ground truth labels at training \citep{eigen2014,fu2018deep,Lapdepth}, which is not trivial in many application domains. Self-supervision without depth labels was achieved by enforcing multi-view photometric consistency during training \citep{Zhou2017, zhan2018unsupervised,Godard2018}.
In the medical domain, supervised depth learning was addressed by \citet{Visentini-Scarzanella2017} with autoencoders and by \citep{shen2019context} with GANs, both using ground truth from phantom models.
Other works based on GANs were trained with synthetic models~\citep{chen2019slam,mahmood2018unsupervised,mahmood2018deep,rau2019implicit}, and ~\citep{Cheng2021} added a temporal consistency loss.

Self-supervised learning is a natural choice for endoscopies to overcome the lack of depth labels on the target domain \citep{sharan2020domain,ozyoruk2021endoslam,recasens2021endo}. 
Although depth or stereo are not common for in-vivo procedures, several works use them for training \citep{luo2019details,xu2019unsupervised,huang2021self}. Others train in phantoms \citep{turan2018unsupervised} or synthetic data \citep{freedman2020detecting,hwang2021unsupervised}, facing the risk of not generalizing to the target domain. We study the limits of such generalization. SfM supervision was addressed by \citet{Lui2020} using siamese networks and by \citet{widya2021self} using GANs. Note that none of these references address uncertainty quantification, which we cover in this work. \citet{Kendall2017} combine epistemic and aleatoric uncertainty by using a MC Dropout approximation of the posterior distribution. This approach obtains pixel-wise depth and uncertainty predictions in a supervised setting.
Traditional self-supervised losses to regress depth are limited due to the aleatoric uncertainty of input images \citep{li2021sins}. 
\citet{poggi2020uncertainty} address such problem by introducing a teacher-student architecture to learn depth and uncertainty. As key advantages, teacher-student architectures provide aleatoric uncertainty for depth and avoid photometric losses and pose regression networks, which are frequently unstable. 
In this chapter, we present the evaluation for supervised and self-supervised approaches in colonoscopies images and propose a novel teacher-student approach that includes teacher uncertainty during training. Among the scalable Bayesian methods for single-view depth prediction, deep ensembles show the best calibrated uncertainty as demonstrated at Chapter \ref{chap:1} \citet{rodriguez2022bayesian} and hence we choose them as our model.
\begin{figure}
  \includegraphics[width=\textwidth]{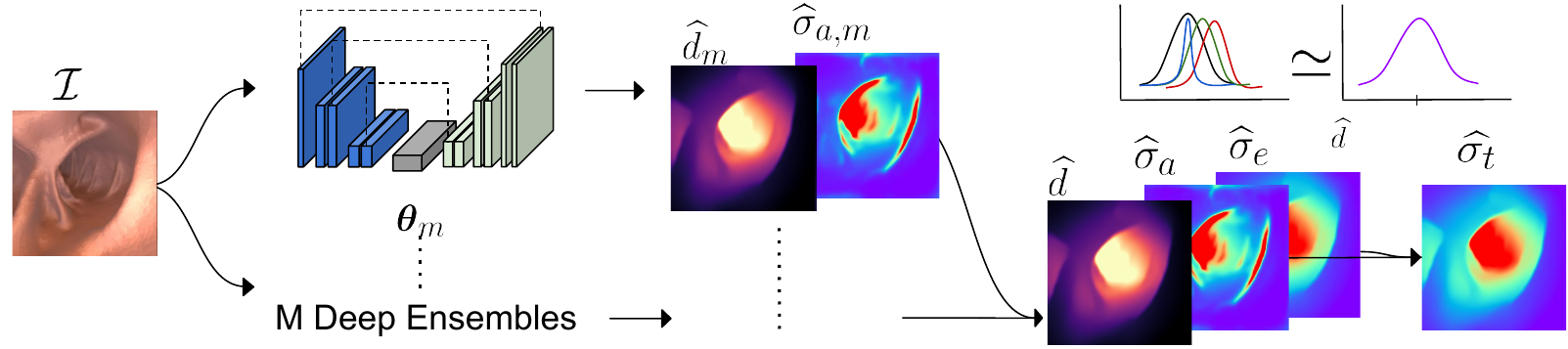}
  \centering
  \caption[Forward propagation of supervised deep ensembles.]{Forward propagation of supervised deep ensembles. Our deep ensembles model a Gaussian distribution $N(\widehat{d},\widehat{\sigma}^2_{t})$.  $\widehat{d}$ comes from averaging all ensembles depth output and $\widehat{\sigma}^2_{t}$ from joining data and model uncertainties}
\label{fig:testsup}
\end{figure}

\section{Supervised learning using deep ensembles}
\label{sec:supervisedlearning}

\begin{figure}
  \includegraphics[width=\textwidth]{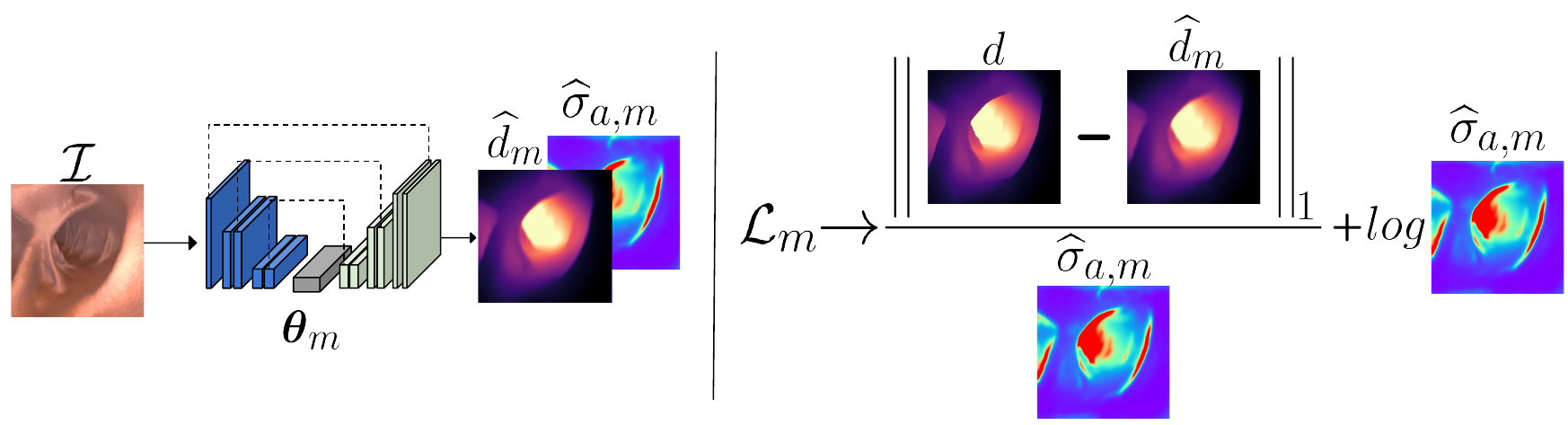}
  \centering
  \caption[Supervised training of a single ensemble \textit{m}.]{Supervised training of a single ensemble \textit{m}. Depth $\widehat{d}_m$ and aleatoric uncertainty $\sigma_{a,m},$ estimation for a target image $\mathcal{I}$. We define graphically the loss for a single ensemble. $\widehat{d}_m$ depth prediction, $d$ depth target  and $\sigma_{a,m}$ the aleatoric uncertainty of the ensemble. }
\label{fig:trainsup}
\end{figure}

Let our dataset $\mathcal{D}=\{ \{\mathcal{I}_1,d_1 \},\hdots,\{\mathcal{I}_N,d_N \}\}$ be composed by $N$ samples, where each sample $i \in \{1,\hdots,N\}$ contains the input image $\mathcal{I}_i \in \{0,\hdots,255\}^{w \times h \times 3}$ and per-pixel depth labels $d_i \in \mathbb{R}^{w \times h}_{>0}$. Regarding our network, we use an encoder-decoder architecture with skip connections, inspired by Monodepth2~\citep{Godard2018}, with two output layers. Thus, for every new image $\mathcal{I}$ the network predits its pixel-wise depth $\widehat{d}(\mathcal{I}, \boldsymbol{\theta}) \in \mathbb{R}^{w \times h}_{>0}$ and data variance $\widehat{\sigma}_{a}^2(\mathcal{I}, \boldsymbol{\theta}) \in \mathbb{R}^{w \times h}_{>0}$. As commented in Section \ref{sec:Related_work}, we use a MAP loss, with $\mathcal{L}_{prior} = ||\boldsymbol{\theta}||^2$ and :

\begin{equation}
    \mathcal{L}_{LL} = \frac{1}{w \cdot h} \sum_{\boldsymbol{j} \in \Omega_{i}} \left(\frac{||d\!\left[ \boldsymbol{j} \right] - \widehat{d}\!\left[ \boldsymbol{j} \right]||_1}{ \widehat{\sigma}_{a}\!\left[ \boldsymbol{j} \right]} +\log \widehat{\sigma}_{a}\!\left[ \boldsymbol{j} \right] \right)
    \label{eq:losssupervised}
\end{equation}
where $\left[\cdot \right]$ is the sampling operator and $\boldsymbol{j} \in \Omega$ refers to the pixel coordinates in the image domain $\Omega$. The per-pixel depth labels $d$ can be obtained from ground truth depth $d^{GT}$ or from SfM 3D reconstructions $d^{SfM}$~\citep{schonberger2016structure}. Figure \ref{fig:trainsup} presents the training of a single ensemble. 
A \emph{deep ensemble model} is composed by $M$ networks with weights $\{\theta_m\}_{m=1}^M$, each of them trained separately starting from different random seeds. We denote as  $(\widehat{d}_m,\widehat{\sigma}^2_{a,m})$ the output of the $m^{th}$ ensemble (see Figure \ref{fig:testsup}). We obtain the mean depth of the ensemble $\widehat{d}$ and its epistemic uncertainty $\widehat{ \sigma}^2_{e}$ using the total mean and variance of the full model.
The total uncertainty $\widehat{\sigma}^2_{t}  = \widehat{\sigma}^2_{a} + \widehat{\sigma}^2_{e}$ combines the data $\widehat{\sigma}^2_{a}$ and model $\widehat{\sigma}^2_{e}$ uncertainties which results from the law of total variance.
\begin{equation}
\widehat{d} = \frac{1}{M} \sum_{m=0}^{M} \widehat{d}_m, \ \  \widehat{\sigma}^2_{a} = \frac{1}{M} \sum_{m=0}^{M} \widehat{\sigma}^2_{a,m}, \ \ 
\widehat{\sigma}^2_{e} = \frac{1}{M} \sum_{m=0}^{M} \left(\widehat{d} - \widehat{d}_m\right)^2
\label{eq:supervisedL1}
\end{equation}
\section{Self-supervised learning using deep ensembles} 

Self-supervised methods aim at learning \emph{without} depth labels, the training data being $\mathcal{D}=\{ \mathcal{I}_1 ,\hdots, \mathcal{I}_N\}$ and the supervision coming from multi-view consistency. For each instance $m$ of a deep ensemble, two deep networks are used \citep{Godard2018}. 
\begin{figure}[h!]
    \centerline{\includegraphics[width=\textwidth]{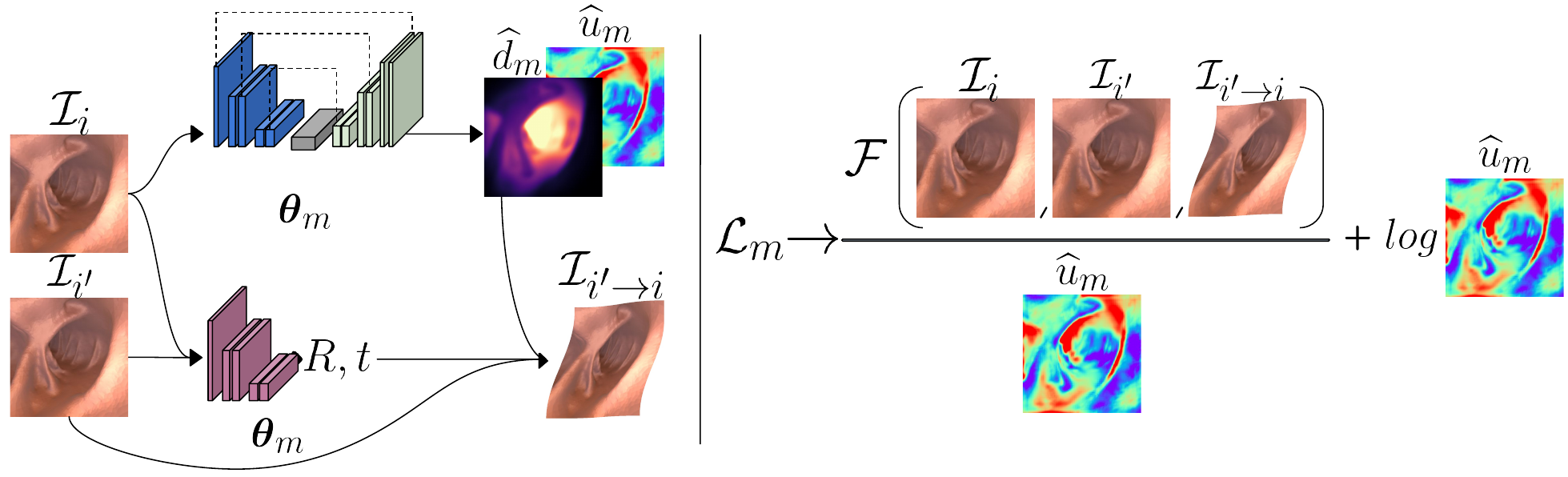}}
    \caption{Self-supervised training of deep ensembles.}
    \label{fig:Approach_Self-Supervised}
\end{figure}

The first one learning depth and a photometric uncertainty parameter $\widehat{u}$ and the second one learning to predict relative camera motion. We use a pseudo-likelihood for the loss function, that uses both networks for photometric consistency:
\begin{equation}
    \mathcal{L}_{LL,m} = \frac{1}{w \cdot h} \sum_{\boldsymbol{j} \in \Omega_{i}} \left( \frac{\mathcal{F}_p\!\left[{\boldsymbol{j}}\right] }{ \widehat{u}_{m}\!\left[{\boldsymbol{j}}\right]} +\log \widehat{u}_{m}\!\left[{\boldsymbol{j}}\right]\right)
    \label{eq:unsupervised_loss}
\end{equation}

where $\mathcal{F}_p$ is the photometric residual and $\widehat{u}_{m}$ an uncertainty prediction (See Figure \ref{fig:Approach_Self-Supervised}) The photometric residual $\mathcal{F}_p\!\left[{\boldsymbol{j}}\right]$ of pixel $\boldsymbol{j}$ in a target image $\mathcal{I}_i$ is the minimum --between the warped images $\mathcal{I}_{i^{\prime} \rightarrow i}$ from the previous and posterior images $\mathcal{I}_{i^\prime}$ to the target one $\mathcal{I}_i$-- of the sum of the photometric reprojection error and Structural Similarity Index Measure (SSIM)~\citep{wang2004image}:

\begin{equation}
    \mathcal{F}_p\!\left[{\boldsymbol{j}}\right] = \min ( (1-\alpha) \lVert \mathcal{I}_{i}\!\left[{\boldsymbol{j}}] \! - \mathcal{I}_{i^{\prime} \rightarrow {i}}\![{\boldsymbol{j}}\right] \rVert_1 +
    \frac{\alpha}{2} (1 -  \text{SSIM}\! (\mathcal{I}_{i},\mathcal{I}_{i^{\prime} \rightarrow {i}},\boldsymbol{j}) \!)
\end{equation}
being $\alpha \in \left[0,1\right]$ the relative weight of the addends; and $\mathcal{I}_{i}\!\left[{\boldsymbol{j}}\right]$ and $\mathcal{I}_{i^{\prime} \rightarrow {i}}\!\left[{\boldsymbol{j}}\right] = \mathcal{I}_{i^\prime}\!\left[{\boldsymbol{j^\prime}}\right]$ the color values of pixel $\boldsymbol{j}$ of the target image $\mathcal{I}_{i}$ and of the warped image $\mathcal{I}_{i^{\prime} \rightarrow i}$. To obtain this latter term, we warp every pixel $\boldsymbol{j}$ from the target image domain $\Omega_{i}$ to that of the source image $\Omega_{i^{\prime}}$ using:
\begin{equation}
   {\boldsymbol{j}^{\prime} = \pi \! \left( \mathbf{R}_{i^{\prime} {i}} \pi^{\!-\!1}\!(\boldsymbol{j},\widehat{d}_{i}\!\left[{\boldsymbol{j}}\right] ) +  \mathbf{t}_{i^{\prime} {i}}\right)}
\label{eq:backandforthprojection}
\end{equation}
$\mathbf{R}_{i^{\prime} {i}} \! \in \! \ensuremath{\mathrm{SO}(3)}$ and $\mathbf{t}_{i^{\prime} {i}} \! \in \! \mathbb{R}^3$ are the rotation and translation from $\Omega_{i}$ to $\Omega_{i^{\prime}}$, and $\pi$ and $\pi^{-1}$ the projection and back-projection functions (3D point to pixel and vice versa).
In this case, the prior loss also incorporates an edge-aware smoothness term $\mathcal{F}_s$, regularizing the predictions \citep{Godard2018}. 
Thus, the prior term becomes $\mathcal{L}_{prior} = ||\theta||^2 + \lambda_u \mathcal{F}_s\!\left[{\boldsymbol{j}}\right]$, where $\lambda_u$ calibrates the effect of the smoothness in terms of the reprojection uncertainty. This prior term is then combined to obtain $\mathcal{L}_{MAP}$ as described in Section \ref{sec:Related_work}.
We obtain the ensemble prediction by model averaging as in the supervised case (Eq. \ref{eq:supervisedL1}).
In this case, the data uncertainty for the depth prediction $\widehat{\sigma}^2_{a,m}$ cannot be extracted from the photometric uncertainty parameter $\widehat{u}$. Due to this, only model uncertainty will be considered in the experiments ($\widehat{\sigma}^2_{t}  = \widehat{\sigma}^2_{e}$).

\section{Teacher-student with uncertain teacher}

\begin{figure}[h!]
    \centerline{\includegraphics[width=\textwidth]{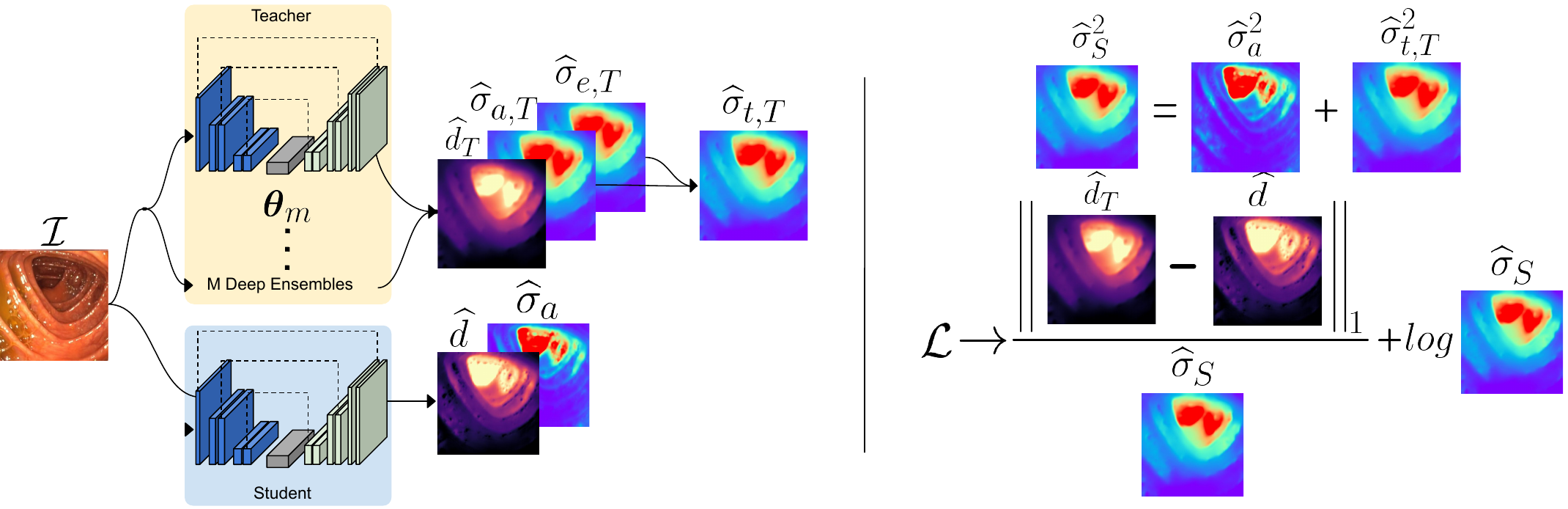}}
    
 \caption{Self-supervised training by a teacher-student approach. 
     }
    \label{fig:Approach_Teacher}
\end{figure}
In the endoscopic domain, accurate depth training labels can only be obtained from RGB-D endoscopes (which are highly unusual) or synthetic data (that is affected by domain change).
We propose the use a of a Bayesian teacher trained on synthetic colonoscopies that produces depth and uncertainty labels.
The teacher's epistemic uncertainty allows us to overcome the domain gap automatically. 
Specifically, our novel teacher-student architecture models depth labels from the predictive posterior of the teacher $d \sim \mathcal{N}(\widehat{d}, \sigma^2_T)$ ($\sigma^2_T$ is the total teacher variance). Thus, the likelihood must incorporate both the teacher and student distributions, which is used in the training loss. As before, the loss is based on a Laplacian likelihood
\begin{equation}
    \mathcal{L}_{LL,m} = \frac{1}{w \cdot h} \sum_{\boldsymbol{j} \in \Omega_{i}} \left( \frac{||\widehat{d}_{T}\!\left[{\boldsymbol{j}}\right] - \widehat{d}\!\left[{\boldsymbol{j}}\right]||_1}{\widehat{\sigma}_{m}\!\left[{\boldsymbol{j}}\right]} +  \log \widehat{\sigma}_{m}\!\left[{\boldsymbol{j}}\right] \right)
    \label{eq:teacher-student}
\end{equation}
where the per-pixel variance is the sum of the teacher predictive variance and the aleatoric one predicted by the student $\widehat{\sigma}_{m}^2 = \widehat{\sigma}_{T}^2 + \widehat{\sigma}_{a,m}^2$. Our student is hence aware of the label reliability, which will be affected by the domain change.

\label{sec:Experimental_results}
\section{Experimental results}
We present findings on both simulated and real colonoscopies.  Our first dataset is the one generated by \citep{rau2019implicit}, containing RGB images rendered from a 3D model of the colon in 15 different textures and illumination conditions. The second one, the EndoMapper dataset \citep{endommaper} consists of real monocular colonoscopies.
\subsubsection{Synthetic colon dataset. } We evaluate three training alternatives: GT (ground truth) depth supervision, SfM supervision and self-supervision. We use $6,\!550$ images for training and $720$ images for testing. 
We observed that training more than $18$ networks per ensemble does not improve the performance significantly, so we use this number in our experiments.
\begin{figure}[h]
  \centering
  \includegraphics[width=0.45\textwidth,angle=90]{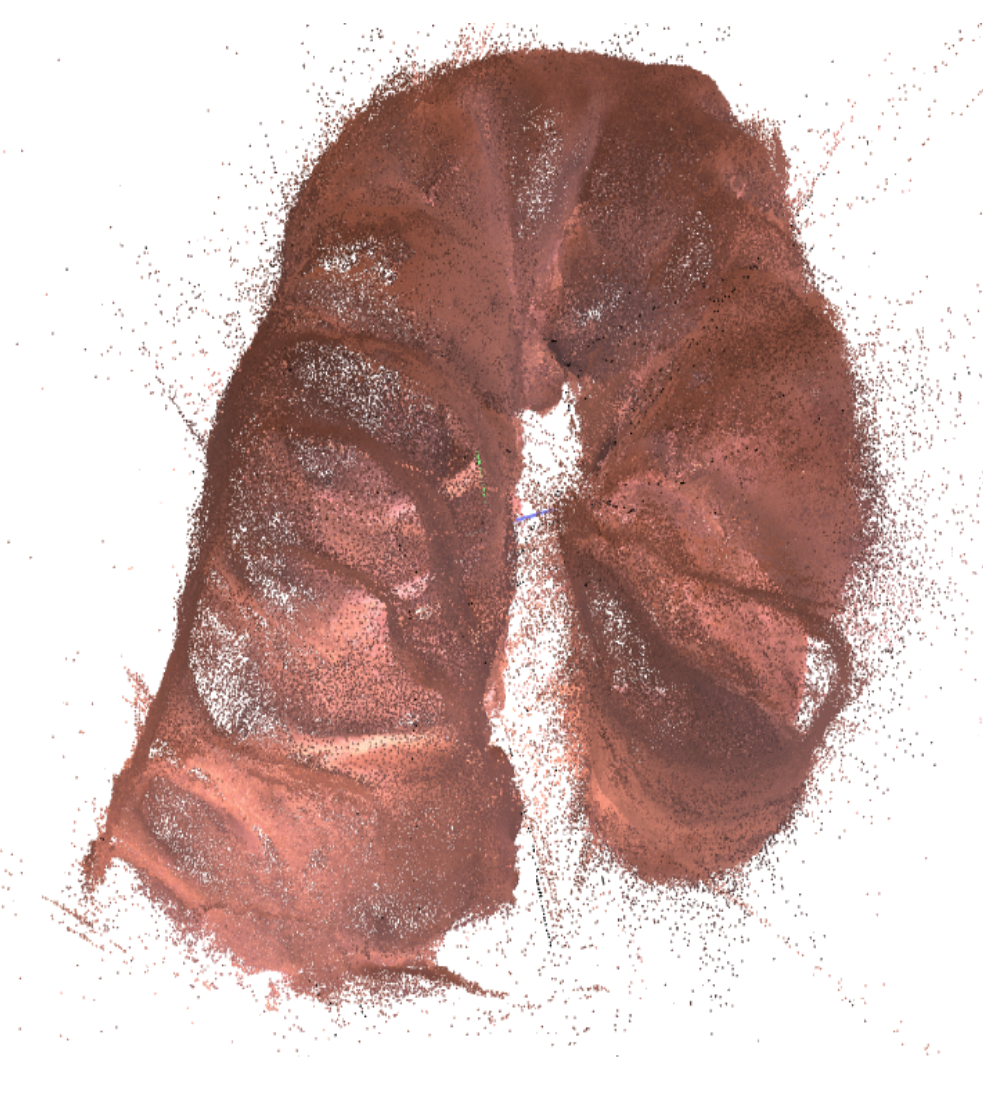}
  \caption{ Multi-view reconstruction from the colonoscopy synthetic images.}
    \label{fig:colmap-synth}
\end{figure}

In SfM-related experiments, we use COLMAP~\citep{schonberger2016structure}. Figure \ref{fig:colmap-synth} shows the 3D reconstruction from SfM. Since $d^{SfM}$ is up to scale, we compute a scale correction factor $s_i$ per image $\mathcal{I}_i$ as follows: 
\begin{equation}
    s^{SfM}_i = \frac{\text{median}(d^{GT}_i)}{\text{median}(d^{SfM}_i)}
\end{equation}
This scale correction is also applied to predictions of self-supervised and supervised SfM models that are also up-to-scale.
Table \ref{tab:MICCAImetricsGT} shows the metrics for the depth error and its uncertainty.
\begin{table}
    \resizebox{\textwidth}{!}{
\begin{tabular}{cccccccccc} Approach  & Abs\textsubscript{Rel} & Sq\textsubscript{Rel}  & RMSE & RMSE\textsubscript{log}  & $\delta < 1.25$  & $\delta < 1.25^2$  & $\delta < 1.25^3$ & AUCE  \\ \hline
 Supervised GT &\textbf{0.050} & \textbf{ 0.335}& \textbf{2.996} &  \textbf{0.102} & \textbf{0.978} & \textbf{0.993} & \textbf{0.997} &  +0.190\\
 
  Supervised SfM & 0.172 & 2.568 & 7.409 & 0.269 & 0.852 & 0.939 & 0.962 & \textbf{-0.116} \\
Self-supervised &  0.179 & 1.774 & 7.601 & 0.243 & 0.792 & 0.938 & 0.972 & -0.152 \\
    \end{tabular}
    }
    \caption{Depth and uncertainty metrics in the synthetic dataset.  RMSE in mm. }
    \label{tab:MICCAImetricsGT}
\end{table}

\begin{figure}[ht]
\centerline{\includegraphics[width=\textwidth]{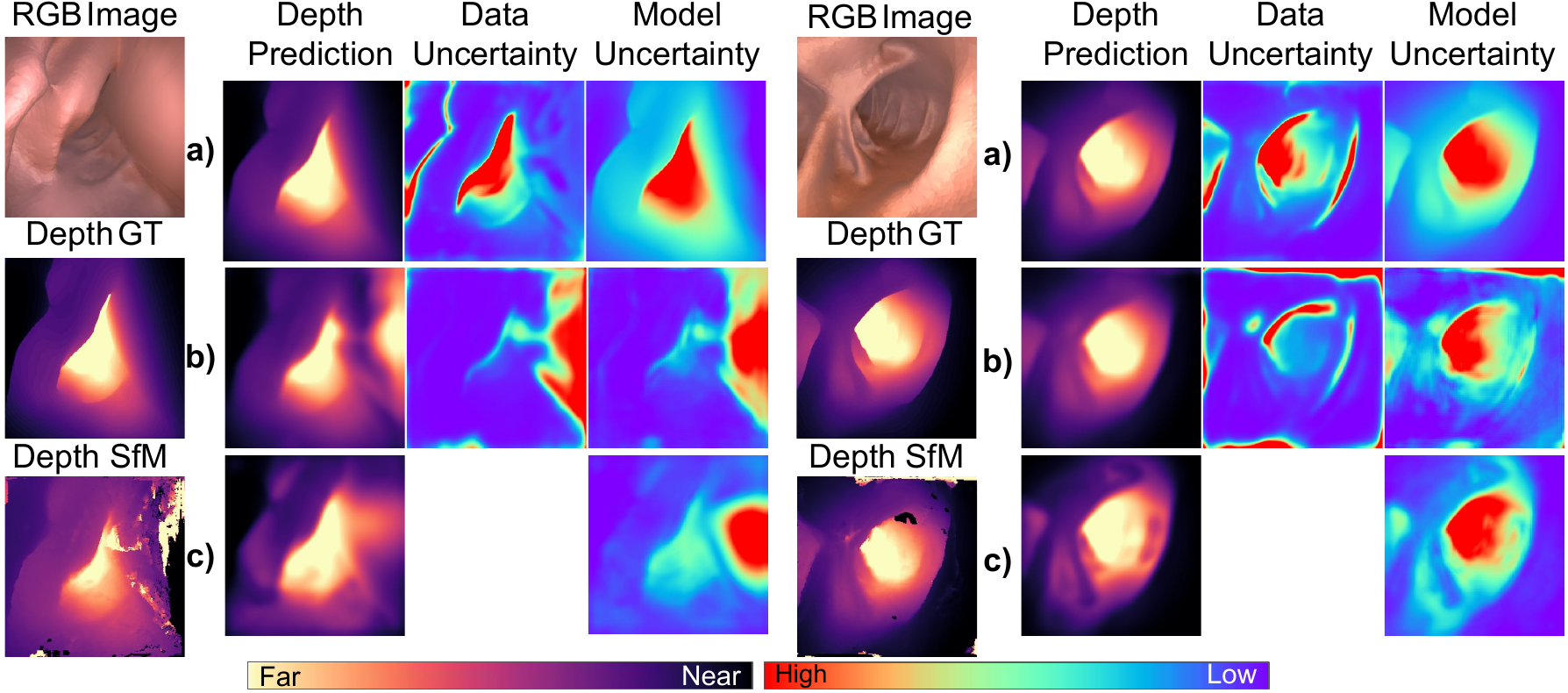}}
\caption[Qualitative depth and uncertainty examples of (supervised learning, supervised learning SfM) and self supervised learning in synthetic images.]{Qualitative depth and uncertainty examples of (supervised learning, supervised learning SfM) and self supervised learning in synthetic images. a) Supervised GT, b) Supervised SfM and c) Self-supervised}
\label{fig:examplesSynth}
\end{figure}

Supervising a deep ensemble with $d^{GT}$ labels achieves the best depth metrics. In terms of uncertainty, self-supervised and supervised with SfM are underconfident, in contrast to supervised with GT depth, which is overconfident and presents higher (worse) absolute AUCE.
Figure~\ref{fig:examplesSynth} shows that the aleatoric uncertainty supervised by GT is high around the haustra and in dark areas. The epistemic uncertainty grows with the scene depth. The uncertainty supervised by SfM is high in areas where there are typically holes in SfM reconstructions. Similarly to models trained with GT, the aleatoric uncertainty is also visible in the haustras and the epistemic in the deepest areas. 
Photometric self-supervision tends to offer the worst performance.
\vspace{1cm}

\subsubsection{EndoMapper dataset.}

\begin{figure}
  \centering
  \includegraphics[width=0.7\textwidth]{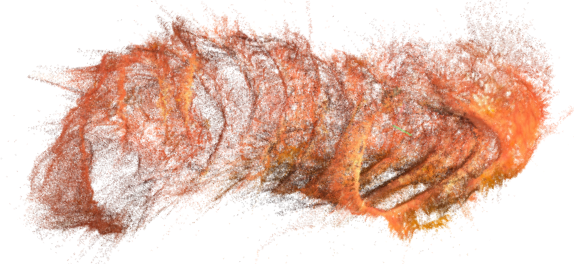}
  \caption{Multi-view reconstruction from the real colonoscopy images of the EndoMapper project using COLMAP.}
  \label{fig:realsfm}
\end{figure}

This experiment evaluates Bayesian depth networks in real colonoscopies.
We use the model previously trained with synthetic ground truth depth (``Supervised GT'') to analyse the effect of the domain change. In addition, we also present results from self-supervised training, a baseline teacher-student method \citep{poggi2020uncertainty} and our novel uncertain teacher approach. In real colonoscopies viewpoints change abruptly, images might be saturated or blurry, a considerable amount of liquid might appear and the colon itself produces significant occlusions. For this reasons, we remove images with partial or total visibility issues. We finally use $6,\!912$ images out of the $14,\!400$ images in the complete colonoscopic procedure. In order to obtain depth and uncertainty metrics, we create a 3D reconstruction of the colon using COLMAP (see Figure \ref{fig:realsfm}). We also use 18-network ensembles for all methods. Table \ref{tab:real} shows the results. Our ``Uncertain teacher'' shows in general the smallest depth errors and the highest correlation between depth errors and predicted uncertainties.
\begin{table}[h]
    \centering
    \resizebox{\textwidth}{!}{
\begin{tabular}{ccccccccc} Approach & Abs\textsubscript{Rel}  & Sq\textsubscript{Rel}  & RMSE  & 
RMSE\textsubscript{log}  & $\delta < 1.25$  & $\delta < 1.25^2$  & $\delta < 1.25^3$  & AUCE  \\ \hline
Supervised GT & 0.240 &  0.644 & 2.595 & 0.308 & 0.645 & 0.898 & 0.962 &-0.148\\
Self-supervised & 0.371 & 1.260 & 4.603 & 0.431 & 0.417 & 0.721 &  0.886 & -0.273 \\
Teacher-student & 0.234 & 0.600 & 2.532 & 0.301 & 0.657 & 0.903 & 0.963 &-0.328 \\
Uncertain teacher (ours) & \bf{0.230} & \bf{0.572} & \bf{2.458} & \bf{0.298} & \bf{0.667} & \bf{0.906} &  \bf{0.964} & \bf{-0.129}\\
\end{tabular}
}
    \caption{Depth and uncertainty metrics in the EndoMapper dataset.}
    \label{tab:real}
\end{table}

\begin{figure}
\centering
\includegraphics[width=0.8\textwidth]{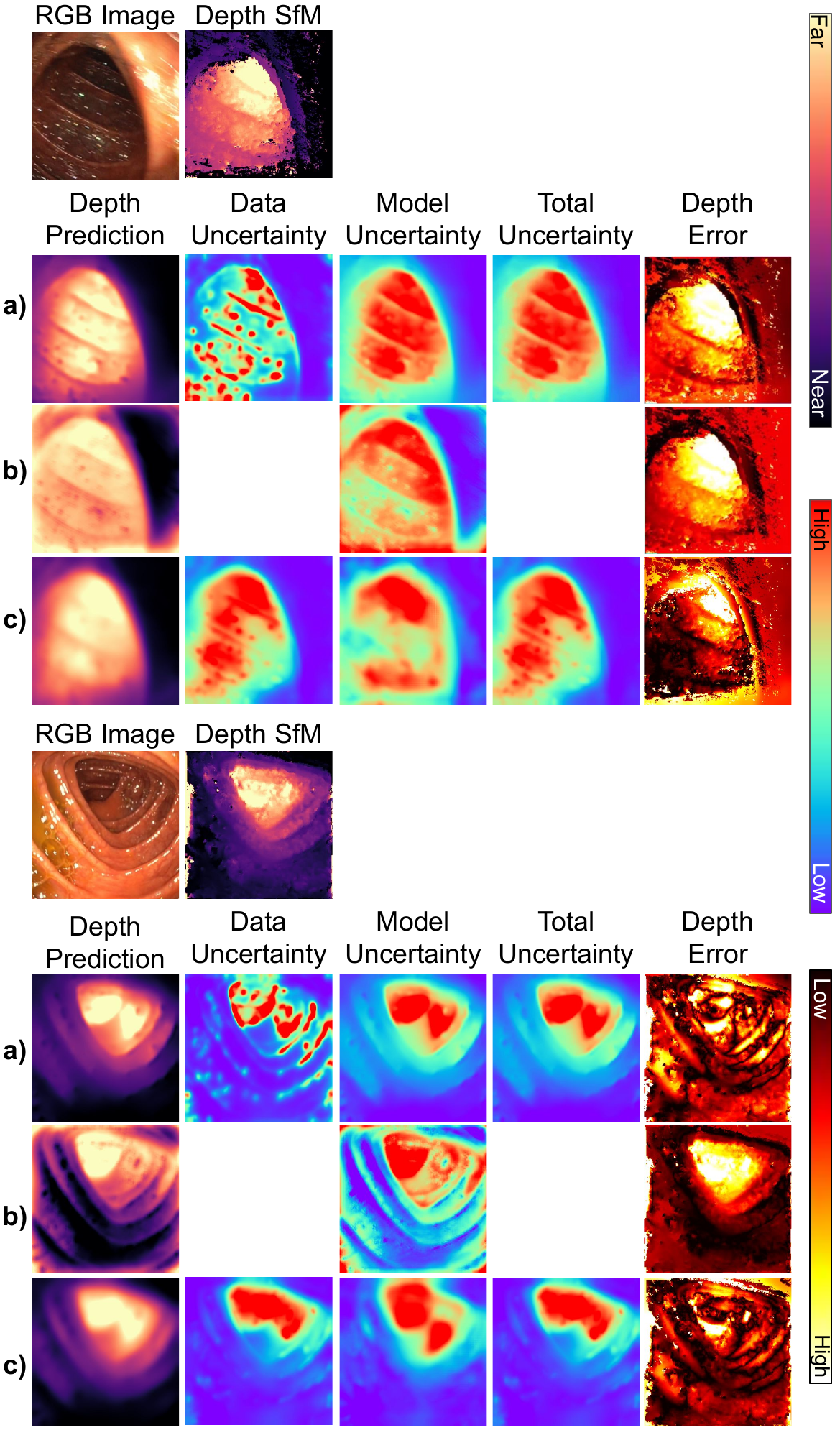}
\caption[Qualitative depth and uncertainty examples for supervised, self-supervised, and uncertain teacher-student learning in real images.]{Qualitative depth and uncertainty examples for a) supervised, b) self-supervised, and c) uncertain teacher-student learning in real images.}
\label{fig:examples}
\end{figure}

\begin{figure}
  \includegraphics[width=\textwidth]{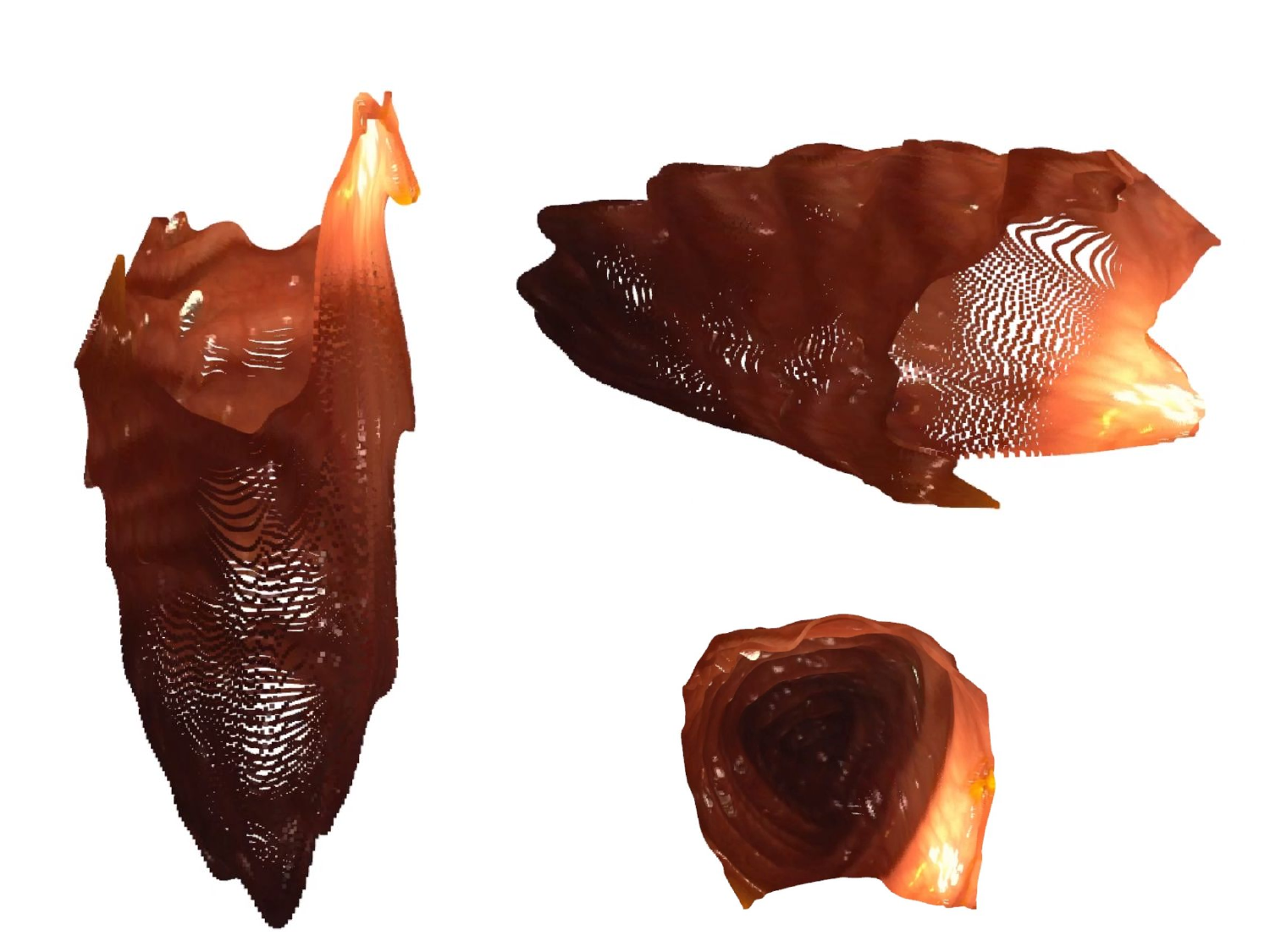}
\caption{3D reconstruction based on input image and depth from our Uncertain teacher method.}
\label{fig:teacherst3D }
\end{figure}

For self-supervised methods, this real setting is challenging due to reflections, fluids and deformations, all of them aspects that are not considered in the photometric reprojection model of self-supervised losses. ``Supervised GT'' is affected by domain change, as it was trained on synthetic data. However, we observe that it successfully generalizes to the real domain and outperforms the self-supervised method. Based on this observation, we use synthetic supervision in the ``Teacher-student'' baseline and our ``Uncertain teacher''. In general, teacher-student depth metrics outperform the models trained with GT supervision in the synthetic domain and with self-supervision in the real domain. However, ``Teacher-student'' presents the worst AUCE metric, as the teacher uncertainty is not taken into account at training time. Our ``Uncertain teacher'' is the one presenting the best depth and uncertainty metrics, as it appropriately models the noise coming from domain transfer in the depth labels.
Figure \ref{fig:examples} shows qualitative results for the ``Supervised GT'', ``Teacher-student'' and ``Uncertain teacher'' models. Note that the data uncertainty captures light reflection and depth discontinuities in supervised learning. On the other hand, the model uncertainty grows for the deeper areas. Observing these results, we can conclude that the domain change from synthetic to real colon images is not significant. Models trained on synthetic data generalize to real images and outperform models trained with self-supervision on the target domain, due to the challenges mentioned in the previous paragraphs. Finally, figure \ref{fig:teacherst3D } shows a 3D reconstruction based on depth from our method ``Uncertain teacher'' and the input image. 

\section{Conclusions}
\label{sec:Conclusion}
All systems building on depth predictions from color images benefit from uncertainty estimates, in order to obtain robust, explainable and dependable assistance and decisions. In this chapter, we have explored for the first time supervised and self-supervised approaches for depth and uncertainty single-view predictions in colonoscopies. 
From our experimental results, we extract several conclusions. Firstly, using ground truth depth as supervisory signal outperforms self-supervised learning and results in better calibrated models. Secondly, approaches based on photometric self-supervision and on SfM supervision coexist in the literature and there is a lack of analysis and results showing which type is more convenient. 
Thirdly, our experiments show that models trained in synthetic colonoscopies generalize to real colonoscopy images. Finally, we have proposed a novel teacher-student architecture that incorporates the teacher uncertainty in the loss, and have shown that it produces lower depth errors and better calibrated uncertainties than previous teacher-student architectures.
\chapter{LightDepth: single-view depth self-supervision from illumination decline}
\label{chap:3}
In this chapter, we introduce a novel self-supervision methodology that exploits illumination decline data of systems where source light and cameras are co-located, paving the way for enhanced depth estimation in endoscopic settings. 
Several works have proposed methods based on supervised learning, which struggle with simulation to real domain shifts. 
On the other hand, methods based on multi-view self-supervision face challenges related to the specific endoscopic domain, such as deformations and small baselines. Assuming both limitations, we approach the problem from a different perspective by including a light model in the method. Our experiments show that by doing that we are able to outperform both limitations in the literature, outperforming all previous approaches.

\section{Introduction}
Minimally invasive medical procedures such as gastroscopies, colonoscopies and bronchoscopies rely on endoscopes that should be as small as possible. As a result, they usually house a single camera and several light points, but neither depth nor stereo cameras. 3D reconstruction is relevant in endoscopies, as it may unlock several functionalities such as the accurate estimation of the size and shape of tumors. However, both single- and multi-view depth estimation methods present significant challenges in this domain. The lack of sufficient depth annotated data hinders the use of supervised depth learning. The presence of fluids that either obscure the view or generate specularities, the sudden illumination changes, the paucity of texture and the surface deformations hamper multi-view methods both for self-supervising deep networks and for geometry estimation. Real in-body textures and fluids are hard to simulate realistically, and the synthetic-to-real gap may be large. 

\begin{figure}[t]
  \includegraphics[width=\linewidth]{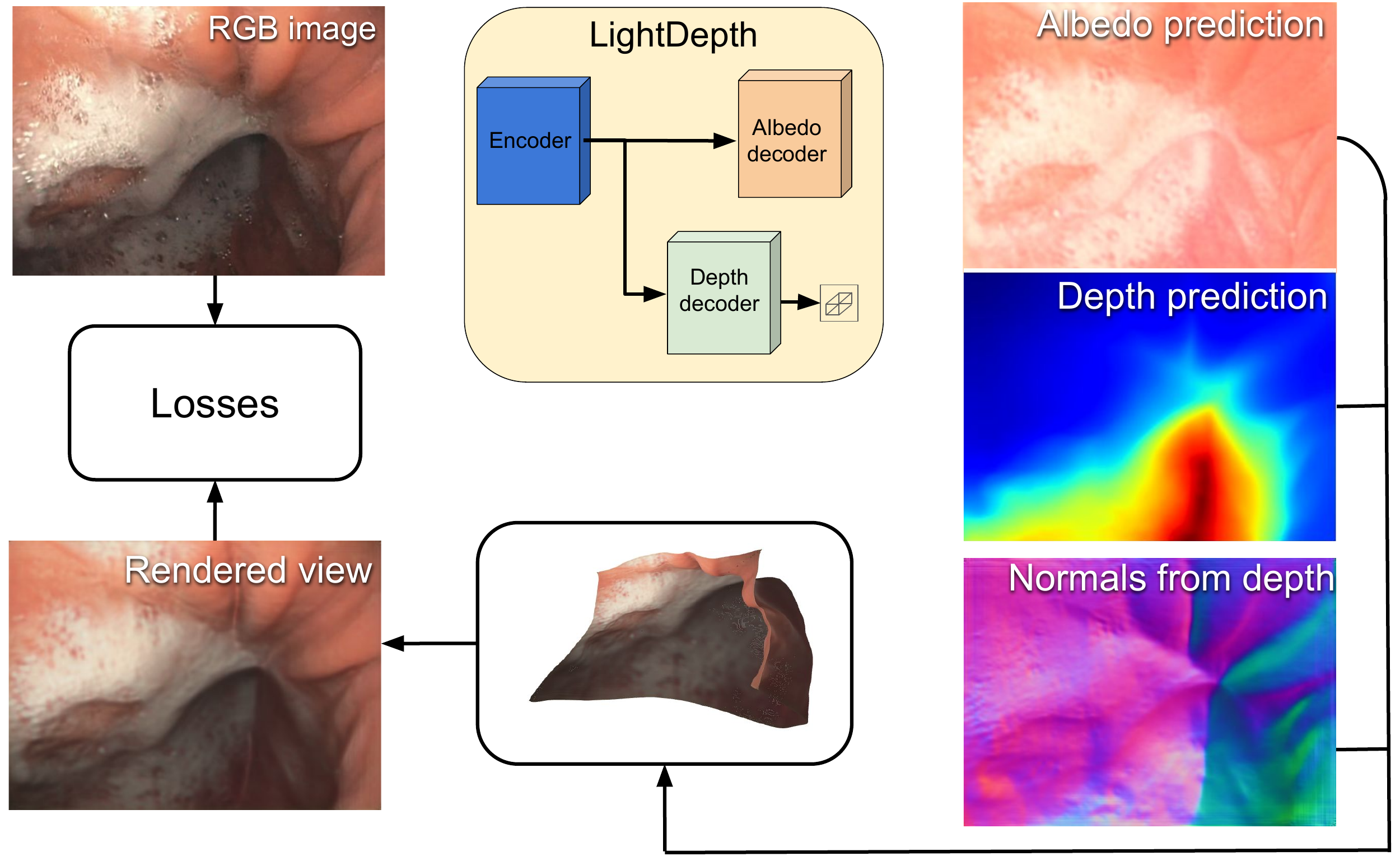}
\caption[Overview: Single-view depth self-supervision in LightDepth.]{{\bf Single-view depth self-supervision in LightDepth.} A two-headed deep network predicts albedo and depth from a single image and estimates surface normals from predicted depths. These are used to render a new image, that takes into account illumination decline and the endoscope's photometric calibration, and can be compared to the original one. Minimizing the difference between the original and rendered images is used at training time to compute the network weights and at inference time to refine the depth predictions.}
\label{fig:teaser}
\end{figure}

We propose a novel approach to depth in endoscopies that overcomes all the above challenges related to depth supervision, multi-view estimation and synthetic-to-real gaps. Our key insight is that, by exploiting a key property of endoscopic imagery, we can provide strong depth self-supervision signals from just one view. In endoscopes, the light source is rigidly located next to the camera and is close to the surface to be reconstructed. As a result, unlike in traditional shape-from-shading (SfS), points with the same albedo are imaged darker the further they are, being the decrease of intensity a function of the square distance to the light source. To exploit this, we introduce a deep network, as depicted by Figure~\ref{fig:teaser}, that estimates depths and albedos from the image, infers normals from depths, and then renders an image while taking into account the attenuation factor due to the distance between the light source and the surface. At training time, we minimize the difference between the original and rendered images. This enforces consistency of the depths, normals, and albedos and provides the required self-supervision without depth annotations. At inference, we use our trained network to predict depth from a RGB image and then, as our method is fully self-supervised, we can perform test-time refinement (TTR) for every monocular image, minimizing the difference between the input and rendered views, further refining the predicted depths.
Our quantitative evaluation on a phantom colon dataset, where ground-truth is available, shows that our {\it self-supervised} approach delivers results that are very close to that of the best supervised one, and significantly superior to that of multi-view self-supervision and synthetic-to-real transfer methods. Crucially, we show quantitatively that our method keeps working on real data, for which there is no ground-truth data that can be used for training and self-supervised alternatives underperform. 
The main specific contributions that led to such results are 1) the inclusion of illumination decline and the endoscope's photometric calibration in the rendering equation, which provides a strong supervisory signal, and 2) a single-view self-supervised method using such renders, including two-headed network architectures LightDepth U-Net and LightDepth DPT that can be trained in large colonoscopy datasets without requiring ground truth labels and even further refined online in the test views.

\section{Related work}

\subsubsection{Generic single-view depth estimation.} 
It has enjoyed a renaissance after the seminal work by~\citet{eigen2014}, which demonstrated the effectiveness of deep neural networks for supervised pixel-wise depth regression in natural images. Subsequent research efforts have made contributions in many different directions. To name a few, network architectures evolved to fully convolutional in the work of~\citet{Laina2016} and more recently to transformers~\citep{Ranftl_2021_ICCV,bhat2021adabins,li2022binsformer}. Some of those works~\citep{bhat2021adabins,li2022binsformer} also discretize the continuous depth space into bins and formulate the problem as an ordinal regression, as done by~\citet{fu2018deep}. Other advances include interpretability~\citep{Dijk_2019_ICCV}, uncertainty quantification~\citep{poggi2020uncertainty,rodriguez2022bayesian}, and modeling camera intrinsics~\citep{facil2019cam,gordon2019depth}. All these approaches are supervised and require depth ground-truth data, which can be difficult and expensive to acquire. 

Self-supervised methods seek to overcome this limitation and reduce the need for ground-truth data, often by exploiting multi-view photometric consistency~\citep{monodepth17,Zhou2017,Zhou2018,Yang2018,johnston2020self,Godard2018}. This also enables depth refinement at test time~\citep{chen2019selfsup,tiwari2020pseudo,luo2020consistent,shu2020feature,watson2021temporal,izquierdo2022sfm}. Unfortunately, this kind of supervision can be noisy, due to inaccuracies in the camera motion estimation, perspective distortions, occlusions or non-Lambertian effects, among others. As result, state-of-the-art self-supervised methods typically suffer from significantly larger inaccuracies than supervised ones. By contrast, our approach avoids these sources of errors and delivers accuracies that are close to those of supervised techniques.

\subsubsection{Endoscopic single-view depth estimation}
Single-view depth estimation has been extensively studied for endoscopic purposes. \citet{Visentini-Scarzanella2017} used CT renderings for depth supervision in bronchoscopies. However, CT scans in particular and ground-truth depth data in general are very rare in endoscopy, which makes self-supervision a quasi necessity. Many works explore multi-view integration \citep{luo2019details,xu2019unsupervised,huang2021self} combined with tracking and SLAM pipelines \citep{recasens2021endo,ozyoruk2021endoslam,ma2021rnnslam}. Others propose video-based training schemes~\citep{karaoglu2021adversarial,freedman2020detecting,hwang2021unsupervised}. Unfortunately multi-view self-supervision  is even more challenging  in endoscopy than in other areas due to the presence of deformations and weak texture. 

Due to the specificity of the domain, synthetic to real transfer has also been extensively explored. For example, a conditional GAN is used for depth recovery while integrating SLAM and multi-view inputs by~\citet{shen2019context}. \citet{chen2019slam} trained a depth network with synthetic images of a simple colon model and fine-tuned with domain-randomized photorealistic images rendered from CT scans. Many other works address the domain shift between simulated and real colons~\citep{mahmood2018unsupervised,mahmood2018deep,rau2019implicit,karaoglu2021adversarial,Cheng2021,Rodriguez2022}. Learning in supervised and transferring the knowledge using uncertainty~\citep{Lui2020} uses monocular videos and multi-view stereo to provide weak depth supervision. 
We will show in the results section that our approach yields more accurate results, especially given that our approach to self-supervision allows further refinement of the estimates at inference time.

\subsubsection{Shape from Shading (SfS).}
Depth estimation from a single image can be traced back to the early SfS methods summarized by \citet{Ruo1999}, and in particular to traditional shape-from-shading~\citep{Horn89}. However, these older techniques rely on strong assumptions that do not hold in endoscopic imagery: the camera and directional point light model are located at infinity; the reflectance is Lambertian; the albedo is constant, and the surfaces are smooth. 

Importantly, lights at infinity result in ill-posed problems~\citep{Prados2005}. By contrast, when the light source is co-located with the camera that is \emph{not} distant from the target surfaces, there is a $\nicefrac{1}{d^2}$ attenuation of pixel intensity with distance $d$ to the surface, which makes the problem well-posed when the albedo is assumed to be constant. Experimental validation shows that this still holds when the light source is translated with respect to the optical centre is provided in~\citep{ Collins2012towards,Visentini2012}, but still assuming constant and known albedo.  Photometric stereo infers depth capturing several images from the same monocular camera under lights at different locations, but requires endoscopic hardware modifications \citep{hao2020photometric,parot2013photometric,collins20123d}.

 More recently, the topic was revisited by SIRFS (Shape, Illumination, and Reflectance from Shading)~\citep{Barron2015} that model the interdependences between shape, illumination and reflectance, and introduces statistical priors on these quantities to disentangle their effects. In subsequent works~\citep{lettry2018unsupervised,li2020inverse,sang2020single,lichy2021shape,zhang2022modeling,sfsnetSengupta18}, priors are learned by deep neural networks using supervision, synthetic-to-real or multi-view self-supervision. In contrast, our approach does not require such priors, which makes its deployment easier.

The SfS methods applied to endoscopy require an accurate geometrical and photometrical model of the camera and light source. This can be obtained with endoscope calibration \citep{modrzejewski2020light, hao2020light, endommaper, batlle2022photometric}.

\newcommand{\bA}{\mathbf{A}}
\newcommand{\bC}{\mathbf{C}}
\newcommand{\bI}{\mathbf{I}}
\newcommand{\bD}{\mathbf{D}}
\newcommand{\bN}{\mathbf{N}}

\section{LightDepth}
\label{sec:method}

\begin{figure}
\includegraphics[width=\textwidth]{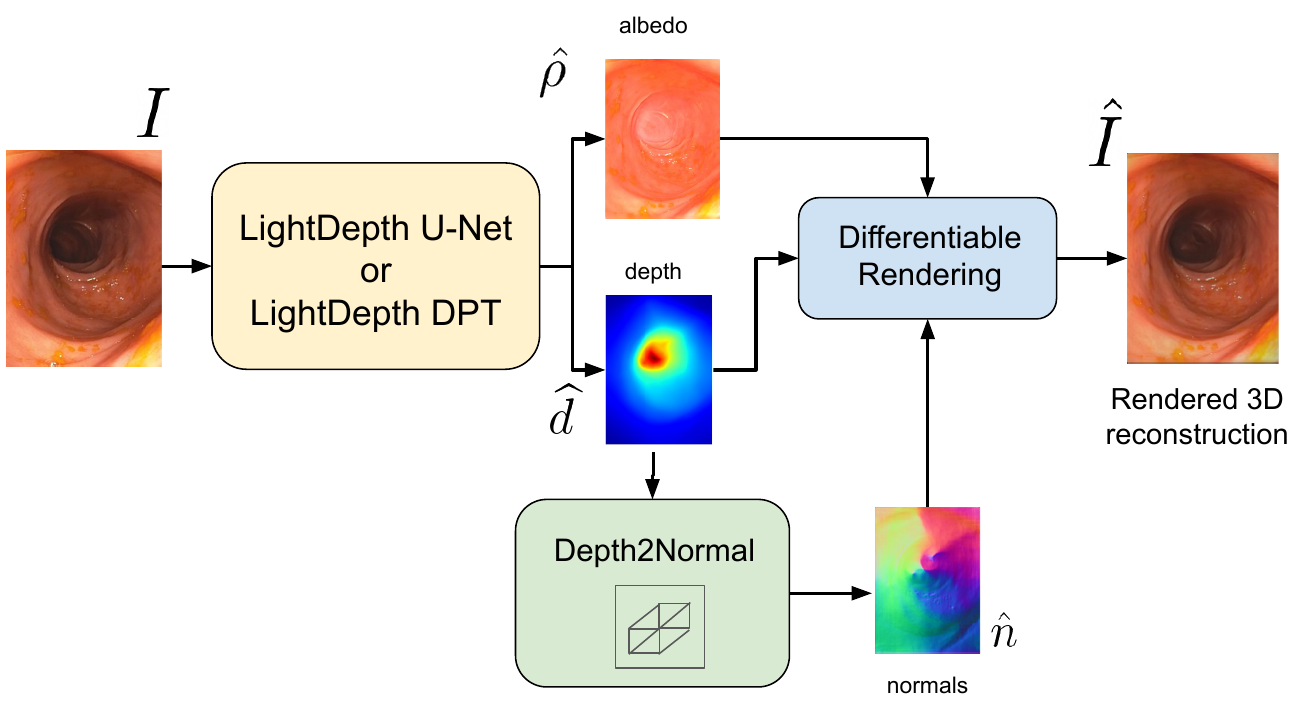}
  \caption[Differentiable Rendering Pipeline for LightDepth.]{{\bf Differentiable Rendering:} The input image is fed into a neural network that predicts albedo and depth values for each pixel. From the estimated depths, we compute the normals at each pixel surface using a kernel-based approach. Then, the depths, albedos, and normals are sent to a differentiable renderer that takes into account illumination decline and the endoscope's photometric model, and generates a synthetic image that should be as similar as possible to the original one. We also use specular reflections in saturated pixels to self-supervise normals.}
  \label{fig:difrend}
\end{figure}

We use a self-supervised single-view approach to train a neural network to predict the albedo, depth, and normals at every pixel of an image so that the image can be resynthesized from these values in a differentiable rendering pipeline (Figure~\ref{fig:difrend}). We exploit this property using a dual-branch network that outputs pixel-wise depths and albedos. The normals are estimated  analytically from the depths, and, together with the albedos, are used to render images that should be as close as possible to the original ones. At the heart of this approach is the fact that the renderer takes into account light decline as a function to distance to the surface. This is what provides the necessary self-supervisory signal. 

\subsection{Photometric model}
\label{sec:photometric-model}

\begin{figure}[ht!]
  \centering
  \includegraphics{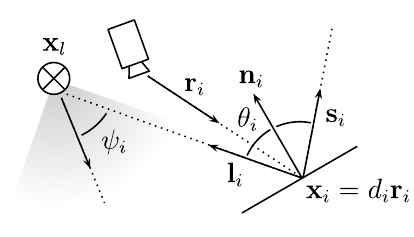}
\caption[Spotlight illumination model.]{Spotlight illumination model, a spotlight source at position $\mathbf{x}_l$ illuminates the surface point $\mathbf{x}_i$. The emission has $R(\psi_i)$ radial fall-off, suffers from an inverse-square decline with $\mathbf{x}_l \rightarrow \mathbf{x}_i$ and attenuates with the incidence angle ($\theta_i$). $\mathbf{l}_i$, $\mathbf{n}_i$, $\mathbf{r}_i$ and $\mathbf{s}_i$ are unit vectors.}
\label{fig:photo-model}
\end{figure}

As in the work of~\citet{batlle2022photometric} and~\citet{modrzejewski2020light}, 
we model scene illumination  as coming from a single spotlight source located at $\mathbf{x}_l \in \mathbb{R}^3$ in the camera reference frame, as depicted by Figure~\ref{fig:photo-model}. Spotlights usually emit with different intensities in each direction. Hence, we adopt the spotlight model (SLS) of~\citet{modrzejewski2020light}.
For surface point  $\mathbf{x}_i$ with off-axis angle $\psi_i$, we write its radiance as
\begin{equation}
    \sigma_{SLS}(\mathbf{x}_i, \psi_i) = \frac{\sigma_0}{\lVert \mathbf{x}_i - \mathbf{x}_l \rVert^2} R(\psi_i) 
\end{equation}

\begin{equation}
    R(\psi_i)  = e ^ {-\mu \left(1 - \cos \left( \psi_i \right)\right)}
\end{equation}
where $\sigma_0$ is the maximum radiance and $R(\psi_i)$ is the radial attenuation controlled by a spread factor $\mu$. Note that the light reaching the surface is subject to the inverse-square law and decays with the propagation distance from $\mathbf{x}_l$ to $\mathbf{x}_i$. 

\subsubsection{Light decline.} 
In endoscopes, the camera and the light source move jointly in a dark environment. Hence, the attenuation of the illumination is an indirect indicator of scene depth as seen from the camera. More specifically, for each pixel, we can write the rendering equation 
\begin{small}
\begin{equation}
\label{eq:im}
    \mathcal{I}(d_i, \rho_i, g)
    =
    {\left(
        \frac{\sigma_0}{\lVert d_i\mathbf{r}_i - \mathbf{x}_l \rVert^2} R(\psi_i)
        \cos\left(\theta_i \right)\;
        \rho_i \;
        g
    \right)}^{1/\gamma} \; ,
\end{equation}
\end{small}
where
$d_i$ is the depth of the $i$-th pixel with image coordinates $\mathbf{u}_i$, $\mathbf{r}_i = \pi^{-1}(\mathbf{u}_i)$ is the camera ray such that $\mathbf{x}_i = d_i\mathbf{r}_i$ and  $\pi^{-1}(\cdot)$ is the inverse projection model of the camera. $\theta_i$ stands for the light's incidence angle with respect to the surface normal $\mathbf{n}_i$, such that, $\cos\theta_i = \mathbf{l}_i \cdot \mathbf{n}_i$. $\rho_i$ represents the albedo of the surface at that point.
$g$ denotes the gain applied by the camera and $\gamma$ is the gamma correction commonly applied by cameras to adapt images to human perception. The resulting $\mathcal{I}(d_i, \rho_i, g)$ is the color captured by the camera.

Our model assumes Lambertian reflections, meaning that the light hitting the surface is scattered equally in all directions. The percentage of reflected light is known as albedo. Specular reflections, which are prevalent in endoscopic images, are not captured by this model but we will consider them in a specific loss that we describe in Section~\ref{sec:losses}.

\subsubsection{Calibration.} Each endoscope has different geometric and photometric parameters, the former affecting the inverse project model $\pi^{-1}$ and the latter impacting both the light position $\mathbf{x}_l$ and spread $R$.
We can estimate these parameters for a particular endoscope by minimizing the reprojection and photometric errors on images of a calibration target, similar to \citep{endommaper, batlle2022photometric}. In our case, the auto-gain values of the endoscope are not known, so radiance measurements of the camera are unitless. Thus, we arbitrarily set $g = 1$, $\sigma_0 = 1$ and obtain up-to-scale reconstructions. Our calibration errors are between $\pm 3$ gray levels.

\subsection{Self-supervision losses}
\label{sec:losses}

Formally, the network of Figure~\ref{fig:difrend} takes as input an image ${I} \in [0, 1]^{w \times h \times 3}$, estimates a depth map $\widehat{d} \in (0, \infty)^{w \times h}$ and an albedo map $\widehat{\rho} \in [0, 1]^{w \times h \times 2}$ . It infers normals $\mathbf{\widehat{{n}}}$ from $\widehat{d}$, and uses $\widehat{d}$, $\mathbf{\widehat{{n}}}$, and $\rho$ to render an image $\widehat{I} \in [0, 1]^{w \times h \times 3}$  that should be as similar as possible to  ${I}$. 
To train this network, we minimize a loss
\begin{equation}
    \mathcal{L} = \mathcal{L}_{p} + \lambda_s \mathcal{L}_{s} + \lambda_{sp} \mathcal{L}_{sp} \; ,
    \label{eq:loss}
\end{equation}
where $\lambda_{s}$ and $\lambda_{sp}$ are scalar weights and $\mathcal{L}_p$,  $\mathcal{L}_s$, and $\mathcal{L}_{sp}$ are the loss terms described below.

$\mathcal{L}_p$ is a photometric loss and we take it to be the squared $L_2$  distance between the original image $I$ and the rendered one $\hat{I}$. Note that because our rendering model is fully differentiable, we can perform end-to-end training.

\begin{equation}
    \mathcal{L}_{p} = \sum_{i \in \Omega} ({I}_i -{\widehat{I}}_i)^2
    \text{,} \quad \text{ where } \quad
    \widehat{I}_i = \mathcal{I}(i, \widehat{d}_i, \widehat{\rho}_i, g)
\end{equation}

%


As in \citep{Godard2018}, $\mathcal{L}_{s}$ is a regularization term that minimizes depth gradients except in areas of high color gradients, that may correspond to depth discontinuities. We write
\begin{equation}
     \mathcal{L}_{s} =|\partial_x \widehat{d}|  e^{-|\partial_x {I}|} +|\partial_y \widehat{d}|  e^{-|\partial_y {I}|}
\end{equation}

Finally, recall that we made a Lambertian assumption in Eq.~\ref{eq:im}, which prevents us to account properly for specular reflections and the overexposed pixels they produce. This is a potential source of error and fails to exploit the very useful information that specularities provide about normals. To remedy this, we introduce specular loss $\mathcal{L}_{sp}$. Given image location $i$, the corresponding  direction $\mathbf{l}_i$ from the surface to the light source and the normal of a the surface $\mathbf{\widehat{{n}}}$, the law of reflection states that
\begin{equation}
    \mathbf{s}_i = \mathbf{l}_i - 2\mathbf{\hat{n}}_i \left( \mathbf{\hat{n}}_i \cdot \mathbf{l}_i \right)
\end{equation}
is the specularly reflected direction. Hence, we take our specular loss term to be
\begin{equation}
    \mathcal{L}_{sp} = \sum_{i\in\Omega}\left(m_i\left( \mathbf{s}_i \cdot ( - \mathbf{r}_i) - 1 \right)\right)^2~,
\end{equation}
\begin{equation}
        m_i = \left\{\begin{matrix}
        1 & I_i > th\\ 
        0 & \text{otherwise}
        \end{matrix}\right. \nonumber
\end{equation}
which minimizes the discrepancy between the expected specular reflection $\mathbf{s}_i$ and the actual direction $(-\mathbf{r}_i)$ where the camera observes the reflection, resulting in pixel with high intensity $th = 0.98$.

Our method takes a single image as input, which makes 3D shape recovery solely from pixel colors an underconstrained problem. According to Eq.~\ref{eq:im},  a change in the brightness of a pixel can be due to changes in depth, albedo, camera exposure or surface normal. For example, if a given pixel is very bright, it can be because the pixel is close to the camera/light; the surface has a different albedo, resulting in more light being reflected; the surface normal is aligned to the light/camera, which increases the reflected light; the camera exposure and digital gain have been increased, which impacts brightness values in the whole image. Given the albedo at each surface point and the camera auto-gain, we could resolve these ambiguities. However, in medical endoscopy, true albedos are unknown, and auto-gain is not provided by the hardware manufacturer. 

\subsubsection{Albedo constancy.}
We observe that endoscopy images exhibit a limited range of colors, with brighter tones being present in close areas and darker tones in deeper regions. Consequently, we hypothesize a significant correlation between albedo and the chromatic attributes, namely Hue and Saturation, in the HSV color space, as well as between depth and the Value Channel. In this way, we constrain the palette of colors that can be explained by the albedo decoder and we enhance the disentanglement between depth and albedo by setting $V=100$ for all albedo values. Hence, to predict the albedo map $\widehat{\rho}$, our network predicts just two channels per pixel, for Hue and Saturation, and assumes Value to be one to convert to the RGB space, in which the loss is formulated. 





\subsection{Network architecture}
\label{sec:network-architecture}

\begin{figure}[h!]
\centering
  \includegraphics[width=0.4\textwidth]{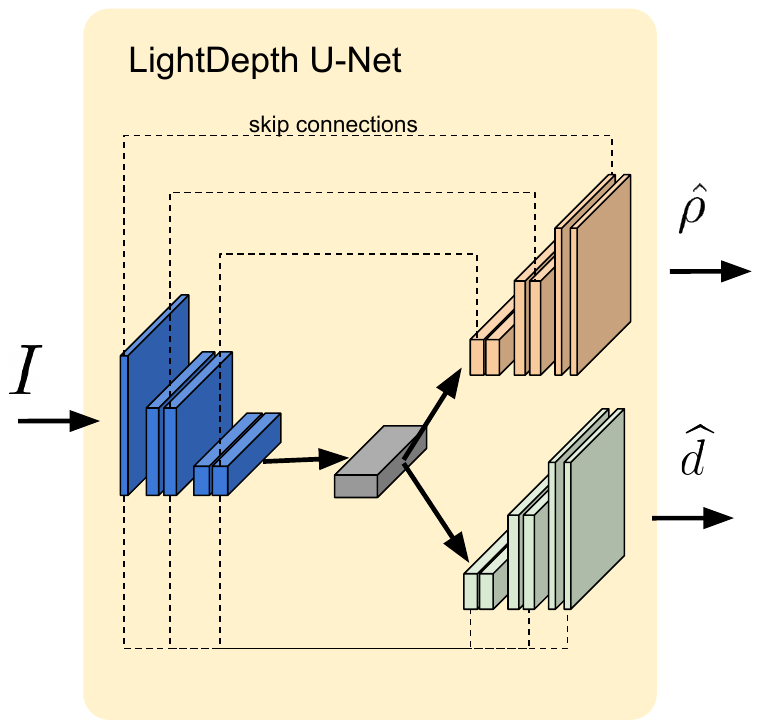}
\caption{LightDepth U-Net is based on a standard U-Net~\citep{Ronneberger2015} with two decoding branches.}
\label{fig:network1}
\end{figure}

\begin{figure}[h!]
  \includegraphics[width=\textwidth]{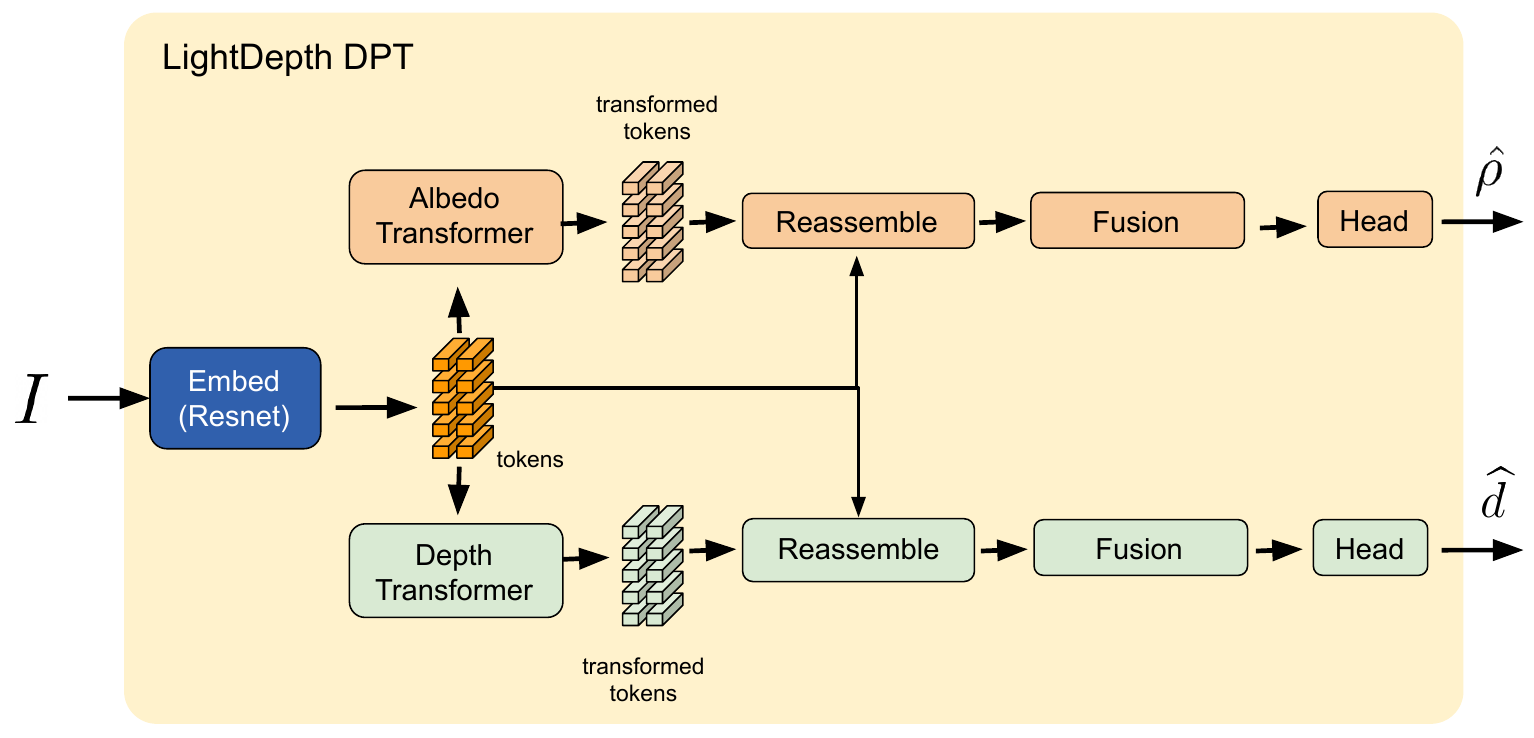}
\caption{LightDepth DPT is based on the DPT-Hybrid architecture~\citep{Ranftl_2021_ICCV}, with a second decoder branch added for the albedo.}
\label{fig:network2}
\end{figure}

Our network outputs depth and albedo maps. In Figure~\ref{fig:network1} and Figure~\ref{fig:network2}, we provide a more detailed depiction of our encoder-decoder architecture. We have tested two different versions. The first one is a U-net with two decoders and skip connections, with a ResNet18 \citep{He_2016_CVPR} serving as the backbone, initialized with the weights from ImageNet \citep{deng2009imagenet}.  Our decoders design is inspired by ~\citep{Godard2018}, our albedo decoder uses sigmoid activation function and our depth decoder ELU+1 activation function after the last convolution. The second one relies on visual transformers for depth estimation \citep{Ranftl_2021_ICCV}, adding a branch for the prediction of albedo decoder. For the depth estimation, we initialize the encoder and depth decoder with DPT Hybrid weights. For albedo estimation, we train the albedo decoder from scratch. In our pipeline, we use the half of resolution than the original images for training, upsampling the outputs with bilinear interpolation. As a backbone, we use a Resnet-50 (DPT-Hybrid) and two decoders that reassemble the tokens and apply attention heads. 

In both versions, to compute the normals at any given pixel, we use a convolution kernel with six-neighborhood (N, NE, E, S, SW, and W) in the depth map. We define six triangles using the central pixel as reference, with each triangle having its own normal. The normal of the central pixel is computed as the average of the normals of the triangles weighted by their area. The use of six neighbors lets us reuse triangles during the convolution pass to speed up computation.


\section{Results}\label{sec:experiments}
\subsection{Datasets}\label{sec:setup} We evaluated LightDepth and relevant baselines on three endoscopy datasets: An in-house \emph{synthetic colon}, \emph{C3VD}~\citep{bobrow2022}, and \emph{EndoMapper}~\citep{endommaper}. With these, we can show quantitative and qualitative results with several levels of realism.

\subsubsection{Synthetic colon.} 
\begin{figure}[!htb]
  \centering
  \includegraphics[width=\linewidth]{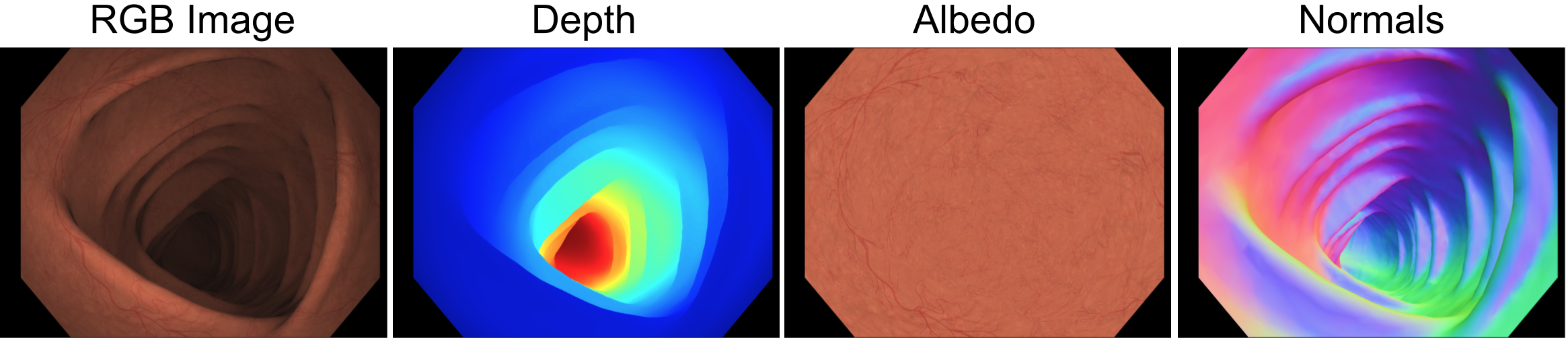}
  \caption{Samples from in-house Fish-eye lambertial synthetic dataset. }
  \label{fig:synthetic}
\end{figure}

We simulate a real Olympus CF-H190L endoscope consisting on a fish-eye camera and a spot-light source, both calibrated as in \citet{endommaper}. This is in contrast to other synthetic datasets that simulate arbitrary camera and illumination configurations, typically pinhole cameras with no or arbitrary distortion and ideal light sources with no radial falloff \citep{rau2019implicit, zhang2020template, ozyoruk2021endoslam, rau2022bimodal}. We rendered the images using ray-casting techniques, in which the colon's geometry and albedo are defined by a triangle mesh obtained from a CT scan of a real colon \citep{incetan2021vr}. We ignore global illumination effects and assume Lambertianity, so there are no specular reflections. The influence of these two effects will be assessed in the two other datasets. Our synthetic data is hence composed by 1620 fish-eye RGB frames annotated with per-pixel albedo, depth and normals. We split it into 1168 images for training and 452 images for test. 
Figure \ref{fig:synthetic} depicts samples from the synthetic dataset, which contains ground truth data on depth, albedo, and normals. 

\subsubsection{Colonoscopy 3D Video Dataset}
\textbf{C3VD}~\citep{bobrow2022} contains real images recorded in a phantom with ground-truth depth. The images have been captured by a real Olympus CF-HQ190L endoscope in a phantom silicone model of a human colon. The data is annotated with ground-truth depth and normals by applying 2D-3D registration of the 3D phantom models. The authors claim that the silicone material is opaque, hence we can assume that the only light source available is in the endoscope. Finally, it includes a geometrical calibration based on the Scaramuzza model \citep{scaramuzza2006}. C3VD provides a good compromise between realism (real endoscope, global illumination effects and specular highlights) and ground-truth labels for quantitative evaluation. Of the 10,088 images available, we use 7,200 for training and 2,888 for testing.
Table \ref{c3vdsplit2} shows which sections of the C3VD were used for training / testing. We split into sections to ensure a fair comparison along the dataset. Regarding real endoscopy images, we use with the sequence 051, 009 and 058 of the EndoMapper dataset. 

\begin{table}[h!]

\centering
    \resizebox{0.5\textwidth}{!}{
\begin{tabular}{|c|c|c|c|c|}
\hline Model  & Texture & Video  & Frames & Stage   \\ \hline
Cecum & 1 & b&765& Train  \\
Cecum & 2 & b&1120& Train  \\
Cecum & 2 & c&595& Train  \\
Cecum & 4 & a&465&   Train\\
Cecum &4 & b&425& Train  \\

Sigmoid Colon &1 & a&800& Train  \\
Sigmoid Colon &2 & a&513& Train  \\
Sigmoid Colon &3 & b&536& Train  \\
Transcending Colon &1 & a&61& Train  \\
Transcending Colon &1 & b&700& Train  \\
Transcending Colon &2 & b&102& Train  \\
Transcending Colon &4 & b&595& Train  \\
Descending Colon &4 & a&74& Train  \\

\hline
Cecum & 1 & a&275& Test  \\
Cecum & 2 & a&370& Test  \\
Cecum & 3 & a&730& Test  \\
Descending Colon &4 & a&74& Test  \\
Sigmoid Colon &3 & a&610& Test  \\
Transcending Colon &2 & a&194& Test  \\
Transcending Colon &3 & a&250& Test  \\
Transcending Colon &4 & a&382& Test  \\
\hline
    \end{tabular}
    }
    \caption{Dataset Split for C3VD }
\label{c3vdsplit2}
\end{table}
\subsubsection{EndoMapper}
\textbf{EndoMapper}~\citep{endommaper} provides the most challenging data, as it contains real colonoscopy and gastroscopy procedures inside the human body, performed by endoscopists on a day-to-day basis. Here we find real textures such as veins, blood and dirt, and other effects such as blur, water and frames very close or even hitting the mucosa. Foam and bubbles are indeed very common in endoscopy images and are usually ignored. LightDepth is capable of disentangling these as part of the albedo and not of the depth. Before processing the dataset, we perform a manual inspection of the selected sequences and we eliminate occluded and excessively blurred frames.

Finally, we train in three procedures, consisting of two colonoscopies and one gastroscopy. There are a total of 24,444, 23,456 and 3,032 frames, respectively. 

\subsection{Metrics, baselines, and training details}

We report results using a median-based scale alignment for all methods, even those supervised with real-scale depth, for fairness.
In our experiments, we compare against models that use depth supervision and multi-view self-supervision. For depth supervision, we use two different architectures, U-Net with L1 loss as a representative of convolutional architectures and DPT-Hybrid~\citep{Ranftl_2021_ICCV} as a state-of-the-art representative of transformer-based models, learning inverse depth with an scale invariant loss.

For a fair comparison, we also evaluate our LightDepth using the same U-Net and DPT architectures.
The U-Net is pre-trained on ImageNet dataset \citep{deng2009imagenet}. For DPT, we initialize with the author-provided weights for encoder and depth decoder. The albedo decoder is trained from scratch.
During training, we select a smoothing weight $\lambda_s = 0.1$ in Eq.~\ref{eq:loss} and a learning rate of $10^{-4}$ for the  Adam optimizer. In the synthetic dataset, we trained our network with $\lambda_{sp}$ = 0, as synthetic dataset has no specular reflections. In C3VD and EndoMapper, we use $\lambda_{sp} = 1$.
\subsubsection{Test-Time Refinement (TTR)}
As our LightDepth enables single-view self-supervision, we can continuously refine the depth predictions online, obtaining much more accurate reconstructions. In the results denoted as ``(TTR)'', we perform online test-time refinement for each test image separately during $N = 20$ optimization steps, using the loss $\mathcal{L}$ in Equation \ref{eq:loss}, as in training time. To mitigate the risk of catastrophic forgetting, we load again the original model trained in the train split after TTR for each image.
 
\begin{figure}
  \includegraphics[width=\linewidth]{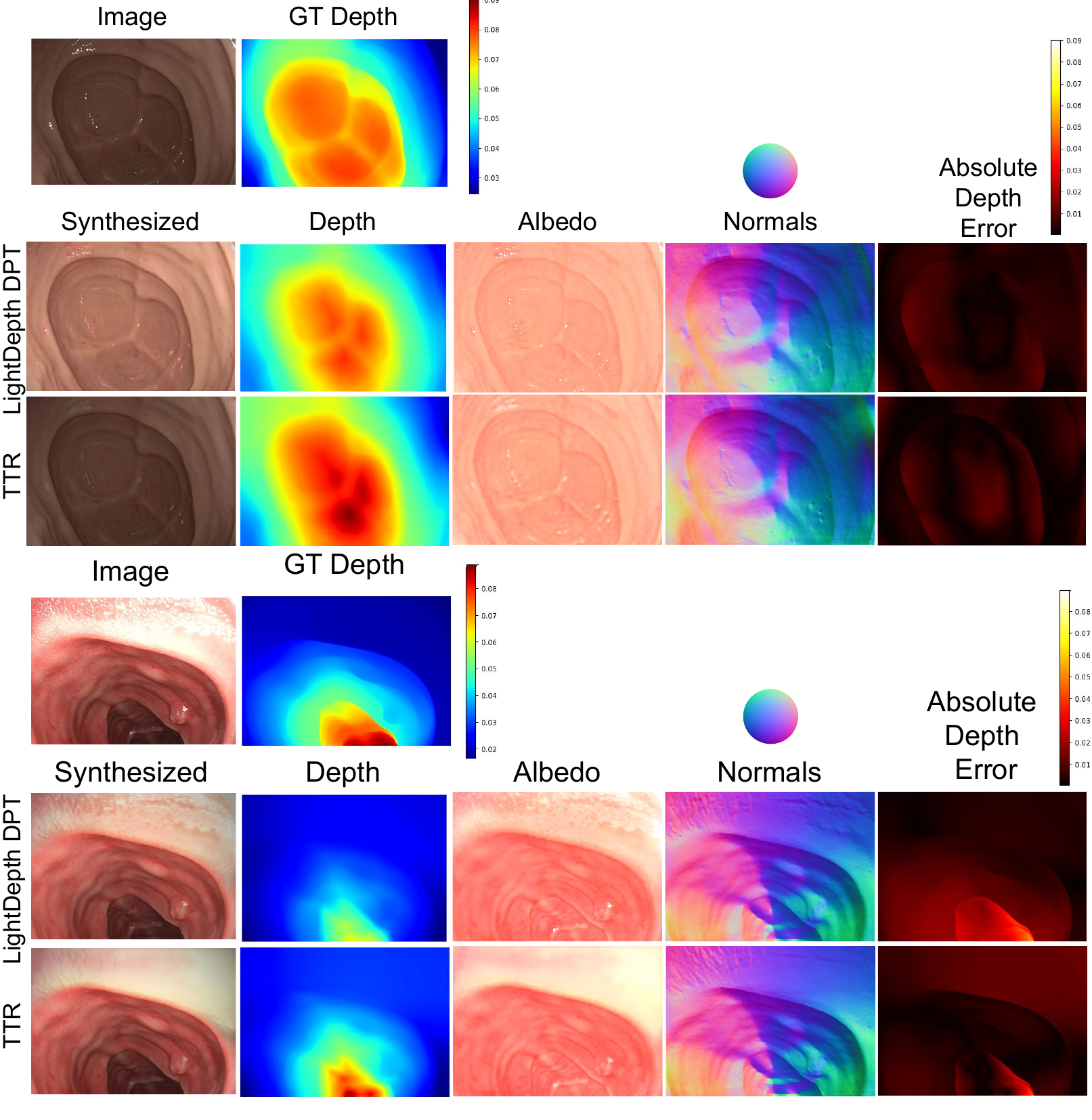}
\caption[LightDepth and LightDepth TTR on C3VD.]{LightDepth and LightDepth TTR on C3VD. Our light decline captures the correct shape of the cecum in the first image and the shape of the polyp in the second. Note how the estimates of normals and albedo are similar before and after TTR. By optimising depth by reducing illumination, DepthLight achieves a darker appearance and improvements in depth estimation.}
\label{fig:TTR}
\end{figure}

\begin{figure*}
  \includegraphics[width=\textwidth]{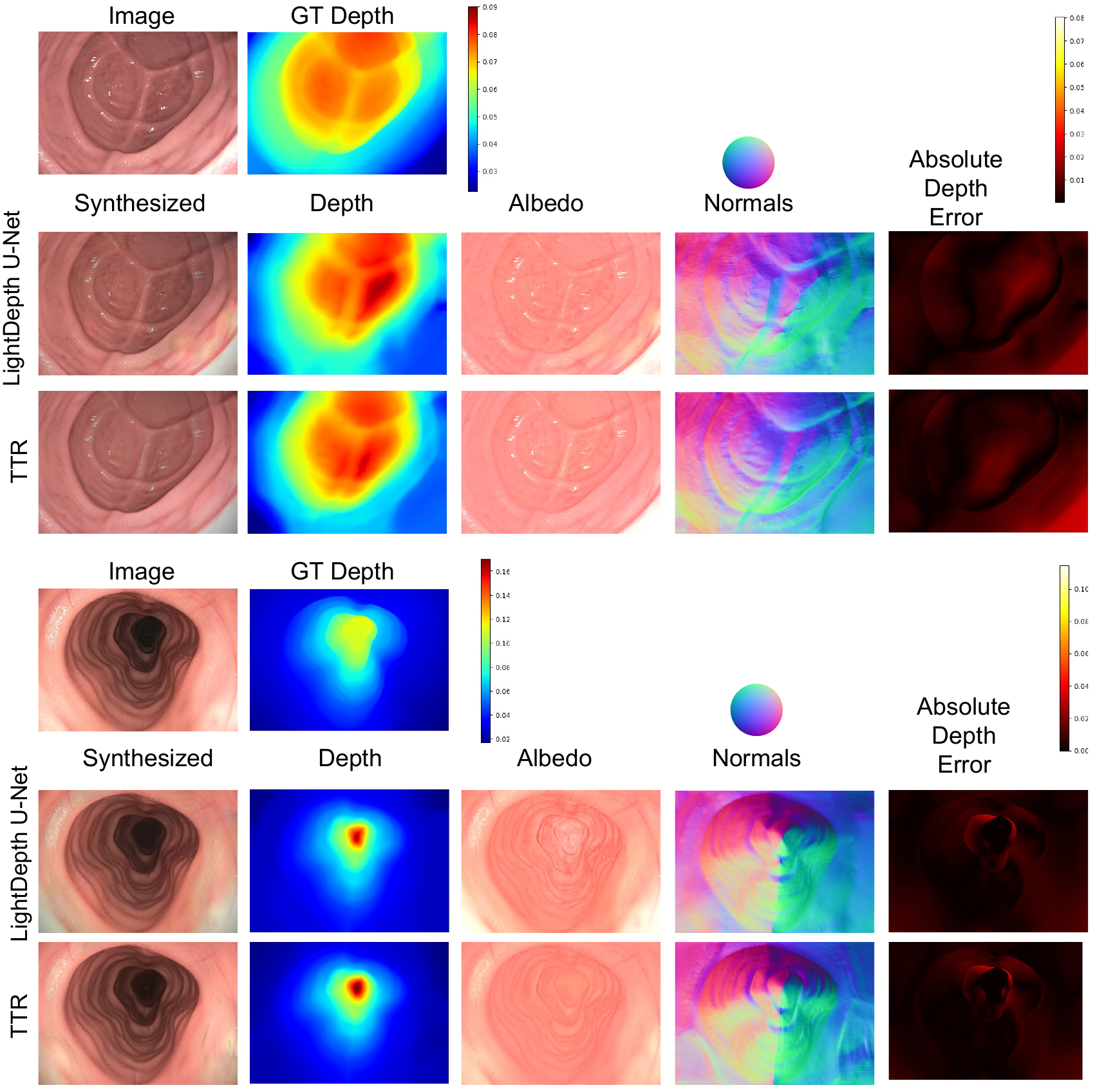}
\caption[Quantitative examples of
LightDepth U-Net in C3VD.]{LightDepth and LightDepth TTR on C3VD. Quantitative examples of LightDepth U-Net in C3VD.}
\label{fig:additionalphantom}
\end{figure*}

Note in Table~\ref{tab:metricsGT} how TTR improves significantly the metrics with respect to LightDepth without TTR for U-Net and DPT architectures. Remarkably, observe how TTR even outperforms the metrics achieved by Depth GT supervision. Figure \ref{fig:TTR} and Figure \ref{fig:additionalphantom} show the improvement given by TTR in the network prediction of depth, normals and albedo and overall in the 3D reconstruction.
Inference time is $\sim 5$ms for LightDepth U-Net and $\sim 22$~ms for LightDepth DPT on a NVIDIA GeForce RTX 3090. We can do TTR in $\sim 90$~ms per optimization step in U-Net and $\sim 190$~ms in DPT.




\subsection{Quantitative and qualitative results on synthetic and phantom}
\label{sec:quant}

\subsubsection{Synthetic colon.}

\begin{table*}[ht!]
    \resizebox{\linewidth}{!}{
        \begin{tabular}{cccccccccccccc}
        & &  & \multicolumn{9}{c}{\textbf{Depth [mm]}} & \textbf{Normals [$^\circ$]} \\
         Architecture & Backbone & Supervision &  MAE $\downarrow$ & MedAE $\downarrow$ & RMSE $\downarrow$ & RMSE\textsubscript{log} $\downarrow$ & Abs\textsubscript{Rel} $\downarrow$ & Sq\textsubscript{Rel} $\downarrow$ & $\delta < 1.25$ $\uparrow$ & $\delta < 1.25^2$ $\uparrow$ & $\delta < 1.25^3$ $\uparrow$ & MAE $\downarrow$ \\ \hline
        U-Net & ResNet18 & Depth GT &   \textbf{4.37} & 2.99 & \textbf{6.38} & \textbf{0.1251} & 0.0965 & \textbf{0.0008} & 0.9057 & \textbf{0.9931} & \textbf{0.9997} & 25.1 \\
        LightDepth U-Net & ResNet18 & Light &  4.76 & \textbf{2.47} & 8.60 & 0.1375 & \textbf{0.0903} & 0.0011 & \textbf{0.9180} & 0.9820 & 0.9935 & \textbf{15.2} \\ \hline       
        \end{tabular}
     }
    \caption{Depth and normal metrics for several architectures and supervision modes. Best results per dataset are boldfaced, second best underlined.}
    \label{tab:metricsGTsynth}
\end{table*}

\begin{figure*}
  \includegraphics[width=\textwidth]{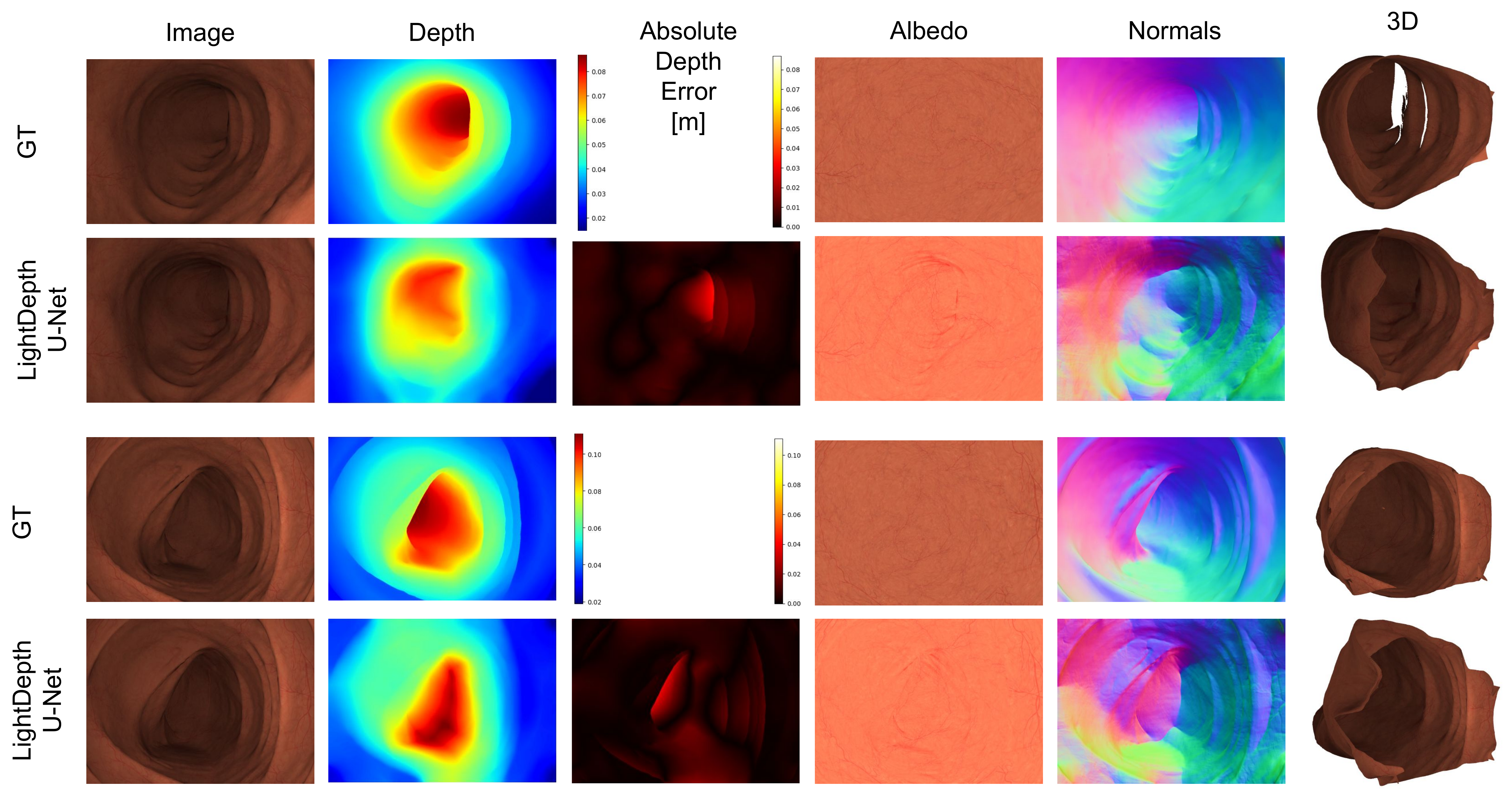}
\caption{Qualitative examples of LightDepth in Synthetic dataset.}
\label{fig:additionalsynth}
\end{figure*}

In Table~\ref{tab:metricsGTsynth}, we report depth and normal metrics for a U-Net supervised with Depth GT, and our self-supervised LightDepth U-Net architecture. Observe that the metrics are similar. This is notable, as self-supervision is consistently reported in the literature to underperform with respect to depth supervision, and suggests that illumination decline provides a very strong self-supervisory signal in endoscopies, which our experiments in the other two datasets confirm.
Furthermore, light self-supervision outperforms Depth GT supervision in MedAE and $\delta < 1.25$, which means that most of the error distribution is lower for light self-supervision and only a small fraction of large errors are better with depth supervision. We observed that it is in far and dark areas where light self-supervision is weaker and this produces a higher depth MAE and RMSE. Observe the significantly lower  error in normal with our light self-supervision, due to the lower errors in most pixels.

\subsubsection{C3VD Phantom.}
\begin{table*}[ht!]
    \resizebox{\textwidth}{!}{
        \begin{tabular}{lccccccccccccc}
         &  &  & \multicolumn{9}{c}{\textbf{Depth [mm]}} & \textbf{Normals [$^\circ$]} \\
         Architecture & Backbone & Supervision &  MAE $\downarrow$ & MedAE $\downarrow$ & RMSE $\downarrow$ & RMSE\textsubscript{log} $\downarrow$ & Abs\textsubscript{Rel} $\downarrow$ & Sq\textsubscript{Rel} $\downarrow$ & $\delta < 1.25$ $\uparrow$ & $\delta < 1.25^2$ $\uparrow$ & $\delta < 1.25^3$ $\uparrow$ & MAE $\downarrow$ \\ \hline
        U-Net & ResNet18 & Depth GT & 4.15 & 3.29 & 5.52 & 0.1139 & 0.0902 & 0.0007 & 0.9172 & {\ul0.9943} & \textbf{0.9994} & 26.5 \\ 
        DPT-Hybrid & ResNet50 & Depth GT  & \textbf{3.22} & 2.77 & \textbf{4.10} & \textbf{0.0860} & \textbf{0.0699} & \textbf{0.0004} & \textbf{0.9640} & 0.9865 & 0.9913 & \textbf{15.1} \\ 
        Monodepth2 &ResNet50& Multi-View & 14.27 & 9.59 & 18.64 & 0.3921 & 0.2971 & 0.0070 & 0.4897 & 0.7313 & 0.8611 & 43.6 \\
        CADepth  & ResNet18 & Multi-View & 52.35 & 17.04 & 87.43 & 0.9144 & 1.1916 & 0.2650 & 0.3664 & 0.5653 & 0.6679 & 67.2 \\
        XDCycleGAN  & ResNet &  Cycle & 17.16 & 11.91 & 22.43 & 0.4953 & 0.3616 & 0.0105 & 0.4291 & 0.6615 & 0.7910 & 64.4 \\
        LightDepth U-Net  & ResNet18 & Light & 4.37 & 2.92 & 6.31 & 0.1183 & 0.0856 & 0.0007 & 0.9315 & 0.9934 & \textbf{0.9994} & 24.0 \\
        LightDepth DPT & ResNet50 &Light&3.94 &2.67 &5.60&0.1080&0.08046&0.0006&0.9476&0.9965&	\textbf{0.9994} &{ \ul 21.3}\\
        LightDepth U-Net  & ResNet18& Light (TTR) & 3.72 & {\ul 2.59} & 5.43 & {\ul 0.1060} & {\ul 0.0770} & {\ul 0.0005} & {\ul 0.9505} & \textbf{0.9971} & \textbf{0.9994} &  23.5 \\
        LightDepth DPT & ResNet50 & Light (TTR) & {\ul 3.70}&\textbf{2.58}& {\ul 5.27}	&0.1073&	0.0780& {\ul 0.0005}	& 0.9525&	0.9961 &{ \ul 0.9992} &	22.5\\\hline
        \end{tabular}
     }
    \caption{Depth and normal metrics for several architectures and supervision modes. Best results per dataset are boldfaced, second best underlined.}
    \label{tab:metricsGT}
\end{table*}

We report depth and normal metrics on the real phantom images of the C3VD dataset in Table~\ref{tab:metricsGT}. Our self-supervised architectures LightDepth U-Net and LightDepth DPT with TTR outperform supervision with Depth GT in MedAE, while the rest of the metrics are very close. As in the case of the synthetic dataset, this is a remarkable result because self-supervised architectures typically lag behind supervised ones in single view depth estimation. 
The fact that LightDepth MedAE is better and RMSE is worse suggests that our errors are better in most of the distribution, and there are a few regions with large errors where Depth GT supervision is able to offer an advantage. Table \ref{tab:metricsGTphoto} details metrics on the quality of the rendered image, which suggest the strength of the self-supervision signal. Observe the improvement of this metrics for the TTR case. 

In Table \ref{tab:metricsGT}, observe that the multi-view self-supervised baselines, Monodepth2~\citep{Godard2018} and CADepth~\citep{yan2021channel}, have a poor performance in our the phantom colon, worse in comparison than results in other datasets. This could be due to the weak textures and changing lighting in the colonoscopy images, resulting in noisy estimations for relative motion and uninformative photometric residuals. Being single-image, our approach is impervious to such difficulties.

\begin{table}[h!]
\centering
    \resizebox{0.5\textwidth}{!}{
        \begin{tabular}{clcccc}
        Dataset & Architecture &  Supervision & SSIM $\uparrow$ & MAE $\downarrow$ \\ \hline
        \multirow{1}{*}{Synthetic} 
         & LightDepth U-Net &Light& {0.9901} & {0.0192} \\ \hline
        \multirow{4}{*}{C3VD} 
         &  LightDepth U-Net &  Light & 0.9765 & 0.0657 \\
         & LightDepth DPT &Light& 0.8873	&0.0599&\\
         & LightDepth U-Net & Light (TTR) & \textbf{0.9811} & \textbf{0.0276} \\
         & LightDepth DPT & Light (TTR) & 0.8977	 &  0.0329
         \\
         \hline
         
        \end{tabular}
     }
     \caption{SSIM and MAE for rendered images in C3VD. Test-time refinement (TTR) gives a substantial improvement.}
    \label{tab:metricsGTphoto}
\end{table}

\subsubsection{Domain shift} As synthetic-to-real is common in endoscopies to address the lack of ground-truth depth for supervision, we also evaluated XDCycleGAN \citep{mathew2020augmenting} as a baseline. Note that the domain shift is still affecting the results. Our single-view LightDepth self-supervision enables training in the target domain, and hence removes completely the domain shift, achangedchieving significantly lower errors. 

\begin{table}[h!]
\centering
\resizebox{0.7\textwidth}{!}{%
\begin{tabular}{cc|l|ccc|c}
 \multicolumn{2}{c}{Dataset}            & & \multicolumn{3}{c}{Depth [mm]}                           & Normals [$^\circ$]    \\
Train                  & Test                   & {Supervision} & MAE  $\downarrow$ & MedAE $\downarrow$ & RMSE$\downarrow$ & MAE $\downarrow$ \\ \hline
\multirow{2}{*}{Synt.} & \multirow{2}{*}{Synt.} & {Depth GT}       & 4.37    & 2.99     & 6.38    & 25.1   \\
                       &                        & {Light}       & 4.76    & 2.47      & 8.60    & 15.2    \\\hline
\multirow{4}{*}{Synt.} & \multirow{4}{*}{C3VD}  & {Depth GT}       &  9.44            & 5.79               & 12.83            & 73.7             \\ 
                       &                        & {Light}        & 5.09            & 3.51      & 7.14    & 27.7   \\ 
                       &                        & {Depth GT (TTR)}  & 4.96            & 3.14 & 7.11 & 25.4 \\
                       &                        & {Light (TTR)}  & {\ul 3.80}           & \textbf{2.51} & 5.54 & { \ul 23.6} \\\hline
\multirow{3}{*}{C3VD}  & \multirow{3}{*}{C3VD}  & {Depth GT}        & { 4.15}    & 3.29      & {\ul 5.52}    & 26.5 \\
                       &                        & {Light}        & 4.37    & { 2.92}      & 6.31    & 24.0 \\
                       &                        & {Light (TTR)}        & \textbf{3.72}    & { \ul 2.59}      & \textbf{5.43}    & \textbf{23.5} \\\hline

\end{tabular}%
}
\caption[Synthetic-to-real domain shift evaluation.]{Synthetic-to-real domain shift. Best results in C3VD test set are boldfaced, second best are underlined. Note the domain shift effect between Synt. and C3VD test data in the bigger errors, and how TTR removes the domain shift effect completely. Notably, our LightDepth TTR delivers similar errors than the models without domain shift, trained in C3VD.}
\label{tab:transfer}
\end{table}

Table \ref{tab:transfer} elaborates further on domain shift by showing depth and normals metrics for a U-Net architecture in these cases. Specifically, we trained a U-Net model with Depth GT supervision and light self-supervision in our synthetic dataset and evaluated their performance in the synthetic and C3VD test sets. Observe how the domain shift affects all metrics significantly. Interestingly, the model trained with light self-supervision and without TTR generalizes significantly better to the C3VD data, as our LightDepth self-supervised model is closer to the physical phenomena than Depth GT supervision. Again, note that single-view self-supervision removes completely the domain shift effect, as models can be trained directly in the target domain. Very remarkably, the performance of our models with domain shift after TTR matches the performance of the models without domain shift.

\begin{figure}
\centering
\begin{tabular}{c  c} 
 Method  & MAE [$^\circ$] \\
 \hline
 U-Net& 16.24\\
 TFtN~\citep{fan2021three}	& 3.89\\
 Open3D~\citep{Zhou2018Open3D} & 1.67 \\
 In-house & \textbf{1.32}\\
 \hline
\end{tabular}
\caption{Normal's MAE for baseline methods.} 
\label{tab:ablationNormals}

\end{figure}

Figure \ref{fig:normals} shows examples of Open3d \citep{Zhou2018Open3D}, in-house, U-Net and TFtN \citep{fan2021three} used in the analysis.

\subsubsection{Normals from depth}
\begin{figure}[h]
\centering
  \includegraphics[width=0.8\textwidth]{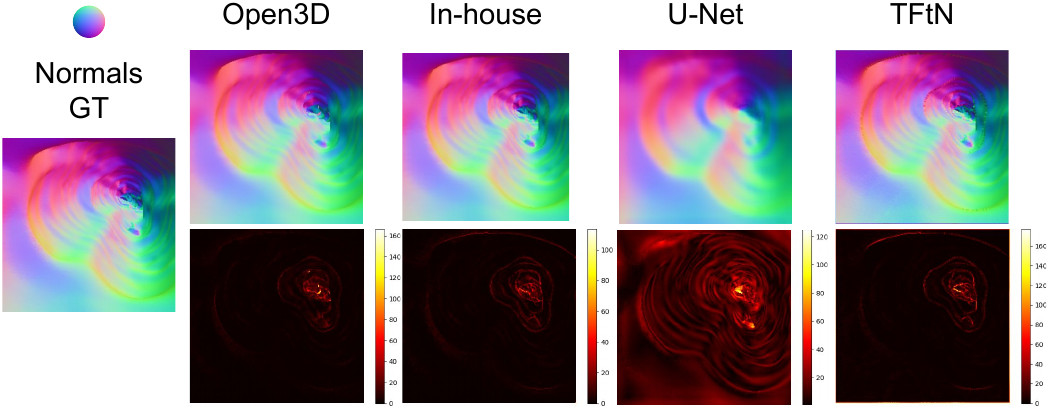}
\caption{Quantitative results of different approaches to obtaining surface normals from a depth map.}
\label{fig:normals}
\end{figure}

The literature details different manners to obtain surface normals from a depth map, e.g.,~ \citep{fan2021three,boulch2016deep}. Table \ref{tab:ablationNormals} shows a MAE analysis of the most promising ones in C3VD. Specifically, we evaluate four methods: a U-net trained to regress normals from depth, the recent TFtN method~\citep{fan2021three}, the implementation in Open3D~\citep{Zhou2018Open3D} that computes normals from a k-nearest neighbourhood in the point cloud, and an in-house method that uses six-neighbourhood in the image. Figure \ref{fig:normals} shows examples of Open3d \citep{Zhou2018Open3D}, in-house, U-Net and TFtN \citep{fan2021three} used in the analysis.
Our analysis shows that an analytic average in a neighbourhood is significantly better than a U-Net and TFTN, and our in-house method that considers a neighbourhood in the image is slightly better, so this last one was our choice.
\subsubsection{Ablation study on the loss} In Table~\ref{tab:loss_ablation}, we ablate the terms of our loss function. The smoothness prior ($\mathcal{L}_s$ term) is remarkably beneficial for both depth and normal prediction. When we do not take advantage of the information of the specular reflections (no $\mathcal{L}_{sp}$ term), we obtain worse results. Adding this new loss term, we see how all the depth and normal metrics improve, especially in the median error, which outperforms the supervised and now matches that obtained in the simulation experiment. Still, the depth MAE and RMSE are slightly higher than those of the baseline due to the far spurious points.

\begin{table}
\centering
\resizebox{0.8\textwidth}{!}{%
\begin{tabular}{cccccc}
 & \multicolumn{3}{c}{Depth [mm]} & Color & Normals [$^\circ$]    \\
Loss & MAE  $\downarrow$ & MedAE $\downarrow$ & RMSE$\downarrow$ & MAE $\downarrow$ & MAE $\downarrow$\\ \hline
$\mathcal{L}_p$ & 6.05 & 3.93  & 8.79    & 0.0637    & 35.5  \\
$\mathcal{L}_p + \mathcal{L}_s$ & 4.95 & 3.04 & 7.23  & 0.0690 &   24.6  \\
$\mathcal{L}_p + \mathcal{L}_s + \mathcal{L}_{sp}$ & 4.37 &  2.92      & 6.31     & 0.0657  &  24.0 \\
                     \hline
\end{tabular}%
}
\caption{Ablation study of the losses with LightDepth U-Net in C3VD dataset. Observe the improvement given by each term.}
\label{tab:loss_ablation}
\end{table}

\subsection{Qualitative results in real endoscopy}
\begin{figure*}
  \includegraphics[width=\textwidth]{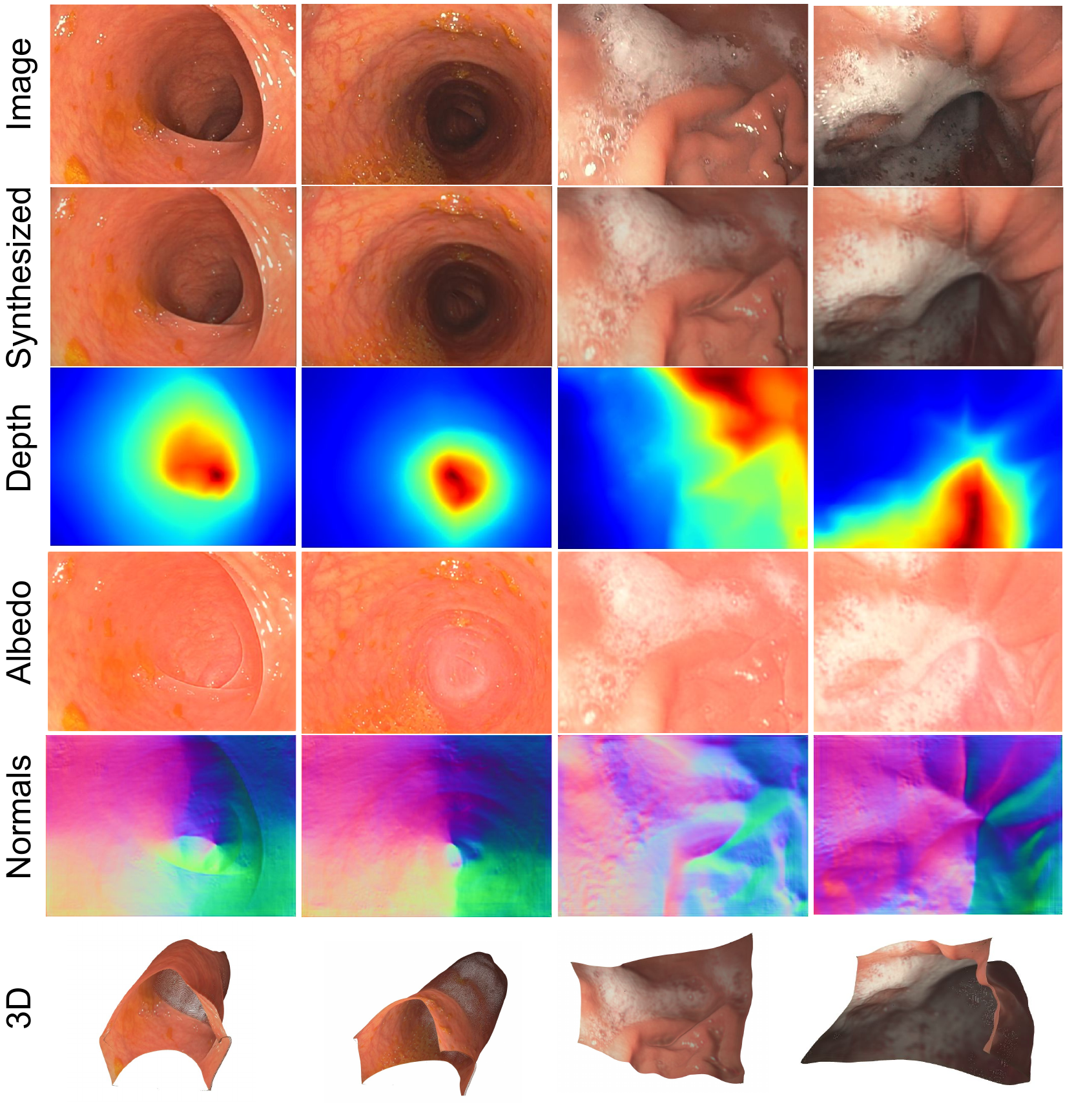}
\caption[Qualitative results on EndoMapper with LightDepth DPT.]{Qualitative results on EndoMapper with LightDepth DPT. In colonoscopies, observe that the normals exhibit a tubular shape specific of the colon. The albedo prediction captures disruptions such as veins, dirt, foam and specularites. Note the influence of light decline in the image and the correlation with the estimated depths.}
\label{fig:ld1}
\end{figure*}
\begin{figure*}
  \includegraphics[width=\textwidth]{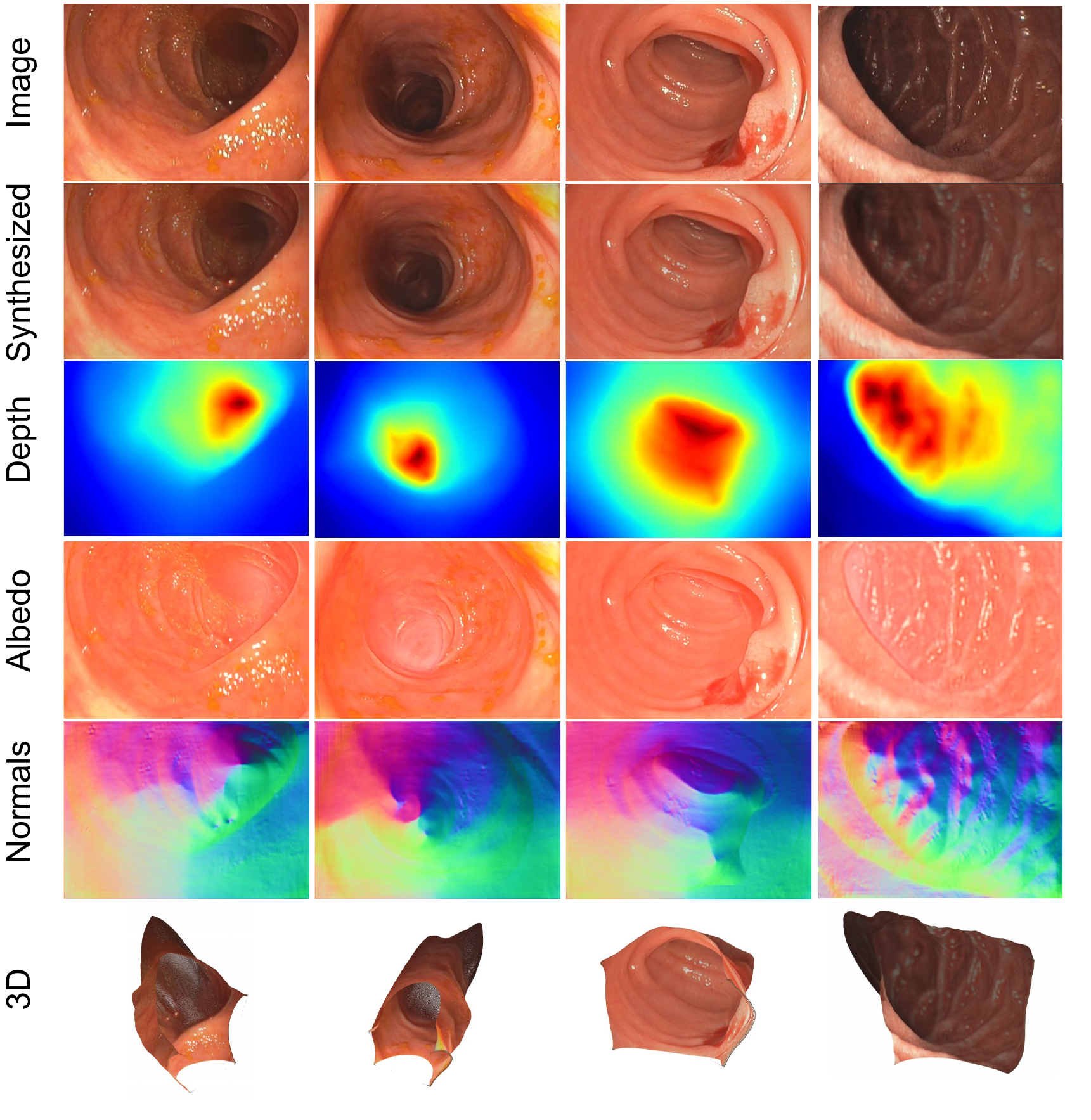}
\caption[Qualitative results on EndoMapper with LightDepth DPT.]{Qualitative results on EndoMapper with LightDepth DPT. The albedo prediction captures blood. Note the influence of light decline in the image and the correlation with the estimated depths.}
\label{fig:ld2}
\end{figure*}

\begin{figure*}
  \includegraphics[width=\textwidth]{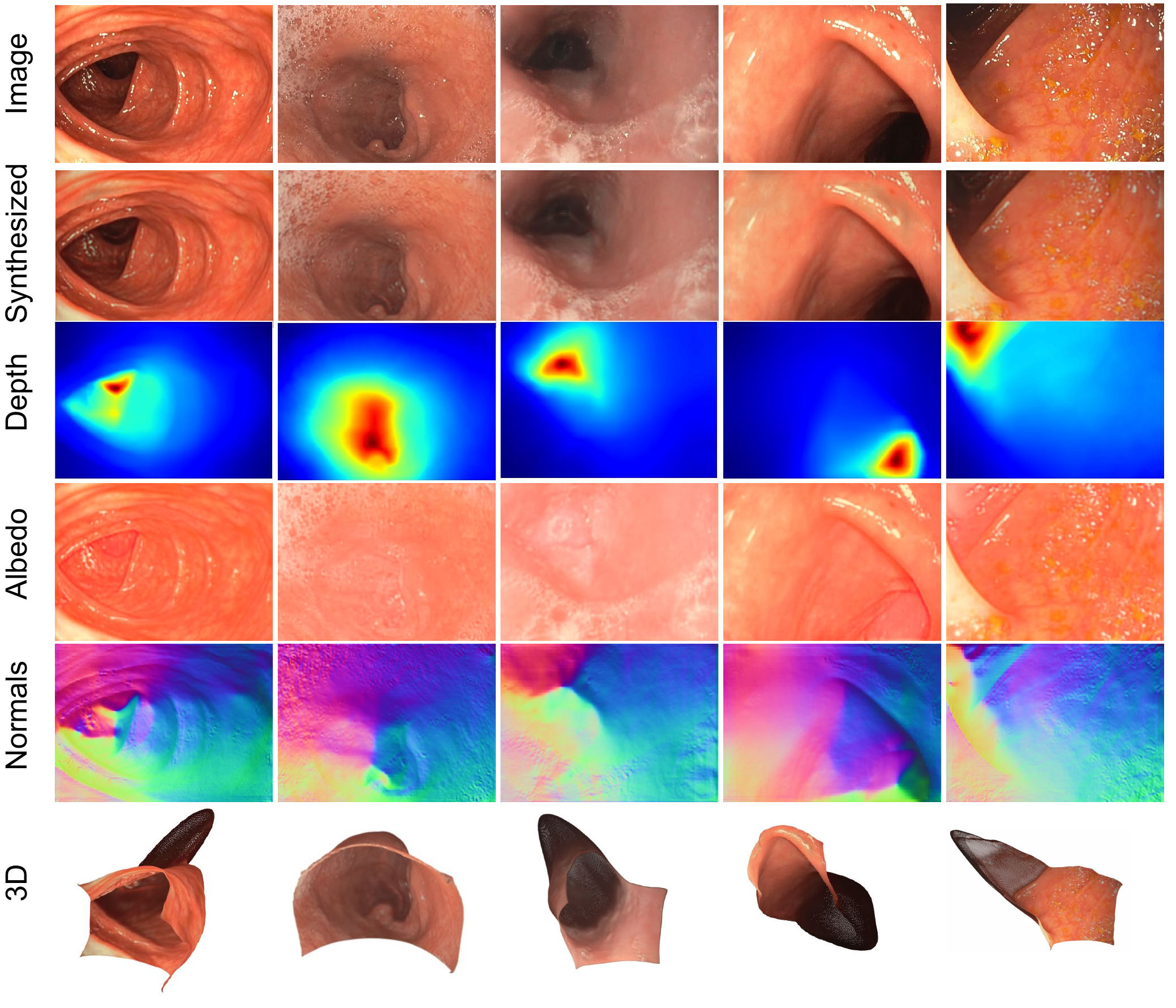}
\caption[Qualitative results on EndoMapper with LightDepth DPT.]{Qualitative results on EndoMapper with LightDepth DPT.}
\label{fig:ld3}
\end{figure*}

We now turn to real images of a human colon from the EndoMapper dataset and present qualitative results in Figure \ref{fig:ld1}, Figure\ref{fig:ld2} and Figure \ref{fig:ld3}. Some details are recovered very accurately, such as the normal maps showing clearly the tubular shape; the depth maps reflecting the discontinuities in the Haustras; the albedos capturing the blood vessels, in particular in the $\text{5}^\text{th}$ column;  and the bubbles and fluids colors in the $\text{6}^\text{th}$ and $\text{7}^\text{th}$ columns, which make the 3d reconstruction of these bubbles and fluids very plausible. 

Unfortunately, there is no ground-truth data available for this dataset, which prevent us from presenting quantitative results, and we do not know of any other dataset with real colonoscopy images that includes ground-truth data. Nevertheless, visual inspection of our results hints that the strengths of our techniques demonstrated quantitatively in Section~\ref{sec:quant} will carry over on truly realistic scenarios like this one. 
\section{Limitations and discussion}
As mentioned in Section \ref{sec:losses}, our depth predictions are up-to-scale. Even if the camera auto-gain were available, the albedo scale may be challenging to learn, so estimating the real scale is not straightforward. In any case, other methods such as multi-view self-supervision or synthetic-to-real cannot guarantee an accurate estimation of the scale either.
We assume that Lambertian reflectance is prevalent in most tissues, and for areas where this does not hold, we use a basic model to capture specularities. Further research could focus on the application of more sophisticated photometric models that cover specularities, e.g., the Phong model.

Thanks to our priors on albedo and depth, we successfully disentangle both factors in our experiments. However, our $V=100$ prior might not hold in areas of clotted blood or with very dark albedos, e.g., because of a disease. These priors might need to be tuned in new application domains for enhanced performance.
Finally, although we demonstrate this technology in the context of endoscopy, its principles are applicable in any setup in which the only light source is close to the target surface and rigidly attached to the camera. In other words, our LightDepth has the potential to open research avenues in many other domains.
\section{Conclusions}In this chapter, we have proposed, for the first time, a single-view self-supervision method for depth learning, which we denote LightDepth, that exploits and is limited to the case of a single spotlight source co-located with a monocular camera, a case that includes, among others, the relevant application of medical endoscopy. As our main contribution, we developed the specific self-supervised learning setup that models the quadratic light decline and enables self-supervised learning. We have implemented two different architectures, a first one based on convolutions and a second one based on transformers, and evaluated their performance against ground-truth supervision, multi-view self-supervision, and domain transfer approaches. Our results show that LightDepth outperforms multi-view self-supervision and synthetic-to-real transfer and matches the performance of fully supervised approaches. Not only that, its training and test-time refinement setup is significantly simpler: LightDepth only requires a reasonable endoscope calibration and does \emph{not} require camera motion estimation nor ground-truth labels nor realistic simulations, all of them challenging in endoscopies. This unlocks, from a practical point of view, relevant potential applications in the medical domain.
\chapter{Conclusions and Future work}
In this thesis, we investigate several aspects related to depth estimation from a single view, presenting a number of contributions. Specifically, our contributions focus on self-supervision and uncertainty quantification.

In the work described in the first chapter, we have made several findings related to uncertainty quantification with Bayesian deep neural networks for supervised learning of single-view depth. Among the most relevant it is the insight that the specific layers in which MC dropout is applied in a Bayesian neural architecture plays a critical role in determining the accuracy of both depth and uncertainty estimates. In our experiments, deep ensembles stand out as the method with the best uncertainty calibration. However, the application of MC dropout, when done in the encoder, manifested a performance that outperformed other variants highlighted in the literature. We observed that aleatoric uncertainty is relatively straightforward to capture. Nonetheless, it is important to highlight the relevance of epistemic uncertainty, which our research identified as a significant part of the total uncertainty. Current scalable methods still have room for improvement, and achieving a reasonable calibration of the uncertainty remains an open goal.

In Chapter \ref{chap:2}, on the uncertain single-view depths in colonoscopies, we conduct research on single-view depth uncertainty quantification, in this case with focus in medical colonoscopy images, bringing a novel perspective to the established literature. Our main motivation is highlighting the importance of both accurate depth and uncertainty estimations in medical imaging. Specifically, in colonoscopy images, depth uncertainty is most prevalent in specular reflections and overexposed regions, dark areas where the color signal is weaker and depth discontinuities such as haustra. 
We demonstrate empirically that the uncertainty captured by deep ensembles behaved coherently when a domain change occurred. This finding has a high relevance, as in the endoscopic domain it is very difficult or impossible to have ground truth annotations, and hence simulation-to-real domain transfer is common. Our research also led to the development of a novel self-supervised method based on an uncertainty-aware teacher-student architecture. 
Our novel architecture incorporates the teacher uncertainty into the learning pipeline. Our experiments show a notable efficiency in reducing depth errors and improving the uncertainty calibration compared to its predecessors. While our method represents a promising advance, its effectiveness is largely dependent on the skill of the teacher model. 
New techniques need to be developed not only to capture uncertainty, but also to be able to apply it to deep learning.

Finally, in the fourth chapter, we propose a new single-view self-supervised learning that models quadratic light attenuation and unlocks single-view self-supervised learning in natural medical images. Our so-called LightDepth method offers an original approach to depth self-supervision. It enables consistent depth estimation without the need for ground truth supervision, based on the use of illumination as a supervisory signal for depth. Our research shows that the illumination-decline self-supervision is significantly superior to multi-view self-supervision, and even matches ground truth supervision.
Moreover, and as a result of using just a single view for self-supervision, a distinct advantage of our method is its ability to refine on-the-fly during testing, which further improves the depth predictions.

Our results indicate that while supervised approaches can often be effective for endoscopic vision, their performance is also limited by the particularities of the use case, specifically the lack of depth sensors in minimally invasive procedures and hence the difficulty of capturing data with depth annotations. We also observed that multi-view self-supervision is limited by the difficulties of estimating relative motion from colonoscopy images, which is also a research challenge in itself.
We believe that our results open a promising path, as our single-view self-supervision overcomes the limitations of both approaches and does only requires RGB data, which does not pose serious difficulties for being acquired . 

\chapter*{Conclusiones y trabajo futuro}
En esta tesis, se investiga varios aspectos relacionados con la estimación de profundidad desde una única vista. Las contribuciones se enfocan en sistemas de aprendizaje autosupervisado y en la cuantificación de incertidumbre en redes neuronales profundas.
En el primer capítulo, hemos aportado varios descubrimientos relativos a la cuantificacion de incertidumbre para la estimación de profundidad con redes bayesianas profundas en un aprendizaje supervisado utilizando una única vista. 
Entre los más relevantes se encuentra la observación de que la aplicación de MC dropout en diferentes capas de la arquitectura de la red tiene un papel determinante respecto a la precisión de la estimación de profundidad e incertidumbre. 
En nuestros experimentos, deep ensembles destaca como el método con mejor calibración de incertidumbre. Sin embargo, la aplicación de MC dropout, cuando se hace en el encoder, ha demostrado un rendimiento mayor que otras variantes de la literatura.  
Hemos observado que la incertidumbre aleatórica es razonablemente fácil de capturar. No obstante, es importante destacar la relevancia de la incertidumbre epistémica como una parte significativa de la incertidumbre total. 
Los métodos escalables tienen todavía potencial de mejora, dado que, lograr una calibración perfecta de la incertidumbre sigue siendo un objetivo abierto.

En el Capítulo \ref{chap:2}, realizamos una investigación sobre la cuantificación de la incertidumbre respecto a la estimación de profundidad. En este caso, nos enfocamos en las imágenes médicas de colonoscopia, aportando una perspectiva novedosa a la literatura establecida.

Nuestra motivación es resaltar la importancia de estimaciones precisas y calibradas con respecto a la profundidad y la incertidumbre en las imágenes médicas. En concreto, en las imágenes de colonoscopia, la incertidumbre de profundidad es más pronunciada en los reflejos especulares y las regiones sobreexpuestas, en zonas oscuras donde la señal de color es más débil y también las discontinuidades de profundidad, como las haustras. 
Se demuestra empíricamente que la incertidumbre capturada por deep ensembles se comporta de forma coherente cuando se produce un cambio de dominio. Este hallazgo tiene una gran relevancia, ya que en el dominio endoscópico es muy difícil disponer de anotaciones ground truth y, por lo tanto, la transferencia de dominio de simulación a real es una práctica común.

En nuestra investigación proponemos un novedoso método autosupervisado basado en una arquitectura teacher-student que incorpora la incertidumbre del teacher en el proceso de aprendizaje. Nuestros experimentos muestran que existe una notable eficacia a la hora de reducir los errores de profundidad y mejorar la calibración de la incertidumbre en comparación con otros métdos de aprendizaje. Nuestro método representa un avance prometedor, aunque su eficacia depende en gran medida de la habilidad del teacher.
Por último, en el cuarto Capítulo, proponemos un nuevo aprendizaje autosupervisado desde una única vista, el cual modela la atenuación cuadrática de la luz y desbloquea el aprendizaje autosupervisado de una sola vista para imágenes médicas naturales. Nuestro método denominado LightDepth ofrece un nuevo enfoque para la autosupervisión de la profundidad, ya que permite una estimación de profundidad sin la necesidad de supervisión con ground truth, basándose en el uso de la iluminación como señal supervisora. Nuestra investigación demuestra que la autosupervisión utilizando la información de la iluminación es significativamente superior a la autosupervisión multivista, e incluso iguala a los métodos de supervisión con ground truth.
Dado que el aprendizaje propuesto requiere únicamente el uso de una sola vista para el entrenamiento, hace que nuestro método sea capaz de refinar sus predicciones con una optimización local, lo que mejora aún más las predicciones de profundidad.

Nuestros resultados indican que, aunque los métodos supervisados pueden ser eficaces para la visión endoscópica, su rendimiento se ve limitado por las particularidades del entorno. La falta de sensores de profundidad en los procedimientos mínimamente invasivos supone una dificultad a la hora de capturar datos con anotaciones de profundidad ground truth, valiosos para el entrenamiento supervisado. 
Finalmente, observamos que la autosupervisión multivista se ve limitada por las dificultades para estimar el movimiento relativo de la cámara entre imágenes, lo que también constituye un reto de investigación en sí mismo.
Creemos que nuestros resultados abren un camino prometedor, la autosupervisión de una sola vista supera las limitaciones de ambos enfoques y sólo requiere datos RGB, que no plantean dificultades para ser adquiridos.

\bibliographystyle{plainnat}  
\bibliography{bibliography}

\nocite{*} 

\newpage
\renewcommand\listfigurename{List of Figures}
\listoffigures

\newpage
\renewcommand\listtablename{List of Tables}
\listoftables

\newpage
\clearpage
\addappheadtotoc

\end{document}